%% file: main.tex
\PassOptionsToPackage{table}{xcolor}
\documentclass{article}
\usepackage{iclr2027_conference}
\usepackage{times}

\input{math_commands.tex}

\usepackage{hyperref}
\usepackage{url}
\usepackage{graphicx}
\usepackage{booktabs}
\usepackage{multirow}
\usepackage{amsmath,amssymb}
\usepackage{xcolor}
\usepackage{tikz}
\usetikzlibrary{arrows.meta,positioning,shapes.geometric,calc}
\usepackage{caption}
\usepackage{subcaption}
\usepackage{enumitem}
\usepackage{tcolorbox}
\usepackage{alltt}

\tcbuselibrary{breakable}
\usepackage[capitalise,noabbrev]{cleveref} 
\graphicspath{{figures/}}

\newcommand{\changed}[1]{#1}
\newcommand{\changednew}[1]{#1}

\newcommand{\changenote}[1]{}

\hypersetup{hidelinks}

\input{data/micropolis-macros.tex}

\input{data/combined_score_macros.tex}
\input{data/combined_sensitivity_macros.tex}
\input{data/simulator_scores_macros.tex}
\input{data/generalization_macros.tex}
\input{data/freeciv_macros.tex}
\input{data/freeciv_grouping_macros.tex}

\input{data/baseline_macros.tex}

\input{data/external_benchmarks_macros.tex}

\title{ForecastBench-Sim: Forecasting Simulated Worlds}

\author{Jaeho Lee\textsuperscript{1}, Nick Merrill\textsuperscript{1}, Fabio D Rocha\textsuperscript{1}, Ezra Karger\textsuperscript{2}\\[3pt]
\normalfont\textsuperscript{1}\href{https://forecastingresearch.org}{Forecasting Research Institute}\\
\normalfont\textsuperscript{2}\href{https://chicagofed.org}{Federal Reserve Bank of Chicago}}

\iclrfinalcopy
\begin{document}
\maketitle

\input{sections/01_abstract}
\input{sections/02_introduction}
\input{sections/03_benchmark_design}
\input{sections/04_micropolis}
\input{sections/05_freeciv}
\input{sections/06_starsim}
\input{sections/07_real_world_forecasting}
\input{sections/08_related_work}
\input{sections/09_discussion}

\subsubsection*{AI use statement}
\input{sections/10_ai_use}
\subsubsection*{Reproducibility statement}
\input{sections/11_reproducibility}

\bibliography{references}
\bibliographystyle{iclr2027_conference}

\clearpage
\appendix

\input{appendix/A_worlds_and_protocol}
\input{appendix/B_prompts_and_templates}
\input{appendix/C_full_results}
\input{appendix/D_ablations}

\input{appendix/E_validation}
\input{appendix/F_release}

\end{document}

%% file: math_commands.tex
\usepackage{amsmath,amsfonts,bm}

\def\eqref#1{equation~\ref{#1}}

\def\1{\bm{1}}

\DeclareMathAlphabet{\mathsfit}{\encodingdefault}{\sfdefault}{m}{sl}
\SetMathAlphabet{\mathsfit}{bold}{\encodingdefault}{\sfdefault}{bx}{n}



%% file: data/micropolis-macros.tex
\newcommand{\MPDRhoBinaryShort}{0.80}
\newcommand{\MPDRhoBinaryCIModels}{$[0.60,\,0.89]$}

\newcommand{\MPDPBinary}{\ensuremath{<}0.001}
\newcommand{\MPDNModelsBinary}{24}
\newcommand{\MPDNQuestionsBinary}{983}

\newcommand{\MPDBestBinaryModel}{GPT-5.6 Sol}

\newcommand{\MPDRhoTailShort}{0.75}
\newcommand{\MPDRhoTailCIModels}{$[0.48,\,0.87]$}

\newcommand{\MPDPTail}{\ensuremath{<}0.001}
\newcommand{\MPDNModelsTail}{24}
\newcommand{\MPDNQuestionsTail}{940}

\newcommand{\MPDBestTailModel}{GPT-5.5}

\newcommand{\MPDRhoContinuousShort}{0.82}
\newcommand{\MPDRhoContinuousCIModels}{$[0.56,\,0.92]$}

\newcommand{\MPDPContinuous}{\ensuremath{<}0.001}
\newcommand{\MPDNModelsContinuous}{24}
\newcommand{\MPDNQuestionsContinuous}{600}

\newcommand{\MPDRhoHorizonMin}{0.71}
\newcommand{\MPDRhoHorizonMax}{0.82}

\newcommand{\MPDMinParseBinary}{98.3}
\newcommand{\MPDMinParseContinuous}{92.0}

\newcommand{\MPDPromptsBinary}{60}
\newcommand{\MPDItemsBinary}{3,000}

\newcommand{\MPDPerPromptBinary}{50}

\newcommand{\MPDPromptsContinuous}{30}

\newcommand{\MPDPerPromptContinuous}{20}
\newcommand{\MPDCostTotal}{32.31}
\newcommand{\MPDCallsTotal}{2,160}

\newcommand{\MPDBandNQuestions}{3,000}
\newcommand{\MPDBandZeroShare}{27.5}

\newcommand{\MPDBandTailShare}{31.3}

\newcommand{\MPDBandMidShare}{32.8}

\newcommand{\MPDBandTopShare}{8.4}

\newcommand{\MPDNExcludedZero}{826}
\newcommand{\MPDNExcludedTop}{251}
\newcommand{\MPDMeanQBinary}{0.28}

\newcommand{\MPDHorizonYears}{3, 5, 7 and 10}

\newcommand{\MPDRhoBinaryCICities}{$[0.74,\,0.84]$}
\newcommand{\MPDRhoTailCICities}{$[0.66,\,0.81]$}
\newcommand{\MPDRhoContinuousCICities}{$[0.66,\,0.85]$}

\newcommand{\MPDBatchNModels}{8}

\newcommand{\MPDBatchTailOne}{0.241}

\newcommand{\MPDBatchTailTwo}{0.154}

\newcommand{\MPDBatchTailFour}{0.137}

\newcommand{\MPDBatchDeltaTailFour}{$-0.104$}

\newcommand{\MPDBatchDeltaTailFourCIModels}{$[-0.20,\,-0.04]$}

\newcommand{\MPDBatchSizeEight}{7--8}

\newcommand{\MPDBatchDeltaMidEight}{$-0.0153$}

\newcommand{\MPDBatchDeltaMidEightCIModels}{$[-0.03,\,-0.00]$}

\newcommand{\MPDBatchTailBiasOne}{$+0.151$}

\newcommand{\MPDBatchTailBiasFour}{$+0.096$}

\newcommand{\MPDBatchTailBiasSixtyFour}{$+0.072$}

\newcommand{\MPDBatchDiscrimMin}{0.60}
\newcommand{\MPDBatchDiscrimMax}{0.65}

\newcommand{\MPDBatchAnchorModel}{GPT-5 Mini}
\newcommand{\MPDBatchAnchorShareSmallest}{24.5}
\newcommand{\MPDBatchAnchorShareSecond}{1.9}

\newcommand{\MPDBatchCopierModel}{GPT-4.1}

\newcommand{\MPDBatchTailSmallestNoExample}{0.208}

\newcommand{\MPDVarNModels}{24}
\newcommand{\MPDVarNVariants}{11}
\newcommand{\MPDVarNQuestions}{600}
\newcommand{\MPDVarNCities}{15}
\newcommand{\MPDVarNPrompts}{30}

\newcommand{\MPDVarMain}{\texttt{20yhist}}

\newcommand{\MPDVarRhoMin}{0.64}
\newcommand{\MPDVarRhoMax}{0.86}
\newcommand{\MPDVarRhoMinVariant}{\texttt{100yhist}}
\newcommand{\MPDVarRhoMaxVariant}{\texttt{preamble2}}

\newcommand{\MPDVarSemanticMean}{0.054}

\newcommand{\MPDVarSemlongerMean}{0.034}

\newcommand{\MPDVarHundredYHistMean}{0.020}

\newcommand{\MPDKnowNStatements}{206}
\newcommand{\MPDKnowNPairs}{84}
\newcommand{\MPDKnowNTrue}{94}
\newcommand{\MPDKnowNFalse}{112}
\newcommand{\MPDKnowNHoneypots}{20}
\newcommand{\MPDKnowNEngine}{152}
\newcommand{\MPDKnowNCities}{28}
\newcommand{\MPDKnowNDynamics}{26}
\newcommand{\MPDKnowNDifficultyZero}{28}
\newcommand{\MPDKnowNDifficultyOne}{91}
\newcommand{\MPDKnowNDifficultyTwo}{87}

\newcommand{\MPDKnowPerPrompt}{103}
\newcommand{\MPDKnowNModels}{24}
\newcommand{\MPDKnowCost}{0.93}

\newcommand{\MPDKnowNAbstainers}{3}
\newcommand{\MPDKnowAbstainers}{GPT-5 Nano, Claude Haiku 4.5, GPT-5.4 Nano}
\newcommand{\MPDKnowUnknownMax}{98}
\newcommand{\MPDKnowUnknownMaxModel}{GPT-5 Nano}
\newcommand{\MPDKnowHoneypotTrueMean}{16}
\newcommand{\MPDKnowHoneypotTrueMax}{60}
\newcommand{\MPDKnowHoneypotTrueMaxModel}{GPT-5 Mini}
\newcommand{\MPDKnowNHoneypotClean}{7}
\newcommand{\MPDKnowAuthor}{Claude Fable}
\newcommand{\MPDKnowAuthorScore}{0.90}
\newcommand{\MPDKnowAuthorRank}{1}
\newcommand{\MPDKnowRhoAllNoAuthor}{0.65}

\newcommand{\MPDResamples}{10,000}
\newcommand{\MPDSeed}{2026}

\newcommand{\MPDFBRhoBinaryShort}{0.38}
\newcommand{\MPDFBRhoBinaryCIModels}{$[-0.20,\,0.80]$}
\newcommand{\MPDFBPBinary}{0.135}

\newcommand{\MPDFBRhoTailShort}{0.45}
\newcommand{\MPDFBRhoTailCIModels}{$[-0.03,\,0.78]$}
\newcommand{\MPDFBPTail}{0.067}

\newcommand{\MPDFBRhoContinuousShort}{0.35}
\newcommand{\MPDFBRhoContinuousCIModels}{$[-0.22,\,0.76]$}
\newcommand{\MPDFBPContinuous}{0.163}

%% file: data/combined_score_macros.tex
\newcommand{\CombinedRho}{0.86}
\newcommand{\CombinedLo}{0.69}
\newcommand{\CombinedHi}{0.93}
\newcommand{\CombinedResidual}{0.35}
\newcommand{\CombinedLooLo}{0.86}
\newcommand{\CombinedLooHi}{0.88}
\newcommand{\CombinedCoreAgreementLo}{0.74}
\newcommand{\CombinedCoreAgreementHi}{0.92}
\newcommand{\CombinedFreeAgreementLo}{0.20}
\newcommand{\CombinedFreeAgreementHi}{0.65}

%% file: data/combined_sensitivity_macros.tex
\newcommand{\CombinedRhoThree}{0.863}
\newcommand{\CombinedCompleteRho}{0.860}
\newcommand{\CombinedRepsRho}{0.874}
\newcommand{\CombinedBinaryRho}{0.888}
\newcommand{\CombinedMatchedEciRho}{0.794}
\newcommand{\CombinedLooMaxMove}{five}
\newcommand{\CombinedCellSdLo}{0.149}
\newcommand{\CombinedCellSdHi}{0.892}
\newcommand{\CombinedEqualSimRho}{0.878}
\newcommand{\CombinedEqualSimAgreement}{0.986}
\newcommand{\CombinedEqualSimMaxMove}{four}
\newcommand{\FreeCivMidTailRankCorr}{\ensuremath{-}0.703}

%% file: data/simulator_scores_macros.tex
\newcommand{\SimScoreRhoLo}{0.68}
\newcommand{\SimScoreRhoHi}{0.82}
\newcommand{\SimScorePartialLo}{0.14}
\newcommand{\SimScorePartialHi}{0.43}

%% file: data/generalization_macros.tex
\newcommand{\GenHalfRho}{0.94}
\newcommand{\GenHalfMin}{0.89}
\newcommand{\GenHalfSplits}{1{,}000}
\newcommand{\GenHalfMicropolis}{0.94}
\newcommand{\GenHalfFreeCiv}{0.85}
\newcommand{\GenHalfStarsim}{0.86}
\newcommand{\GenLosoLo}{0.79}
\newcommand{\GenLosoHi}{0.84}

%% file: data/freeciv_macros.tex
\newcommand{\FCMidMean}{0.141}

\newcommand{\FCMidRho}{0.18}

\newcommand{\FCRhoMin}{0.18}
\newcommand{\FCRhoMax}{0.59}

\newcommand{\FCMidNegBias}{24}
\newcommand{\FCMidBiasMin}{\ensuremath{-}0.33}
\newcommand{\FCMidBiasMax}{\ensuremath{-}0.07}
\newcommand{\FCMidDiscMin}{0.09}
\newcommand{\FCMidDiscMinModel}{Llama 4 Scout}
\newcommand{\FCMidDiscMax}{0.27}
\newcommand{\FCMidDiscMaxModel}{Gemini 3.7 Flash}
\newcommand{\FCMidDiscRho}{0.72}
\newcommand{\FCMidDiscLo}{0.48}
\newcommand{\FCMidDiscHi}{0.85}
\newcommand{\FCMidBiasEciRho}{\ensuremath{-}0.07}
\newcommand{\FCMidExcessAbsBiasRho}{0.87}
\newcommand{\FCMidExcessDiscRho}{\ensuremath{-}0.39}
\newcommand{\FCMidCompBias}{\ensuremath{+}0.00}
\newcommand{\FCMidCompDisc}{0.48}
\newcommand{\FCMidUnderLo}{0.21}
\newcommand{\FCMidUnderHi}{0.26}
\newcommand{\FCMidHorizonLo}{0.136}
\newcommand{\FCMidHorizonHi}{0.149}
\newcommand{\FCMidHorizonFirst}{0.136}
\newcommand{\FCMidHorizonLast}{0.139}
\newcommand{\FCMidRelPLow}{0.26}
\newcommand{\FCMidRelQLow}{0.09}
\newcommand{\FCMidRelPHigh}{0.38}
\newcommand{\FCMidRelQHigh}{0.91}
\newcommand{\FCMidRelWhere}{they lie above the truth in the two lowest intervals and below it in the eight highest}
\newcommand{\FCMidLowF}{0.049}
\newcommand{\FCMidLowQ}{0.436}
\newcommand{\FCMidLowRows}{3,189}
\newcommand{\FCMidHighF}{0.922}
\newcommand{\FCMidHighQ}{0.699}
\newcommand{\FCMidHighRows}{261}
\newcommand{\FCMidDecTotal}{0.141}
\newcommand{\FCMidDecBias}{0.041}
\newcommand{\FCMidDecVarF}{0.050}
\newcommand{\FCMidDecVarQ}{0.075}
\newcommand{\FCMidDecCov}{0.025}
\newcommand{\FCMidDecBiasShare}{29\%}
\newcommand{\FCMidWorstItems}{69}
\newcommand{\FCMidWorstShare}{14\%}
\newcommand{\FCMidMeanQ}{0.501}
\newcommand{\FCTailMean}{0.33}

\newcommand{\FCTailRho}{0.34}

\newcommand{\FCTailMeanFMin}{0.13}
\newcommand{\FCTailMeanFMax}{0.34}
\newcommand{\FCTailMeanQ}{0.02}

\newcommand{\FCTailHorizonFirst}{0.28}
\newcommand{\FCTailHorizonLast}{0.41}
\newcommand{\FCTailWorstItems}{63}
\newcommand{\FCTailWorstShare}{36\%}
\newcommand{\FCMirBest}{0.263}
\newcommand{\FCMirBestModel}{Gemini 3 Flash}
\newcommand{\FCMirWorst}{0.664}
\newcommand{\FCMirWorstModel}{GPT-5 Nano}
\newcommand{\FCMirFlat}{0.233}
\newcommand{\FCMirMeanQ}{0.982}
\newcommand{\FCMirRho}{0.10}
\newcommand{\FCMirRhoLo}{\ensuremath{-}0.36}
\newcommand{\FCMirRhoHi}{0.51}
\newcommand{\FCMirRhoP}{0.64}

%% file: data/freeciv_grouping_macros.tex
\newcommand{\FCGroupDirection}{Tail loss falls while mid-range, mirror and continuous losses rise}
\newcommand{\FCGroupRhoChangeMax}{0.06}
\newcommand{\FCGroupRankCorrLo}{0.68}
\newcommand{\FCGroupRankCorrHi}{0.85}
\newcommand{\FCGroupMoves}{13, 14, 7 and 12}
\newcommand{\FCGroupOverlapSingle}{0.42}
\newcommand{\FCGroupOverlapGrouped}{0.33}
\newcommand{\FCGroupOverlapTruth}{0.49}
\newcommand{\FCGroupShuffledDiffMax}{0.013}
\newcommand{\FCGroupShuffledRankLo}{0.90}
\newcommand{\FCGroupShuffledRankHi}{0.95}

%% file: data/baseline_macros.tex
\newcommand{\BLMPMidBeat}{17}

\newcommand{\BLMPTailBeat}{16}

\newcommand{\BLFCMid}{0.074}

\newcommand{\BLFCMidBest}{0.103}

\newcommand{\BLFCTailBeat}{9}

\newcommand{\BLMPCont}{0.037}
\newcommand{\BLMPContBeat}{11}

\newcommand{\BLFCCont}{0.070}

\newcommand{\BLFCContBest}{0.402}

\newcommand{\BLSSCont}{1.68}
\newcommand{\BLSSContBeat}{19}

\newcommand{\BLSSContSIR}{0.414}
\newcommand{\BLSSContSIRBeat}{7}

\newcommand{\BLFCNatcond}{0.064}

\newcommand{\BLSSInt}{2.86}
\newcommand{\BLSSIntBeat}{23}

\newcommand{\BLFCNatcondTrueP}{0.015}

\newcommand{\BLSSIntTrueNoVax}{5.37}

\newcommand{\BLSSIntOwnBeat}{22}
\newcommand{\BLSSIntOwnN}{23}

\newcommand{\BLSSSIRErrMin}{2}
\newcommand{\BLSSSIRErrMax}{12}

\newcommand{\BLFCMidConst}{0.075}

\newcommand{\BLFCMidAsked}{0.086}
\newcommand{\BLFCTailAsked}{0.304}
\newcommand{\BLFCCandidates}{23,241}
\newcommand{\BLFCContCandidates}{1,462}
\newcommand{\BLFCQuestionsPerRate}{28}
\newcommand{\BLFCSelected}{0.075}

\newcommand{\BLFCSelectedBest}{0.107}
\newcommand{\BLFCMidMeanForecast}{0.31}
\newcommand{\BLFCMidMeanQ}{0.50}
\newcommand{\BLFCMidMeanRate}{0.48}
\newcommand{\BLFCTailMeanForecast}{0.21}
\newcommand{\BLFCTailMeanQ}{0.02}
\newcommand{\BLFCTailMeanRate}{0.17}
\newcommand{\BLFCMidCorrForecast}{0.19}
\newcommand{\BLFCMidCorrForecastMax}{0.27}
\newcommand{\BLFCMidCorrRate}{0.47}
\newcommand{\BLFCRankCorrForecastLevel}{0.94}

%% file: data/external_benchmarks_macros.tex
\newcommand{\ExtFETheta}{0.90}

\newcommand{\ExtFEThetaGivenEci}{0.59}
\newcommand{\ExtFEThetaGivenEciCI}{[-0.04, 0.88]}
\newcommand{\ExtFEEciGivenTheta}{0.52}

\newcommand{\ExtPATheta}{0.86}

\newcommand{\ExtPAThetaGivenEci}{0.67}
\newcommand{\ExtPAThetaGivenEciCI}{[-0.47, 1.00]}
\newcommand{\ExtPAEciGivenTheta}{0.22}

\newcommand{\ExtFBTheta}{0.37}

\newcommand{\ExtFBThetaGivenEci}{-0.25}
\newcommand{\ExtFBThetaGivenEciCI}{[-0.70, 0.49]}
\newcommand{\ExtFBEciGivenTheta}{0.57}

\newcommand{\ExtMatchedN}{14}
\newcommand{\ExtMatchedFETheta}{0.85}

\newcommand{\ExtMatchedFBTheta}{0.46}

\newcommand{\ExtMatchedDiffTheta}{0.38}
\newcommand{\ExtMatchedDiffThetaCI}{[0.03, 0.79]}

\newcommand{\ExtBetweenFBFE}{0.77}
\newcommand{\ExtBetweenFBPA}{0.60}
\newcommand{\ExtBetweenFEPA}{0.86}

%% file: sections/01_abstract.tex
\begin{abstract}
Predicting how a complex system will evolve is among the most general tests of reasoning under uncertainty, and among the hardest to benchmark. Real-world forecasting benchmarks often resolve slowly and provide sparse evidence of accuracy in forecasting rare or counterfactual events. We introduce ForecastBench-Sim, a forecasting benchmark that asks models to forecast the outcomes of simulated worlds. We generate questions about rare events, conditional outcomes, and responses to causal interventions. Across 24 models, three simulations, and nine distinct tasks, model scores closely track general measures of model capabilities ($\rho=\CombinedRho$ [$\CombinedLo,\CombinedHi$]). ForecastBench-Sim provides tasks whose question sets can be refreshed, varied in difficulty, and immediately scored. We discuss potential mechanisms to prevent saturation and avoid benchmark gaming, focusing on the automatable extensibility of our framework to new unseen questions.
\end{abstract}

%% file: sections/02_introduction.tex
\section{Introduction}
\label{sec:intro}

Forecasting is an increasingly valuable skill for large language models (LLMs), and recent results indicate that LLM-based forecasters are approaching the best human forecasters \citep{halawi2024forecasting,fri2026forecastbenchparity}. However, existing forecasting benchmarks (like ForecastBench; \citealp{karger2024forecastbench}) rely on observed, real-world forecasting questions, which bring a few notable limitations. First, live forecasting questions must resolve, which can take weeks, if not months, forcing evaluators to wait to see evaluation results. Historical backtesting permits immediate scoring, but resolutions may already be encoded in model parameters (and, in a competitive benchmark, we cannot rely on reported knowledge cutoffs to rule out contamination). Second, rare events are sparse in real-world data by nature, making it difficult to estimate how well models perform on forecasting such questions. Third, interventional and correlative conditional questions are poorly suited to real-world benchmarks. The real world provides us with only a single timeline. For interventions (i.e., causal effects), the real world cannot reveal both an intervention outcome and its counterfactual. For correlative conditionals, the real world does not always provide suitable repeated observations to estimate the true likelihood of two events co-occurring.

We introduce ForecastBench-Sim, a simulated-world forecasting benchmark. Fresh, privately generated worlds allow immediate evaluation without relying on reported knowledge cutoffs. ForecastBench-Sim relies on forecasting simulated worlds: it runs a world up to a snapshot point, then describes its state in a world report. The benchmark then poses questions to the model under evaluation about that world's future state. These questions are resolved by playing the world forward from the snapshot (\Cref{fig:schematic}). World plays are deterministic given a seed; by replaying the same snapshot with different seeds, we can estimate true event probabilities. We can then score a forecast by its excess score (i.e., the loss beyond that of the ground-truth probability itself). \Cref{sec:design} defines these terms.

Our contributions are as follows.
\begin{itemize}[leftmargin=*,itemsep=1pt,topsep=2pt]
\item A benchmark design (\Cref{sec:design}) under which any simulation can yield forecasting questions that are evaluable against true probability estimates, including rare and conditional forecasting questions.
\item Results for a shared panel of 24 models on three simulations: Micropolis \citep{micropolis2008} for tail questions, FreeCiv \citep{freeciv2026} for natural conditionals and Starsim \citep{kerr2024starsim} for interventional conditionals (\Cref{sec:micropolis,sec:freeciv,sec:starsim}).
\item A case study demonstrating that a forecasting failure observed in our simulated worlds also appears in a real-world forecasting case (historical measles cases) (\cref{sec:real-world-forecasting}), motivating future work on simulation-based forecasting training.
\item A validation against the Epoch Capabilities Index (ECI) \citep{epoch2026eci} and existing forecasting benchmarks (\cref{sec:validation}). We find high correlation with two out of three existing forecasting benchmarks ($\rho=\ExtFETheta$ on Metaculus FutureEval and $\rho=\ExtPATheta$ on Prophet Arena). The partial-correlation estimates remain positive after controlling for ECI, although both confidence intervals include zero, suggesting that ForecastBench-Sim may measure some aspects of forecasting ability beyond pure capability.
\item A planned release of all code required to reproduce our results, or extend our results with new simulations (\cref{app:release}).
\end{itemize}

%% file: sections/03_benchmark_design.tex
\section{Benchmark design}
\label{sec:design}
\label{sec:benchmark-design}

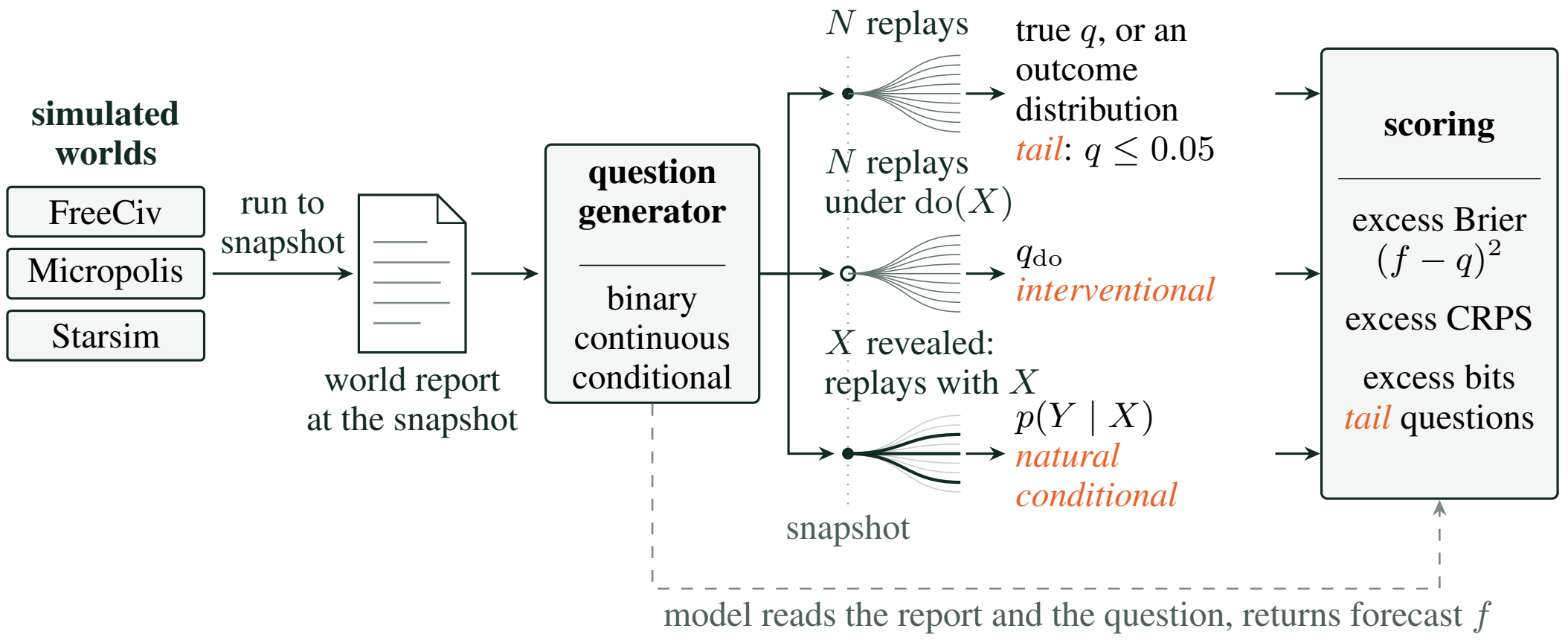
\begin{figure*}[t]
  \centering
  \input{figures/fig_schematic}
  \caption{ForecastBench-Sim generates world reports and forecasting questions from simulator snapshots. Replays under new random seeds estimate outcome probabilities or distributions, against which model forecasts are scored by their excess loss.}
  \label{fig:schematic}
\end{figure*}

\href{https://github.com/forecastingresearch/forecastbench-sim/tree/arxiv}{ForecastBench-Sim} uses \emph{simulators} to generate forecasting questions. We use two games and one disease model; the important feature is that the simulators are (a) complex and (b) deterministic given a seed. A \emph{world} is a simulated scenario within a simulation: a city in Micropolis, a game in FreeCiv, or a scenario containing one or two regions in Starsim. A simulator therefore supplies many worlds, and every question is asked about a snapshot of one world. We run the world to a chosen turn, the \emph{snapshot}, and write a \emph{world report}, a text document that gives the state at the snapshot, the history of the main quantities and the events so far. Forecasters (in practice, models) see only the world report and questions. Worlds are then run many times with different seeds, creating a variety of \emph{replays}, from which we can estimate the ground truth probability for any given event (\cref{fig:schematic}).

\subsection{Questions and ground truth}
\label{sec:design-questions}

ForecastBench-Sim allows multiple types of questions. For \emph{binary questions}, the ground-truth probability $q$ is the share of the $N$ replays in which the event occurs. For a \emph{continuous question}, the ground truth is the outcome distribution of the asked quantity over the replays; forecasters answer with per-quantile estimates.  A \emph{tail question} (i.e., a question with a rare outcome) is a binary question with $0 < q \le 0.05$, and a \emph{mid-range question} one with $0.05 < q < 0.95$; both are selected on $q$ before any forecast is made, so a world contributes as many rare events as its replays contain.

ForecastBench-Sim also allows conditional questions. A \emph{natural conditional} reveals one true fact $X$ about the interval after the snapshot and asks the same question $Y$ again. Its ground truth is $p(Y\mid X)$, the share of replays in which $Y$ occurs among those in which $X$ holds. An \emph{interventional conditional} applies an intervention (e.g. a vaccination campaign) to a snapshot and asks about the world conditional on the intervention being applied. Its ground truth comes from replays of the world with and without the intervention.

\subsection{Scoring}
\label{sec:design-scoring}

We score forecasts against replay ground truth using proper scoring rules or their quantile approximations, and report the \emph{excess score}: expected forecast loss minus expected loss at the ground truth. Mid-range binary questions use \emph{excess Brier score}, $(f-q)^2$ \citep{brier1950verification}, and tail questions use \emph{excess bits}, $\mathrm{KL}(q\,\|\,f)$. Continuous questions use pinball losses at the elicited quantiles to approximate CRPS \citep{matheson1976scoring}; subtracting the score at the replay distribution's own quantiles gives \emph{excess CRPS}, normalized by fixed scales to give \emph{excess nCRPS}. \Cref{app:scoring} gives derivations, normalization, clipping and finite-quantile caveats.

\paragraph{Baselines.}\label{sec:design-baselines}
For binary questions, \emph{base rates} average probabilities for matching questions and horizons in other worlds. For continuous questions, \emph{climatological change} adds other worlds' changes to the current value; for conditionals, \emph{independence} uses the unconditional baseline. Starsim also has an SIR forecast using only the world report. \Cref{app:baselines} gives constructions and alternatives.

\subsection{Simulations, models and protocol}
\label{sec:design-worlds}
\label{sec:design-protocol}
\label{sec:design-models}

\Cref{tab:worlds} summarizes the three simulations; \cref{sec:micropolis,sec:freeciv,sec:starsim} give the particulars of each, and \cref{app:worlds} the program versions, seeds, run counts and replacements. We ran the same 24 models in every simulation through OpenRouter. We chose them so that each has a score on the Epoch Capabilities Index (ECI), a single capability scale estimated across many public benchmarks by an item-response model \citep{ho2025rosetta,epoch2026eci}. We requested the lowest supported reasoning setting where controllable, or a 1,024-token budget for budget-only models; four models offered no control, and requested settings did not equalize compute (\cref{app:roster}).

\begin{table}[t]
  \caption{The three simulators and their parameters. Parameters necessarily vary somewhat between simulators, as each simulator and question under evaluation (tail events, interventional conditionals, natural conditionals) requires a bespoke setup.}
  \label{tab:worlds}
  \centering
  \small
  \setlength{\tabcolsep}{4pt}
  \begin{tabular}{@{}>{\raggedright\arraybackslash}p{2.55cm}>{\raggedright\arraybackslash}p{3.35cm}>{\raggedright\arraybackslash}p{3.5cm}>{\raggedright\arraybackslash}p{3.35cm}@{}}
    \toprule
    & Micropolis & FreeCiv & Starsim \\
    \midrule
    Substrate & Open-source release of SimCity Classic \citep{micropolis2008}; no player input; disasters on & Open-source clone of Civilization \citep{freeciv2026}; five civilizations competing with in-game AI on ``hard'' & Starsim \citep{kerr2024starsim}; disease simulator \\
    Worlds & 15 cities, 2 snapshots each & 8 games & 4 worlds, 1 snapshot each \\
    Snapshot & game years 20 and 30 (turns 960 and 1,440) & turn 60 & day 20 \\
    Horizons & 3, 5, 7 and 10 game years after the snapshot & turns 90, 120, 150, 180 and 210 & days 40 and 60 \\
    Replays per snapshot & 1,000 & 1,000 & 1,000 per arm for continuous rescore; binary retains original matched seeds \\
    Items per model & \MPDItemsBinary\ binary asked, of which \MPDNQuestionsBinary\ mid-range and \MPDNQuestionsTail\ tail; \MPDNQuestionsContinuous\ continuous & 750 mid-range; 300 tail; 300 continuous; 400 natural-conditional cells & 8 continuous; 24 interventional; 24 historical binary items in appendix \\
    \bottomrule
  \end{tabular}
\end{table}

%% file: figures/fig_schematic.tex
\definecolor{figgreen}{HTML}{102B23}%
\definecolor{figorange}{HTML}{E8632C}%
\begin{tikzpicture}[
  x=1cm, y=1cm,
  font=\footnotesize,
  >={Stealth[length=4pt,width=3.5pt]},
  every node/.style={inner sep=2pt},
  box/.style={draw=figgreen, line width=0.6pt, fill=figgreen!4, rounded corners=1.5pt,
              align=center, inner sep=3.5pt},
  wbox/.style={draw=figgreen, line width=0.6pt, fill=figgreen!4, rounded corners=1pt,
               minimum width=1.75cm, minimum height=0.44cm, inner sep=1pt},
  ttl/.style={align=center, text=figgreen, font=\footnotesize\bfseries, inner sep=1pt},
  capt/.style={align=left, text=figgreen, inner sep=1pt},
  res/.style={align=left, inner sep=1pt},
  tag/.style={text=figorange, font=\footnotesize\itshape},
  arr/.style={->, draw=figgreen, line width=0.6pt},
  fan/.style={draw=figgreen!65, line width=0.3pt},
  fank/.style={draw=figgreen, line width=0.8pt},
  fanf/.style={draw=figgreen!22, line width=0.3pt},
  fpath/.style={->, draw=figgreen!55, line width=0.5pt, dashed},
]

\node[ttl] at (0.85,2.85) {simulated\\worlds};
\node[wbox] at (0.85,2.15) {FreeCiv};
\node[wbox] at (0.85,1.6)  {Micropolis};
\node[wbox] at (0.85,1.05) {Starsim};
\draw[arr] (1.8,1.6) -- (3.05,1.6)
  node[midway, above=2pt, capt, align=center] {run to\\snapshot};

\draw[draw=figgreen, line width=0.6pt, fill=white]
  (3.1,0.9) -- (3.1,2.3) -- (3.8,2.3) -- (4.05,2.05) -- (4.05,0.9) -- cycle;
\draw[draw=figgreen, line width=0.5pt] (3.8,2.3) -- (3.8,2.05) -- (4.05,2.05);
\foreach \yy/\len in {1.88/0.48, 1.7/0.68, 1.52/0.6, 1.34/0.68, 1.16/0.42}
  \draw[figgreen!55, line width=0.5pt] (3.23,\yy) -- ++(\len,0);
\node[capt, align=center, anchor=north] at (3.575,0.82) {world report\\at the snapshot};
\draw[arr] (4.1,1.6) -- (4.7,1.6);

\node[box, minimum width=1.9cm, anchor=west] (gen) at (4.75,1.6)
  {{\bfseries question}\\{\bfseries generator}\\[1pt]\rule{1.3cm}{0.3pt}\\[1pt]
   binary\\continuous\\conditional};
\draw[arr] (gen.east) -- ++(0.25,0) |- (7.33,3.2);
\draw[arr] (gen.east) -- (7.33,1.6);
\draw[arr] (gen.east) -- ++(0.25,0) |- (7.33,0.0);

\draw[figgreen!45, line width=0.4pt, dotted] (7.45,-0.45) -- (7.45,3.55);
\node[capt, align=center, anchor=north, text=figgreen!75] at (7.45,-0.5) {snapshot};

\node[capt, anchor=south west] at (7.2,3.62) {$N$ replays};
\fill[figgreen] (7.45,3.2) circle (1.5pt);
\foreach \i in {-4,...,4}
  \draw[fan] (7.45,3.2) .. controls (7.95,3.2) and (7.95,{3.2+0.085*\i}) .. (8.45,{3.2+0.085*\i});
\draw[arr] (8.5,3.2) -- (8.85,3.2);
\node[res, anchor=west] at (8.9,3.2)
  {true $q$, or an\\outcome\\distribution\\{\color{figorange}\itshape tail}: $q \le 0.05$};

\node[capt, anchor=south west, align=left] at (7.2,2.02) {$N$ replays\\under $\mathrm{do}(X)$};
\draw[draw=figgreen, line width=0.7pt, fill=white] (7.45,1.6) circle (1.7pt);
\foreach \i in {-4,...,4}
  \draw[fan] (7.45,1.6) .. controls (7.95,1.6) and (7.95,{1.6+0.085*\i}) .. (8.45,{1.6+0.085*\i});
\draw[arr] (8.5,1.6) -- (8.85,1.6);
\node[res, anchor=west] at (8.9,1.6)
  {$q_{\mathrm{do}}$\\{\color{figorange}\itshape interventional}};

\node[capt, anchor=south west, align=left] at (7.2,0.42) {$X$ revealed:\\replays with $X$};
\fill[figgreen] (7.45,0.0) circle (1.5pt);
\foreach \i in {-4,-2,-1,1,3,4}
  \draw[fanf] (7.45,0.0) .. controls (7.95,0.0) and (7.95,{0.085*\i}) .. (8.45,{0.085*\i});
\foreach \i in {-3,0,2}
  \draw[fank] (7.45,0.0) .. controls (7.95,0.0) and (7.95,{0.085*\i}) .. (8.45,{0.085*\i});
\draw[arr] (8.5,0.0) -- (8.85,0.0);
\node[res, anchor=west] at (8.9,0.0)
  {$p(Y \mid X)$\\{\color{figorange}\itshape natural}\\{\color{figorange}\itshape conditional}};

\node[box, minimum width=2.1cm, minimum height=4.0cm, anchor=west] (score) at (11.65,1.6)
  {{\bfseries scoring}\\[1pt]\rule{1.8cm}{0.3pt}\\[3pt]
   excess Brier\\$(f-q)^2$\\[5pt]
   excess CRPS\\[5pt]
   excess bits\\{\color{figorange}\itshape tail} questions};
\draw[arr] (11.25,3.2) -- (score.west |- 0,3.2);
\draw[arr] (11.25,1.6) -- (score.west |- 0,1.6);
\draw[arr] (11.25,0.0) -- (score.west |- 0,0.0);

\draw[fpath] (gen.south) |- (9.5,-1.2) -- (score.south |- 0,-1.2) -- (score.south);
\node[capt, align=center, text=figgreen!75, anchor=north] at (9.5,-1.27)
  {model reads the report and the question, returns forecast $f$};
\end{tikzpicture}%

%% file: sections/04_micropolis.tex
\section{Micropolis: rare events in a city simulator}
\label{sec:micropolis}

To evaluate performance on tail events, we use Micropolis as our simulator. Micropolis is the open-source release of the classic 1989 SimCity, the original city-builder game. Its cellular-automaton-based engine simulates the evolution over time and space of crime, pollution, car traffic, population, and other variables. In the original game, a human player takes the role of Mayor and affects city development, but Micropolis can be run in a headless, no-Mayor mode. We use rollouts of these headless simulations, starting from the pre-built cities that ship with the game.

A game mechanic of interest is that of disasters. Some of these occur at fixed background rates, others with rates proportional to city characteristics, and others occur only when a variable crosses some threshold. They provide an interesting setting for studying forecasts of rare events.

\textbf{Setup.}
We ran 15 of the game's built-in cities from one seed and paused each at game years 20 and 30. For each of these 30 snapshots, models were asked to make forecasts with horizons of 3, 5, 7, and 10 game years. The world report describes the city state, includes time series of 6 metrics with 6-month resolution and an event log that includes all disasters that happened so far. At every snapshot and horizon we ask 25 binary questions, a total of 3,000 questions per model.  The analysis assigns each of these to a set by its realized $q$, as defined in \cref{sec:design-questions}: \MPDNQuestionsTail\ tail questions with $0<q\le 5\%$, scored in excess bits, and \MPDNQuestionsBinary\ mid-range questions with $5\%<q<95\%$, scored in excess Brier. The \MPDNExcludedZero\ questions with $q=0$ and the \MPDNExcludedTop\ with $q\ge 95\%$ are left out of both sets. The continuous set asks for five percentiles of 5 metrics at each horizon, \MPDNQuestionsContinuous\ questions per model. A snapshot's binary questions are split across two prompts of \MPDPerPromptBinary\ questions, and its continuous questions share one prompt of \MPDPerPromptContinuous\ questions. The prompt casts the model as a superforecaster and names the scoring rule (see \cref{app:worlds-micropolis} for details).

\subsection{Results}
\label{sec:micropolis-results}

\begin{figure}[t]
\centering
\includegraphics[width=0.78\linewidth]{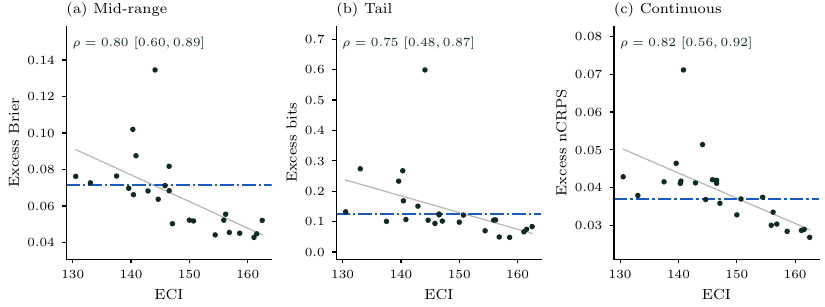}
\caption{Micropolis forecasting scores against the Epoch Capabilities Index for \MPDNModelsContinuous\ models; lower is better in every panel. (a) Mid-range questions ($0.05<q<0.95$): excess Brier, \MPDNQuestionsBinary\ questions. (b) Tail questions ($0<q\le 0.05$): excess bits, \MPDNQuestionsTail\ questions. (c) Continuous: excess nCRPS, \MPDNQuestionsContinuous\ questions. $\rho$ is sign-adjusted Spearman with its 95\% percentile bootstrap over models.
Blue dash-dotted: the baseline, the base rate in (a) and (b) and the climatological change in (c) (\cref{sec:design-baselines}).
}
\label{fig:micropolis-capability}
\end{figure}

Scores on the three question sets track ECI: $\rho$ is 0.80 for mid-range, 0.75 for tail and 0.82 for continuous (\cref{fig:micropolis-capability}). Per-model scores are in \cref{tab:micropolis-models}. For the binary questions, pooling all 3,000 (including $q = 0$ and $q \ge 95\%$) gives ECI correlations of 0.80 for excess bits and 0.82 for excess Brier.

\paragraph{Mid-range questions.}
Individual model biases run in both directions and 23 of 24 models beat a constant 0.5 forecast; \BLMPMidBeat{} beat the base rate (\cref{fig:micropolis-capability}a); the rest are among the less capable models. Mid-range loss follows \emph{discrimination}, the correlation between a model's forecasts and replay truth: across models the two correlate at $-0.88$, and discrimination rises with ECI ($\rho=0.77$).

\paragraph{Tail questions.}
Every model overestimates the probability of rare events (mean forecasts 0.05--0.30 against a mean $q$ of 0.015)
and none beats a constant 0.05 forecast, which knows the tail set's range of $q$; \BLMPTailBeat{} beat the base rate, which does not. More capable models overshoot less ($\rho=-0.68$ of mean forecast with ECI). The size of this bias
depends on elicitation details (\cref{app:ablation-batching}).

\paragraph{Continuous questions.}
Loss rises with horizon, from 0.028 excess nCRPS at 3 years to 0.050 at 10 (\cref{tab:micropolis-continuous}). The binary loss does not (\cref{fig:micropolis-horizon}), which reflects the question mix changing with horizon (\cref{app:micropolis-robustness}). \BLMPContBeat{} of 24 models beat the climatological change (\BLMPCont; \cref{fig:micropolis-capability}c), which scores close to the median model.

%% file: sections/05_freeciv.tex
\section{FreeCiv: natural conditionals in a strategy game}
\label{sec:freeciv}

\subsection{Worlds and question sets}
\label{sec:freeciv-setup}

To evaluate performance on natural conditional questions, we use FreeCiv as our simulator.  FreeCiv is an open-source, turn-based strategy game based on the Civilization series \citep{freeciv2026}, run through the CivRealm environment \citep{qi2024civrealm}. Each world involves five auto-controlled civilizations on a 78 by 52 map. We played eight games to turn 60, paused them, then replayed each game for 1,000 unique ``continuations'' to turn 210 (\cref{app:worlds-freeciv}). The world report, about 4,900 tokens, gives each civilization's state at turn 60 and at every fifth turn before it, the diplomatic state of every pair and a log of the events so far (\cref{app:prompts}). We draw the questions from \changed{23,241 generated candidates}: the mid-range set ($0.05<q<0.95$) takes 30 questions from each of five equal-width bands of $q$ at each horizon, balanced across question families, and the tail set ($0<q\le 0.05$) takes 60 per horizon (\cref{app:worlds-freeciv}). Baselines use all eligible candidates.

\subsection{Results}
\label{sec:freeciv-results}

\begin{figure}[t]
  \centering
  \includegraphics[width=0.8\linewidth]{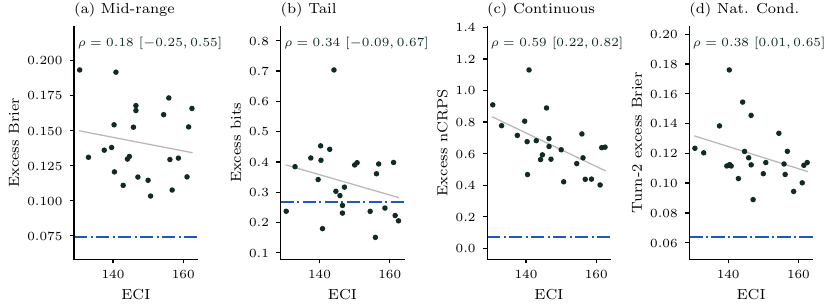}
  \caption{FreeCiv scores against the Epoch Capabilities Index for 24 models; lower is better in every panel. (a) Mid-range questions ($0.05<q<0.95$): excess Brier, 750 questions. (b) Tail questions ($0<q\le 0.05$): excess bits, 300 questions. (c) Continuous: excess nCRPS, 300 questions. (d) Natural conditionals: turn-2 excess Brier score against $p(Y \mid X)$, 400 cells. $\rho$ is sign-adjusted Spearman with its 95\% percentile bootstrap over models. Blue dash-dotted: the baseline, the base rate in (a) and (b), the climatological change in (c) and independence in (d) (\cref{sec:design-baselines}).}
  \label{fig:freeciv-capability}
\end{figure}

\paragraph{Mid-range questions.}
On the 750 mid-range questions, excess Brier averages \FCMidMean{} and correlates weakly with ECI ($\rho=\FCMidRho$; \cref{fig:freeciv-capability}a). Every model has negative mean forecast error on the mid-range set and loses to constant 0.5, a reference that benefits from the draw's even spread of $q$ (\cref{app:freeciv-weights}). \changed{No model beats the base rate either (\BLFCMid{}, against \BLFCMidBest{} for the best model; \cref{fig:freeciv-capability}a).} The forecasts barely follow $q$: their correlation with it is \BLFCMidCorrForecast{} across parsed forecasts and at most \BLFCMidCorrForecastMax{} for any one model, \changed{against \BLFCMidCorrRate{} for the base rate (\cref{app:baselines}).} Discrimination---the correlation between forecasts and replay truth---rises with ECI at $\rho=\FCMidDiscRho$. The truth-binned plot does not itself measure calibration (\cref{fig:freeciv-reliability}). On all 1,100 selected binary questions with equal item weights, nine models beat universal 0.5 (\cref{app:freeciv-baselines}); the per-band comparisons do not imply uniformly poor mixed-set performance.

\paragraph{Tail questions.}
On the 300 truth-selected tail questions, excess bits average \FCTailMean{} and correlate with ECI at $\rho=\FCTailRho$ (\cref{fig:freeciv-capability}b). Mean forecasts are \FCTailMeanFMin--\FCTailMeanFMax{} against mean replay probability \FCTailMeanQ. All models lose to constant 0.05 on this subset, a reference chosen with knowledge of its probability range. \changed{The base rate, which has no such knowledge, is beaten by \BLFCTailBeat{} models. They are the models that forecast lowest overall, which also costs them on mid-range questions:} models' mean tail and mid-range forecasts rank alike ($\rho=\BLFCRankCorrForecastLevel$). Joint binary comparisons depend on the question mixture (\cref{app:freeciv-baselines}).

\paragraph{Continuous questions.}
On the 300 continuous questions, excess nCRPS averages 0.65 and correlates with ECI at $\rho=0.59$ (\cref{fig:freeciv-capability}c), with the same sign at every horizon. Mean excess more than doubles from turn 90 to turn 210. The elicited 5th-to-95th percentile intervals contain 22\%--77\% of replay values. This undercoverage can reflect location error as well as insufficient width; it does not by itself identify the cause of the scores. \changed{No model beats the climatological change (\BLFCCont{} against \BLFCContBest{} for the best model; \cref{fig:freeciv-capability}c). Most quantities grow strongly after turn 60, and captures, razings and wonders had hardly occurred by then; other games show these changes more directly than the turn-60 report (\cref{app:baselines}).}

\paragraph{Natural conditionals.}
After the fact is revealed, turn-2 excess Brier averages 0.119 and correlates with ECI at $\rho=0.38$ (\cref{fig:freeciv-capability}d). Mean gain over keeping the turn-1 forecast is near zero ($-0.002$), but this masks substantial heterogeneity: mechanical same-series questions (C1) lose $0.053$ on average, while the other 344 cells gain $0.006$; 7 of 24 models have positive overall mean gain. However, models usually update in the right direction—on the 185 cells where the conditional-related shift is above 0.05, the update is directionally correct 86\% to 94\% of the time. The models' ultimate failure is in sizing the update—update sizes vary widely by question group, and do not point in a specific direction across the board (\cref{fig:freeciv-natcond}). \changed{No model beats independence, the base rate of $Y$ (\BLFCNatcond).} Ignoring the fact but using the true $p(Y)$ would score \BLFCNatcondTrueP{} (\cref{app:baselines}).

%% file: sections/06_starsim.tex
\section{Starsim: interventional conditionals in an epidemic model}
\label{sec:starsim}

To evaluate performance on interventional conditionals, 
we use Starsim as our simulator. Starsim is an agent-based disease model \citep{kerr2024starsim}. In this simulator, the world report specifies disease parameters, then asks models to forecast how a given vaccination program will change future infections (as compared to the counterfactual in which no vaccine is distributed).

\subsection{Worlds and question sets}
We use four worlds, each a region of 5,000 agents in Starsim 3.3.4 \citep{starsim2026software}. Worlds differ only in transmission probability, giving $R_0 \approx 1.8$, $2.4$, $3.0$ and $3.9$. For each world, we run 300 seeds and display the single run whose day-20 cumulative infections are closest to the median. The resulting world report gives the disease parameters, cumulative infections at five-day intervals through day 20, and the number infectious on day 20 (\cref{app:prompts-starsim}). (Full parameters are in \cref{app:worlds-starsim}).

We ask how many people will be newly infected between the end of day 20 and the end of day 40 or 60, excluding earlier infections that would dilute the vaccine's effect. Models return the 10th, 25th, 50th, 75th and 90th percentiles. In the interventional prompt, we  add one sentence: ``On day 21, a share of the population receives a vaccine that reduces infection probability per exposure by 95 percent.'' We vary the share of the population that receives the vaccine between 25\%, 50\%, and 90\%. In total, these conditions give 24 interventional items. To estimate the true probability, we perform 1,000 replays per item. Full elicitation details and exceptions are in \cref{app:worlds-starsim}; full scoring details are in \Cref{app:starsim-scoring}.

\begin{figure}[t]
\centering
\includegraphics[width=0.8\linewidth]{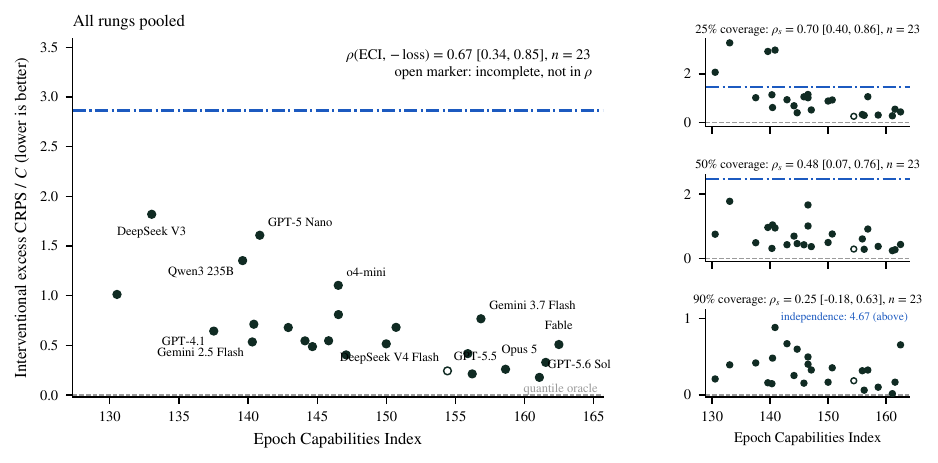}
\caption{Accuracy of forecasts under vaccination in Starsim versus the Epoch Capabilities Index. Scores are normalized excess CRPS averaged across both forecast horizons; lower is better. The main panel combines all three vaccination coverage levels, while the side panels show each separately. Saved forecasts are rescored using corrected replays from the same full simulation state at day 20. Each panel reports Spearman's $\rho_s=\rho(\mathrm{ECI},-\mathrm{loss})$ and a 95\% interval from 2,000 bootstrap resamples of models. Positive correlations indicate that more capable models make more accurate forecasts. DeepSeek V4 Flash (open marker) answered 21 of 24 questions and is excluded from all reported correlations. Blue dash-dotted: independence (\cref{sec:design-baselines}).}
\label{fig:starsim-interventional}
\end{figure}

\subsection{Results}

Models with higher ECI scores tend to forecast epidemic outcomes more accurately under vaccination (\cref{fig:starsim-interventional}). Across 23 models, the correlation between ECI and negative excess score is $\rho=0.67$ [$0.34,0.85$], where lower excess scores indicate better forecasts and the interval reflects uncertainty from resampling models. At 25\%, 50\%, and 90\% vaccination coverage, the correlations are $0.70$, $0.48$, and $0.25$, respectively, with mean normalized excess scores of 1.08, 0.69, and 0.34. All \BLSSIntBeat{} models beat independence (\BLSSInt; \cref{app:baselines}). Higher ECI also predicts better forecasts without vaccination: $\rho=0.75$ [$0.49,0.88$] across 24 models. \BLSSContBeat{} of 24 models beat the climatological change (\BLSSCont), but only \BLSSContSIRBeat{}, mostly among the most capable, beat an SIR forecast (\BLSSContSIR). This forecast integrates the report's SIR equations \citep{hethcote2000mathematics} from its day-20 state (\cref{app:baselines,fig:starsim-unconditional}). \Cref{tab:starsim-rescore-uncertainty} reports uncertainty from resampling models separately from uncertainty due to the finite number of simulation replays.

Forecasts under vaccination are less accurate at the longer horizon when scored in people rather than normalized units (\cref{fig:starsim-horizon}). Mean excess CRPS rises from 153 people at day 40 to 191 at day 60, with 19 of 23 models receiving worse scores. Among the 17 models shared between the two benchmarks, better performance here is associated with better overall ForecastBench performance ($\rho=0.45$ [$-0.14,0.88$]). However, the interval includes zero (\cref{tab:starsim-rescore-uncertainty}).

%% file: sections/07_real_world_forecasting.tex
\section{Relevance to real-world forecasting}
\label{sec:real-world-forecasting}

\begin{figure}[t]
  \centering
  \includegraphics[width=\linewidth]{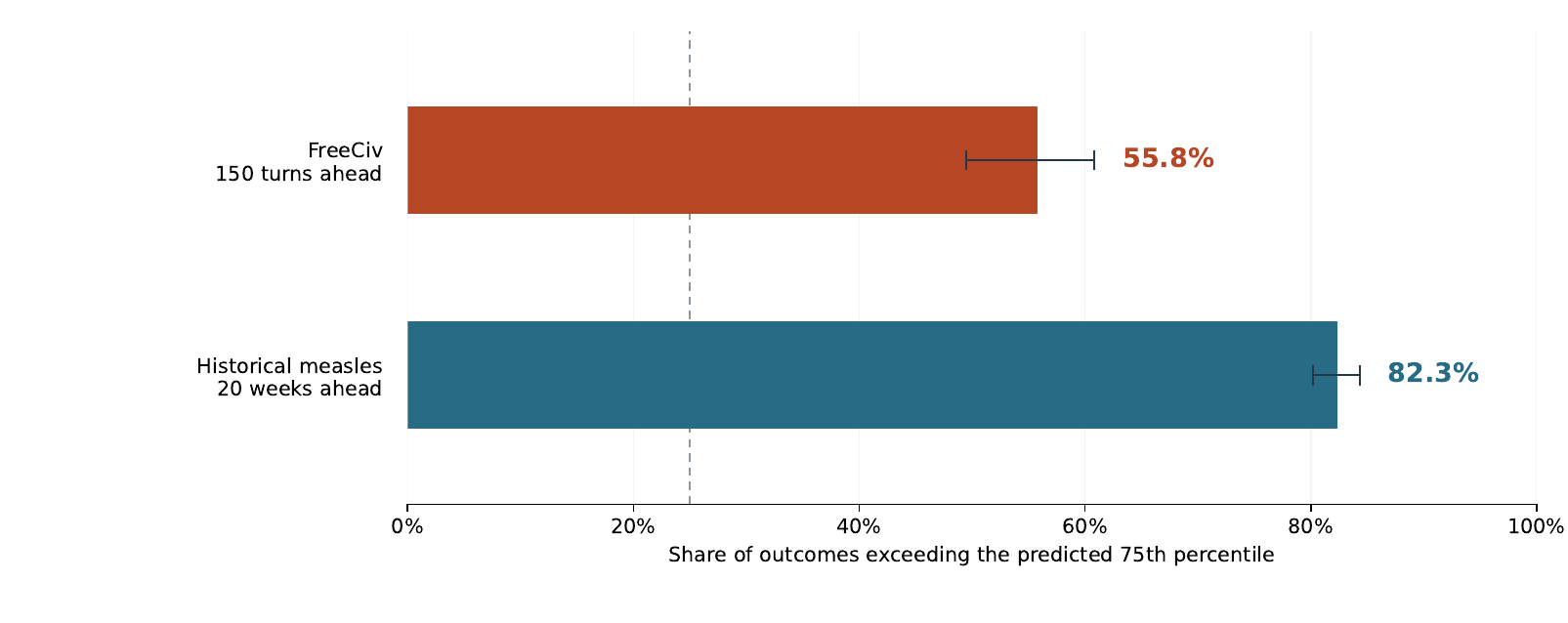}
  \caption{Identifying real-world forecasting errors using ForecastBench-Sim. In simulations, we observe that models set upper quantiles too low at longer timescales. We observe this same bias in historical measles cases.}
  \label{fig:calibration-case-study}
\end{figure}

In this section, we show that our benchmark can identify forecasting failures in simulated worlds that also appear in real-world forecasting. In our FreeCiv benchmark, we find that, at longer timescales, models set upper quantiles too low, such that outcomes exceed them much more often than their forecasts imply. Outcomes should exceed a calibrated 75th-percentile forecast at most 25\% of the time. Yet, across all continuous question families, at 150 turns ahead, the mean exceedance rate across 24 models is 55.8\%. We find that the same failure occurs in forecasts of historical US measles data\changednew{ \citep{tycho2018measles}}. In a cohort spanning season starts from 1928 to 1962 (1,339 state-seasons across 35 seasons), outcomes exceed the predicted 75th percentile 82.3\% of the time at 20 weeks, averaged across 28 models. All 28 models exceed the nominal 25\% rate (\cref{fig:calibration-case-study}). The aggregate error appears at every evaluated horizon in both domains, and persists on the smaller matched measles subset with explicit domain labels (\cref{app:real-world-forecasting}).

%% file: sections/08_related_work.tex
\section{Related work}
\label{sec:related}

\paragraph{Forecasting benchmarks.}
Benchmarks of LLM forecasting draw their questions from tournaments, markets and the news. ForecastQA and Autocast assembled resolved questions \citep{jin2021forecastqa,zou2022autocast}, ForecastBench generates new questions over time and scores models beside human forecasters \citep{karger2024forecastbench}, and Prophet Arena scores live market questions \citep{yang2026prophetarena}. Each resolves a question once, against a single realized outcome. Alternatives pose already-resolved questions with evidence frozen before resolution \citep{gao2025pastcasting,ye2026hindcast}, or check consistency without a truth value \citep{paleka2024consistency}. Moreover, live date-filtered retrieval can leak post-cutoff information \citep{ellahib2026temporalleakage}, and the related study did not test frozen snapshots. A 2026 survey \citep{xu2026llmforecastingagents} includes the FreeCiv-only design of \citet{lee2026forecastbenchsim}, which resolved each question against a single hidden continuation. A relevant human reference class comes from forecasting tournaments \citep{tetlock2005expert,mellers2014psychological}, which could not yet assess accuracy on low-probability questions about long-run risks \citep{karger2025xpt}. We retain the scoring rules of the forecasting literature but resolve against replays of a privately stored game state. This means we can use fresh question sets with unreleased answers, which reduces the likelihood of contamination.

\paragraph{Scoring rules and rare events.}
The uncertainty term of the Brier score \citep{murphy1973vector} depends partly on the probability of the event occurring, and \citet{paleka2025pitfalls} show that averaging raw scores across base rates hides rare-event skill. \citet{lerch2017forecasterdilemma} also showed that scoring only the cases in which an extreme event occurred rewards forecasters who overstate its probability, so we avoid this failure mode by selecting tail questions on $q$ before any specific outcome is drawn. Because calibration on overall questions does not translate well to calibration on rare events \citep{allen2024tailcalibration,merkle2013choosing}, we also score tail questions in excess bits.

\paragraph{Causal, conditional and updating evaluation.}
Causal benchmarks for LLMs take their truth from a symbolic oracle over a small causal graph \citep{jin2023cladder,jin2024corr2cause,kasetty2024evaluating} or let an agent intervene in a small generative model \citep{gandhi2025boxinggym,yang2026causalab}. InterveneBench evaluates causal study designs and intervention reasoning against expert-verified annotations \citep{shi2026intervenebench}, without scoring predicted outcome distributions. \citet{tetlock2017bringing} argued that policy conditionals are most critical to decision makers, but that they cannot be resolved in the real world. IARPA's FOCUS program proposed in 2017 to score counterfactual forecasts by rerunning a simulated world, with forecasters reading reports of AI-played Civilization games \citep{lehner2017focus,wiblin2019tetlock80k}. An interventional conditional follows the same idea with a proper rule, paired replays of a disease model and a language model as the forecaster.
 EvolveCast updates LLM forecasts on real news and can compare each update only with those of human forecasters \citep{yuan2025evolvecast}. Other work tests whether updates are consistent with Bayes' theorem without an event-level truth \citep{imran2025bayesian,he2025martingale}, and BayesBench scores beliefs against exact posteriors under a specified likelihood \citep{samanta2026bayesbench}. Binary conditional forecasting is established \citep{mccracken2022binaryconditional}; here we evaluate language-model updates after revealed facts against conditional probabilities estimated by replay.

%% file: sections/09_discussion.tex
\section{Discussion}
\label{sec:discussion}

Compared to existing forecasting benchmarks, ForecastBench-Sim allows us to examine forecasting errors against estimated outcome distributions by replaying the same simulated state. We find, for example, that models set upper quantiles too low in FreeCiv, a pattern we also find in historical measles forecasts (\cref{sec:real-world-forecasting}). The fact that a shared failure mode exists makes ForecastBench-Sim a promising setting for reinforcement learning with verifiable rewards (RLVR): fresh worlds can supply large numbers of forecasting questions, including rare events and conditional questions, with rewards computed against replay-based outcome distributions, without waiting for real-world events to resolve or being limited to a historical record. Whether improvements learned in simulation transfer to real-world forecasting remains to be tested.

One limitation of our evaluation is that replay distributions are conditioned on the full simulator state, whereas models receive only a partial textual report. Some excess loss may therefore reflect an unavoidable information gap rather than forecasting error. (The Starsim SIR baseline helps interpret this gap: it uses only information supplied in the report and outperforms most evaluated models, demonstrating that better forecasts were possible with the available information (\cref{sec:starsim,app:baselines}).) Future work could supply full simulator states, potentially via an agentic harness. Alternatively, future work could systematically vary models' access to information, measuring how observability affects forecasting performance.

Further, our simulators, however complex, are necessarily simpler than the real world, and the specific simulators we chose are likely in models' corpora. Fresh worlds and varied configurations could reduce contamination risk, while withholding selected simulator rules would test whether models can infer how unfamiliar worlds work. While this variation does not eliminate the risk of gaming, it will allow us to at minimum test whether models adapt to new conditions, or whether they reproduce patterns from our published work.

\section{Conclusion}
\label{sec:conclusion}

ForecastBench-Sim uses simulated worlds to evaluate models' forecasting ability against estimated outcome distributions. Across three simulated worlds, more capable models (as measured by ECI) generally score better on our benchmark (\cref{tab:validation}). These skill rankings also agree with two real-world forecasting leaderboards (FutureEval, $\rho=\ExtFETheta$; Prophet Arena, $\ExtPATheta$), although agreement with a third is weaker (ForecastBench, $\rho=\ExtFBTheta$). After adjusting for ECI, the partial-correlation estimates are $\ExtFEThetaGivenEci$, $\ExtPAThetaGivenEci$ and $\ExtFBThetaGivenEci$, respectively; all three 95\% intervals include zero (\cref{app:validation-forecastbench}). These results are consistent with ForecastBench-Sim capturing forecasting ability beyond general model capability. We also identify the same tendency to set upper quantiles too low in FreeCiv and historical measles forecasts (\cref{sec:real-world-forecasting}). Future work may use our benchmark as a controlled setting to study and train LLM forecasters alongside real-world evaluation.

%% file: sections/10_ai_use.tex
We used language-model assistants for drafting and literature retrieval in the paper-writing process, as well as to produce statistical summaries and figures. In the experimental setting, models conducted code generation and simulation and elicitation monitoring. Models also helped construct the Micropolis knowledge test (\cref{app:ablation-knowledge}). Models also inspected prior notes about the research setup, audited scoring and provenance to ensure the paper's numbers were attributed correctly, and implemented the retrospective Starsim continuous repair. The authors are responsible for the final analyses, citations and claims.

%% file: sections/11_reproducibility.tex
Benchmark code:\\
\url{https://github.com/forecastingresearch/forecastbench-sim/tree/arxiv}\\
Analysis code:\\ 
\url{https://github.com/forecastingresearch/forecastbench-sim-paper/tree/arxiv}\\
Data release:\\
\url{https://zenodo.org/records/24401629}

The data release holds the raw model responses, question sets, world reports, full prompts and the other data files not already included in these repositories; \Cref{app:release} describes its contents.

\Cref{sec:design-scoring} defines all the relevant scoring rules used in the paper. \Cref{app:worlds} describes each simulation, as well as listing the tested models with details on reasoning settings (and cost). \Cref{app:prompts} contains reproductions of the prompts, question templates in full, and excerpts of the world reports. \Cref{app:full-results} reports the scores included in every figure, split by model and horizon. \Cref{app:release} indicates which scripts score the outputs and produce each figure.

%% file: appendix/A_worlds_and_protocol.tex
\section{Simulations and evaluation protocol}
\label{app:worlds}

Replay truths condition on the full saved simulator state, while forecasters see a report; zero excess is therefore an oracle reference, not necessarily attainable from the report alone. \Cref{sec:design} defines ground-truth probability $q$ and the scoring terms; \cref{app:prompts} gives the forecasting prompts and templates. Throughout, a horizon $T$ is the turn or day at which a question resolves, and ``the window'' means the turns after the snapshot up to and including $T$.

\subsection{Micropolis}
\label{app:worlds-micropolis}

\textbf{Program.}
Micropolis is the open-source release of the SimCity Classic engine \citep{micropolis2008}. We use MicropolisCore, its C++ and WebAssembly port, through the port's command-line runner. The runner loads a city from a built-in save and advances the engine with no player input and with random disasters enabled. One turn is 16 engine ticks, and 48 turns make one game year. Our version of the engine fixes an out-of-bounds crash in the engine and a reseeding of its random number generator from the clock on every load; with the fixes, the city, the seed and the number of turns determine a run completely. Transport, police and fire funding are fixed at 100\%, and the treasury has no effect on the city. We therefore ask no question about money, and the world report withholds the city's funds. The benchmark repository vendors this engine; at commit \texttt{e0f49b5} it reproduces the trunk runs and every replay of the paper (\cref{app:release-contents}).

\textbf{Cities, snapshots and replays.}
We use 15 of the 19 populated built-in cities (\cref{tab:micropolis-cities}). We run each city from seed 42 and pause it at turns 960 and 1,440, game years 20 and 30, which gives 30 snapshots. From each snapshot we run 1,000 replays. The runner takes a byte copy of the engine's memory at the snapshot, restores it for every replay, reseeds the random number generator and runs to the requested horizon. A replay's output depends only on the city, the seed used to reach the snapshot, the snapshot turn, the horizon and the replay seed. The horizons are 144, 240, 336 and 480 turns past the snapshot, or 3, 5, 7 and 10 game years. One replay of 480 turns takes about 110~ms on one core, so 1,000 replays of one snapshot take under two minutes. The world report for a snapshot gives the current state, the reported metrics every 24 turns since the start, the counts of rubble, burning and road tiles, and the disasters and advisories so far (\cref{app:prompts-micropolis} reproduces one).

\textbf{Binary questions.}
\Cref{app:prompts-micropolis} lists the 25 questions we ask at every snapshot and horizon, as they appear in the prompt. Every question concerns the window from the snapshot turn $S$, exclusive, to the horizon turn $H$, inclusive. Event questions count the engine's own messages in the window. One plane crash emits both a plane and a helicopter message, and we count only the first. Population, class, score, pollution and tile counts are read from the engine's per-turn statistics at the end of the named turn, and a yearly checkpoint is a turn that is a multiple of 48. The city classes, in increasing order, are Village, Town, City, Capital, Metropolis and Megalopolis. B8 resolves YES if the largest population at a yearly checkpoint in the window exceeds the largest at any yearly checkpoint up to $S$. The analysis then assigns the 3,000 question instances per model (25 questions, 30 snapshots, 4 horizons) to sets by the realized $q$: \MPDNQuestionsTail\ tail instances with $0<q\le 0.05$ and \MPDNQuestionsBinary\ mid-range instances with $0.05<q<0.95$ per model. The \MPDNExcludedZero\ instances with $q=0$ and the \MPDNExcludedTop\ with $q\ge 0.95$ are in neither set; they enter only the check in \cref{sec:micropolis-results} that pools all 3,000 under one scoring rule.

\textbf{Continuous questions.}
The continuous set asks for the value at the horizon of population, average traffic, average pollution, average crime and average land value. With five metrics, four horizons and 30 snapshots, that is \MPDNQuestionsContinuous\ questions per model. The model returns the 10th, 25th, 50th, 75th and 90th percentiles. The score is the excess nCRPS of \cref{sec:design-scoring}: the CRPS of those five percentiles against the replay distribution, less the floor the replays themselves score, divided by a scale computed per city and metric --- that metric's mean over the city's turns up to its first snapshot, floored at 10,000 for population, 10 for traffic and 40 for pollution, crime and land value.

\textbf{Elicitation.}
A snapshot's binary questions are split across two prompts of \MPDPerPromptBinary\ numbered questions, ordered by horizon; its continuous questions share one prompt of \MPDPerPromptContinuous\ (5 metrics at 4 horizons). The prompt opens ``You are an expert superforecaster, familiar with the work of Tetlock and others'', describes the no-mayor setting, and names the scoring rule: the Brier score for the binary set and the CRPS for the continuous set. It ends with a fixed answer block, one line per question. The full prompt details are given in \cref{app:prompts-micropolis}. We send each prompt once. Reasoning requests and uncontrolled-model exceptions are listed in \cref{app:roster}.

\begin{table}[htbp]
  \caption{The 15 Micropolis cities and their population when loaded. Each city is paused at turns 960 and 1,440 and replayed 1,000 times from each snapshot.}
  \label{tab:micropolis-cities}
  \centering
  \small

  \begin{tabular}{@{}lr@{\hspace{2em}}lr@{\hspace{2em}}lr@{}}
    \toprule
    City & Population & City & Population & City & Population \\
    \midrule
    haight & 214,260 & kamakura & 90,380 & senri & 20,700 \\
    joffburg & 178,680 & yokohama & 81,380 & wetcity & 19,960 \\
    kobe & 153,120 & freds & 63,440 & southpac & 15,240 \\
    kowloon & 137,640 & bruce & 14,440 & linecity & 12,780 \\
    kyoto & 108,820 & happisle & 105,420 & deadwood & 6,580 \\
    \bottomrule
  \end{tabular}
\end{table}

\paragraph{Scale and selection scope.}
All Micropolis prompt variants use the same city-and-metric normalization table, computed from history through the main run's first snapshot at year 20, with the stated floors. The scale is not recomputed for each variant's age. Binary prompts name Brier scoring while headline tail evaluation uses excess log loss; both are proper for truthful probabilities, but we have not tested whether models respond the same way when the named rule changes.

\paragraph{Continuous-answer range check.}
For all Micropolis continuous results, we mark a set of percentiles as unparsed if its median is outside the engine's possible range: 0--255 for average pollution, crime and land value, 0--612 for average traffic, or negative for population. This post-hoc rule removed no answers from the main run; its motivation and effect on the prompt variants are reported in \cref{app:ablation-variants}.

\subsection{FreeCiv}
\label{app:worlds-freeciv}

\textbf{Program.}
FreeCiv is an open-source, turn-based strategy game \citep{freeciv2026}. We run its server inside the Freeciv-web container \citep{freecivweb2026} through the CivRealm environment \citep{qi2024civrealm}. Each game has five civilizations, created by the server's fill option and played by the game's AI at the ``hard'' level. The one connected player is switched to AI control at the same level and serves only as the observer through which the environment reads the game; no human or model plays. The map is 78 by 52 tiles. Barbarians run at the server's default settings, under which raids begin around turn 60. The container is \texttt{civrealm/freeciv-web:1.4} (January 16, 2024); its server reports version 3.2.91-dev, a Freeciv 3.3 development build, and every game uses the \texttt{classic} ruleset, as the headers of the saved games record.

\textbf{Anchor games and replays.}
We played eight anchor games, seeds 7001, 7003, 7005, 7008, 7010, 7011, 7014 and 7022, to turn 60 and kept the server's saved game at every turn. From the turn-60 save of each anchor we ran 1,000 replays, seeds 5001 to 6000, to turn 210. The replays ran on a fleet of 27 cloud workers on September 2, 2026, 600 at a time, and finished in about three hours. We read every quantity used for resolution from the per-turn saved games of the replay, not from the client. A cross-check of 6,080,000 (turn, civilization) cells between the serialized series and an independent parse of the saved games found no mismatch. Five replays stopped at a fixed turn with every civilization alive and did so again when rerun. They are program failures, and we replaced each with a fresh seed on the same anchor (7001: seeds 5014 and 5841; 7005: 5439 and 5639; 7022: 5823; substitutes 6001 upward). We also replaced two replays in which the observer civilization was eliminated before turn 210 (7001 seed 5482 at turn 117; 7014 seed 5290 at turn 178), so that every replay reaches turn 210. That replacement is conditioned on an outcome. Replacing one of 1,000 replays changes an unconditional frequency by at most 0.001. Conditional frequencies use smaller reveal-specific denominators: with 100 qualifying replays, one replacement can change the conditional frequency by 0.01. These are bounds, not measured changes; the two originals are retained. A ninth anchor, seed 7016, lost its observer in 3.1\% of replays, and we replaced it by seed 7011 before the draw. Replacing an anchor when its observer dies also makes anchor selection depend on survival. Changing one replay outcome can affect eligibility at the $q>0$ tail boundary; its actual effect on selected items or scores has not been established. The corpus occupies 92~GB.

\textbf{World report.}
We generate the report from the turn-60 saved game; it is 17.9 to 20.5~KB, about 4,900 tokens, per anchor. It lists the civilizations and, for each, the score, rank, government, treasury, population, technologies, wonders, cities, territory, units, military, science, trade, food and production at turn 60. It gives the diplomatic state, first-contact turn and embassy status of every pair, whether each pair's territories touch, and each civilization's government changes. It samples the same quantities every five turns from turn 1. It ends with a log of every city founded, captured or destroyed, every diplomatic change, and every technology and wonder completed through turn 60. It contains no map and no research target. A forecaster therefore sees less than the saved game holds, and the ground truth is conditional on the full saved game.

\textbf{Question sets and the draw.}
We generate a large set of candidate questions from which we select a subset (the \emph{draw}). For the mid-range questions, the draw is made to have an approximately uniform distribution of  ground truth probabilities $q$.
Every question is an instance of a family with parameters: a civilization, a pair, a quantity, a threshold, a count, a wonder, a technology or a government. \Cref{tab:app-freeciv-binary-families} lists the 23 binary families. \Cref{tab:app-freeciv-continuous-families} lists the nine continuous families with their normalization constants, and \cref{tab:freeciv-anchors} the items per anchor. We generate candidate questions for every family, civilization, pair and parameter value at each anchor game and horizon; questions about whether a quantity reaches a level use seven levels, from the 5th to the 95th percentile of the quantity across the replays. Dropping candidates with $q=0$ or $q=1$, duplicates, and questions about a civilization eliminated by turn $T$ in any replay (survival and civil-war questions excepted) leaves \changed{23,241 binary candidates}. Within each quota the draw is random but rotates through families, so every family with candidates gets an equal share before any gets more. Caps limit the questions per anchor game and per family, and allow at most one question of a family about the same civilization or pair at a horizon. The mid-range set holds 750 items: 30 from each of five equal-width bands of $q$ on $(0.05, 0.95)$ at each of the five horizons, balanced across families within a band. Its mean $q$ is \FCMidMeanQ. The tail set holds 300 items with $0<q\le 0.05$, 60 per horizon; the 10th, 50th and 90th percentiles of $q$ are 0.001, 0.013 and 0.048, and $q=0$ is excluded by construction. The mirror set holds 50 items with $q\ge 0.95$, 10 per horizon. The continuous set holds 300 items, 60 per horizon: eight count families at about 6 per horizon each, and value-at-$T$ questions on score, population, cities and territory at 3 per metric per horizon. We draw them at random from the \changed{1,462 of 1,502 generated candidates} whose replay distribution at the forecast turn has a positive median and a positive interquartile range. At most two come from each game for each family (or metric) and horizon, with no repeated combination of game, subject, family and horizon. Each continuous item carries its 1,000 replay values, and the model returns the 5th, 25th, 50th, 75th and 95th percentiles. The normalization constant of a family is the median over its items of the 5th-to-95th percentile range of the truth, rounded to one significant figure. The set includes 16 survival questions about elimination, so there is no blanket exclusion of all questions involving elimination. The question generator checks survival in every replay for the listed subject civilizations and designated world-state quantities. Survival and civil-war events are explicit exceptions, and the checks do not cover every world aggregate.

The natural-conditional set holds 400 cells at horizons 120 to 210, 100 per horizon, over 355 distinct questions. A cell pairs a question $Y$ with one fact $X$ about turns 61 to 90; its target $p(Y\mid X)$ is the frequency over the replays in which $X$ held. A cell is admissible when at least 100 replays satisfy $X$, $0.02<p(Y\mid X)<0.98$, at most two facts are paired with one question, and at most eight cells share one event. We drew cells in five structural groups, with 36, 36, 56, 172 and 100 cells: A, irrelevant news; B, related weak news; C1, mechanical same-series; C2, persistence across series; D, inferential. In group A the fact concerns a civilization or pair that the question is not about. In group B the fact is a related but weak event on the state the question asks about: a capture, a government change, a wonder, a cease-fire or a first contact. In group C1 the fact is a change in the same quantity that the question asks about, for instance that a civilization lost three cities when the question asks about its city count at the horizon. In group C2 the fact is the same kind of change on a different series, for instance the same loss of cities when the question compares scores or territory. In group D the fact bears on the outcome only through inference, for instance a war that began when the question asks about war, alliance, civil war or conquest at the horizon. Within each group we drew cells with the strongest kinds of fact first (a loss of three or more cities, a fall of two or more in city count, a war begun), then weaker kinds. The revealed facts are of 14 kinds; the most common are that a civilization lost three cities (191 cells) and that a pair went to war (90 cells). Measured after the draw over all 1,000 replays, 174 cells have an estimated replay shift $|p(Y\mid X)-p(Y)|$ below 0.03, 96 between 0.03 and 0.08, 67 between 0.08 and 0.15, and 63 of at least 0.15. A further 124 turn-1 questions exist only to seed cells and enter no other set. The turn-2 prompt adds one sentence, ``One fact about turns 61 through 90 has been revealed to you: \ldots'', and never discloses the turn of the event.

\textbf{Elicitation.}
We send every prompt through OpenRouter with the report before the questions. On first-party hosts the prompt has two content parts with a cache breakpoint between them; elsewhere it is one string. There is no system message and no persona. The prompt says that answers will be evaluated with a proper scoring rule, without naming the rule. A grouped prompt of the run the paper reports has 8,759 to 17,575 tokens and asks for one line per question in a delimited block, a probability or the five percentiles; a natural-conditional prompt has 7,565 to 8,544 tokens and asks for the probability inside a tag (\cref{app:prompts}). Every request uses provider-default sampling and an output limit of 32,768 tokens (16,384 in the run of September 9, with 24,000 for DeepSeek V4 Flash and 6,000 for Kimi K2). We set reasoning effort to the lowest level each model supports, with a 1,024-token budget for four models and no setting for four others (\cref{tab:roster}), and pin every model to one host (\cref{app:roster}). Every stored row records the host, the prompt, completion and reasoning token counts, the cost, the finish reason, the reasoning parameter sent and, where returned, the reasoning text. We read a grouped reply from its last delimited block, matching each line to its question by the stated number, and fall back to numbered lines outside the block or, for percentiles, to five numbers on the line after a restated question. We read a natural-conditional reply from its tag block or, failing that, from a labeled line, which we flag, and ask once more when neither can be read. A missing binary answer scores as a forecast of 0.5; missing continuous and conditional answers are dropped. The no-news control continues the turn-1 exchange of 100 cells, stratified by group, with a sentence that says the game has continued and no new information is available.

The run of September 9, 2026, asked one question per prompt, 2,023 calls per model (1,524 turn-1, 400 turn-2 and 99 no-news), and produced 48,543 of 48,552 expected rows. Twenty-three models finished in 41 minutes; DeepSeek V3 took 85 minutes because its host limited it to about 8 calls per minute. Every reply from twenty-one models could be parsed. Kimi K2 (26 rows) and DeepSeek V4 Flash (27 rows) ran to the output limit with no answer, DeepSeek V3 lacks 9 turn-2 rows whose turn-1 answer was empty, and GPT-5 Nano at minimal effort needed the lenient parse on about 12\% of its calls. DeepSeek V4 Flash ignores the effort setting and returns 10,000 to 15,000 reasoning tokens per call. The scored rows cost \$393.83 across the 24 models, and a check at medium effort on 7 models and 200 mid-range items a further \$34 (\cref{app:ablation-effort}). The paper uses this run's natural-conditional forecasts for 22 models and compares its other sets with the grouped run in \cref{app:ablation-grouping}.

The run of September 20, 2026, is the one the paper reports. It asked the mid-range set, tail, mirror and continuous questions in groups, with the limits of the Micropolis run: 26 binary prompts of 36 to 50 questions and 18 continuous prompts of 13 to 20, one game per prompt and 44 prompts per model. We sent each prompt once and did not ask again for an answer we could not read. Every model answered at least 98.8 percent of its questions, and 99.9 percent on average. One prompt returned no readable answer at all (o4-mini). The natural-conditional arm kept the protocol of September 9, one question per prompt. We reuse its forecasts for the 22 models whose host did not change. DeepSeek V3 and DeepSeek V4 Flash moved to the hosts that Micropolis used and answered them again. DeepSeek V4 Flash ran at effort low, as in the other two simulations; the setting has no effect on it, and it returned about 13,275 reasoning tokens per grouped call. The grouped calls cost \$26.10 and the two natural-conditional reruns \$2.93.

\begin{table}[htbp]
  \caption{FreeCiv items per anchor game (draw v1.8). The sets are stratified on $q$, horizon and family, not on anchor, so anchors contribute unequally. Extra turn-1 questions seed natural-conditional cells and enter no set.}
  \label{tab:freeciv-anchors}
  \centering
  \small
  \setlength{\tabcolsep}{4pt}
  \input{data/freeciv_items_per_anchor}
\end{table}

\begin{table}[htbp]
  \caption{FreeCiv binary question families (draw v1.8). Each rule paraphrases the generated resolution criteria; the window is turns 61 through $T$ inclusive, and ``at $T$'' means as recorded at the end of turn $T$. A capture is a city changing owner by force, counted per event; a city lost in a civil-war split is not a capture. Great wonders exclude small wonders such as a palace. Twenty-two families appear in the mid-range set; \texttt{NW4\_survival} appears only in the tail set. Columns give items in the mid-range set (750), the tail set (300), the mirror set (50) and the natural-conditional cells (400). \Cref{tab:freeciv-binary-families} gives an example question from each family.}
  \label{tab:app-freeciv-binary-families}
  \centering
  \footnotesize
  \setlength{\tabcolsep}{4pt}
  \input{data/freeciv_binary_families}
\end{table}

\begin{table}[htbp]
  \caption{FreeCiv continuous question families (draw v1.8) with items and the constant that divides the CRPS of each family: the median over the family's items of the 5th-to-95th percentile range of the truth, rounded to one significant figure. \texttt{NC1\_value\_at\_T} has one constant per metric.}
  \label{tab:app-freeciv-continuous-families}
  \centering
  \footnotesize
  \setlength{\tabcolsep}{4pt}
  \input{data/freeciv_continuous_families}
\end{table}

\paragraph{Saved-answer coverage.}
Micropolis drops unparsed forecasts: 71,951 of 72,000 binary and 14,319 of 14,400 continuous answers are scored. Its exported continuous forecast file contains derived scores, not the original five quantiles. FreeCiv retains its distinct policy: one missing mid-range forecast among 18,000 model--question rows is imputed to 0.5; all 7,200 tail and 1,200 mirror rows parsed. Of 7,200 continuous rows, 37 are dropped, and 16 of 9,600 natural-conditional cells lack paired forecasts; all 2,976 extra turn-1 rows parsed. These counts do not account for all failed API attempts. The Micropolis binary missing-answer sensitivity is reported in \cref{app:micropolis-robustness}; continuous and FreeCiv missing-not-at-random sensitivities and complete raw-cache provenance remain unresolved.

\subsection{Starsim}
\label{app:worlds-starsim}

\textbf{Program.}
Starsim is a framework for models of disease spread among individuals \citep{kerr2024starsim,starsim2026software}. Both repositories fix the Starsim version at 3.3.4. The simulation package (scenarios, report generator and paired runs) is \texttt{pandemic\_world} in the ForecastBench-Sim repository at commit \texttt{65204aa} of August 20, 2026, which corrected the vaccination day (an earlier version started the vaccination on day 0; no reported run used it). The code that built the worlds, elicited the models and scored them is the \texttt{results/causal} part of the paper repository; the September 8 run is its commit \texttt{4d3a27a}, and the September 16 run adds the per-model reasoning setting and the host of each call to the same code.

\textbf{World.}
\Cref{tab:starsim-parameters} lists the parameters of the run. A region is 5,000 individuals on a random contact network of about six contacts per person per day under a susceptible--infectious--recovered model with a ten-day mean infectious period, no second infection once a person has recovered, and no deaths; 0.5 percent of individuals are infectious on day 0. Four worlds differ only in the per-contact transmission probability. For each we ran 300 seeds and displayed the run whose cumulative count at day 20 is closest to the median (seeds 27, 33, 62 and 47 for $\beta = 0.03$, $0.04$, $0.05$ and $0.065$; 431, 1,129, 2,316 and 3,994 people ever infected at day 20). The original ground-truth construction pooled seeds whose day-20 count lay within 15 percent of the displayed run's: 131, 157, 202 and 290 of the 300, with vaccination runs using the same seeds. This historical construction is retained for binary scores only; continuous scores now use the retrospective fixed-state procedure below. The two-region worlds of the binary sets pair one displayed run with another at a value of $\beta$ one step apart (Riverton at 0.04 with Southbay at 0.03, 0.05 with 0.04, 0.05 with 0.065, and 0.03 with 0.04); the fourth is the reverse of the first, so two worlds have Riverton ahead and two Southbay.

\textbf{Items.}
The continuous template, \texttt{new\_infections}, asks how many people will be newly infected between the end of day 20 and the end of day $t$, $t \in \{40, 60\}$, and is answered with the 10th, 25th, 50th, 75th and 90th percentiles; the revised truth is the distribution over 1,000 continuations from the fixed displayed state in the specified arm. The binary template, \texttt{cases\_comparative}, asks whether Riverton will have more cumulative infections than Southbay at day $t$, and its ground truth is $q=(n_Rn_S)^{-1}\sum_{i,j}\mathbf{1}(R_i>S_j)$ over the full Cartesian product of the matched regional pools; ties are NO. This is not equal-seed pairing. Within a region, control and vaccination reuse seeds and pre-day-21 histories. Across regions, seed IDs also reuse random roots, so the marginal samples are not independently seeded. The unconditional and 90 percent probabilities lie at the endpoints of the empirical distribution. Bootstrap intervals that collapse to a point do not establish certainty about the underlying population. No question about deaths was run: the model has no deaths. The five sets are the 8 unconditional continuous items, the same 8 under a vaccination at 25, 50 and 90 percent coverage (24 items), the 8 unconditional binary items, the same 8 under a vaccination at 90 percent, and the 8 of the \emph{hold} set, in which the vaccinated region is the one ahead and the coverage is the largest value, in steps of 5 percent, that leaves the ranking the same at both horizons (20, 10, 15 and 20 percent by world; the day-60 ground-truth values are then 0.97, 0.98, 0.01 and 0.03). \Cref{tab:starsim-templates} gives the templates and the intervention sentence.

\textbf{Elicitation and records.}
The four binary worlds use 20,567, 31,714, 58,580 and 20,567 Cartesian comparisons, respectively, but these are dependent comparisons of reused regional trajectories, not independent joint trials. There are 104, 145, 202 and 104 seed IDs that are eligible in both regions. Restricting the comparison to those pairs changes how the regional outcomes are coupled; it does not repair the original target. Matching uses only day-20 cumulative counts within $\pm15\%$, ignoring other displayed history and infectious counts. Requiring an exact match to the full report leaves only the displayed run in each cache. The same cached outcomes were used to select ranking flips and hold coverages, further limiting generalization.

Each prompt contains the provenance statement, the report, an intervention sentence where applicable, the question and the answer format, reproduced in full in \cref{app:prompts}; one question per prompt, no role, no scoring rule named. Calls go through OpenRouter at default sampling settings, with the output limit at the larger of 16,000 tokens and the model's listed limit (65,536 at most), a 15-minute time limit, and up to four attempts when a response reaches the output limit or contains no answer that can be parsed. Five binary and three continuous repetitions were planned; medians over available saved answers are scored. The September 16 run sets each model's reasoning to its lowest level, with the 1,024-token budget for the four budget-only models (\cref{app:roster}); the September 8 run had set no reasoning parameter. The record of every call holds the response, the finish reason, the prompt, completion and reasoning token counts, the cost, the response time, the reasoning setting, and the host that served it. The September 16 records contain 5,045 retained scored responses at a recorded cost of \$41, of which the unconditional binary, unconditional continuous and continuous interventional sets are 3,190 calls and \$27; DeepSeek V4 Flash retained 111 of 216 planned responses across 49 of 56 items, including 21 of 24 continuous intervention items and 4 of 8 hold items; Kimi K2 retained 182 of 216 across all 56 items. The other 22 models have all planned responses. These are successful retained responses, not a verified count of all API attempts. The scorer treats a set as complete when every item has an answer, even if some items have fewer repetitions than planned.

\textbf{Protocol details.}
The unconditional and interventional versions of an item are sent in separate contexts. We do not batch questions: each set has only eight items, and showing both horizons together would reveal that cumulative counts cannot decrease. Eight models returned no reasoning tokens at the lowest setting. DeepSeek V4 Flash returned about 18,000 per recorded call despite that setting; its incomplete continuous intervention and hold sets are excluded from full-set correlations; its complete eight-item binary 90 percent set is included. Kimi K2 returned no answer to many calls on the run day and completed every item on later attempts, some with fewer than three answers.

\textbf{Scoring and normalization.}
\label{app:starsim-scoring}
Binary items are scored in \emph{excess bits}, $\mathrm{KL}(P^* \,\|\, f)$ with $f$ bounded to $[0.001, 0.999]$, the rule that FreeCiv applies to its tail questions (\cref{sec:design-scoring}), here on both sides of one half: every $P^*$ in the unconditional and 90 percent sets is exactly 0 or 1, and on the hold set the day-60 ground-truth values are 0.97, 0.98, 0.01 and 0.03, so no binary item falls in the mid-range where the excess Brier score $(f - P^*)^2$ applies. Continuous items are scored by a five-quantile pinball approximation to the CRPS \citep{matheson1976scoring}, computed as in \cref{sec:design-scoring} and averaged over the 1,000 fixed-state continuation outcomes. The \emph{floor} is the score of the ground-truth distribution's own quantiles at the same five levels, computed the same way; it is the minimum five-quantile score under the empirical replay distribution, and the excess CRPS is the score minus the floor. To put the two horizons on one scale we divide the excess by one constant per horizon, $C_{40} = 200$ and $C_{60} = 300$ people. These historical reference scales were the median 5th-to-95th percentile ranges under the original matched-seed truths, rounded to one significant figure; they are retained unchanged to make the retrospective comparison interpretable, not reestimated from the new distributions. The elicited quantiles stay at the 10th to 90th percentiles, where FreeCiv elicits the 5th to 95th (\cref{app:worlds}). We compute the excess per item and then average over the items. Scores assess outcome distributions under the intervention. They combine baseline prediction and intervention response and do not isolate forecasting of the effect itself.

\paragraph{Retrospective fixed-state rescore.}
\label{app:starsim-rescore}
The continuous results reuse forecasts elicited before this repair. The original target pooled trajectories that approximately matched the displayed cumulative count, rather than continuing the displayed state. We retain the original runs and retrospectively replace only their continuous truths and empirical quantile floors. This changes the statistical target: changes in excess reflect both changed outcomes and changed floors, not model deterioration on an unchanged task.

For each of four serialized end-of-day-20 anchors, we hold the complete agent, network, disease and pre-drawn recovery state fixed. We draw 1,000 future continuation seed sets per world, each with four arms (0, 25, 50 and 90 percent coverage): 16,000 branches in total. Horizons 40 and 60 share a branch, and arms share future random roots within a seed set. The target is $P(Y\mid S_{20},a)$ for a fixed full state $S_{20}$ and intervention $a$, not $P(Y\mid\text{report},a)$. Report-only conditioning would require a distribution over compatible latent states. On day 21, vaccination is Bernoulli uptake among all 5,000 living agents, including infected and recovered agents, with probability equal to coverage and 0.95 leaky efficacy; it is neither exact-size nor susceptible-only vaccination. New infections subtract the exact displayed cumulative baseline. Checks verify timing, recipients, unchanged pre-snapshot histories, paired arms and $0\le Y_{40}\le Y_{60}\le 5000-C_{20}$.

Sample size, continuation seeds and analysis rules were frozen before score inspection. A corrected recipient-checker discrepancy at coverage 0.9 affected no simulator behavior, seeds, retained outcomes or normalization; the stopped attempt and amendment remain in the production record. Binary truths and binary scores were not repaired by this continuous run; binary results are retained only as appendix diagnostics and excluded from headline validation and the nine-cell combined score.

No forecasts were newly elicited or imputed. The numbers of saved calls relative to planned calls are 567/576 (unconditional), 552/576 (25\%), 558/576 (50\%) and 556/576 (90\%). There are respectively 6, 14, 11 and 11 model--item cells below the planned three repetitions, including three wholly missing interventional items for DeepSeek V4 Flash. The median at each elicited quantile uses every saved answer; repeated answer values are retained, and no duplicate repetition keys were found. The panels of models with an answer to every item contain 24 baseline models and 23 interventional models, including models with fewer repetitions. Neither uncertainty calculation below captures this elicitation uncertainty.

For continuous associations, we draw 2,000 paired bootstrap samples of the models while holding the empirical truths fixed. Separately, 1,000 replay bootstrap resamples draw continuation indices independently within each world and use the same sampled indices across arms and horizons, recomputing quantiles, floors, scores and correlations. Combined-score model intervals retain the current paper's 10,000 resamples of fitted model scores without refitting; replay intervals refit the scores while keeping the model panel fixed. These are separate sensitivity intervals, not a joint confidence interval. They do not resample worlds, latent anchor states, elicitation calls or external benchmark measurements. A replay correlation interval that collapses to a point means that the sampled rankings are stable; it does not establish the population association. Original truth estimates are held fixed in before/after comparisons. \Cref{tab:starsim-rescore-uncertainty} reports both uncertainty sources.

\begin{table}[htbp]
  \caption{Starsim world parameters of the run. The binary sets use two such regions with transmission probabilities one step apart; the continuous sets use one.}
  \label{tab:starsim-parameters}
  \centering
  \small

  \begin{tabular}{@{}>{\raggedright\arraybackslash}p{4.6cm} >{\raggedright\arraybackslash}p{8.0cm}@{}}
    \toprule
    Parameter & Value \\
    \midrule
    Regions & one (continuous sets); Riverton and Southbay, isolated from each other (binary sets) \\
    Individuals per region & 5,000 \\
    Contact network & random, about 6 contacts per individual per day \\
    Initial share infectious & 0.005 \\
    Transmission probability per contact & one of $\{0.030, 0.040, 0.050, 0.065\}$ per world ($R_0 \approx 1.8$, $2.4$, $3.0$, $3.9$) \\
    Duration of infection & 10 days, mean \\
    Deaths & none \\
    Seeds per world & 300; the displayed run is the seed closest to the median day-20 count \\
    Replays per item & continuous: 1,000 per fixed state and arm; binary: regional draws $157\!\times\!131$, $202\!\times\!157$, $202\!\times\!290$, $131\!\times\!157$ \\
    Vaccine & given on day 21; reduces a vaccinated person's probability of infection per exposure by 95 percent \\
    Coverage & 25, 50 or 90 percent (continuous); 90 percent, or the per-world \emph{hold} value of 20, 10, 15, 20 percent (binary) \\
    Snapshot & day 20 \\
    Horizons & day 40 and day 60 \\
    Report series & parameters as above; cumulative infections at days 0, 5, 10, 15, 20; number infectious at day 20 \\
    \bottomrule
  \end{tabular}
\end{table}

\begin{table}[htbp]
  \caption{Starsim question templates and the intervention sentence, as given to the models. Braces mark the fields filled per item.}
  \label{tab:starsim-templates}
  \centering
  \small
  \setlength{\tabcolsep}{4pt}

  \begin{tabular}{@{}l >{\raggedright\arraybackslash}p{6.2cm} >{\raggedright\arraybackslash}p{3.6cm}@{}}
    \toprule
    Template & Question & Resolution \\
    \midrule
    \texttt{cases\_comparative} & Will Riverton have more cumulative infections than Southbay at day \{T\}? & Riverton's cumulative infections at day $T$ strictly greater than Southbay's; share of matched seed pairs \\
    \texttt{new\_infections} & How many people in Riverton will be newly infected between the end of day 20 and the end of day \{T\}? (That is: cumulative infections at day \{T\} minus the \{c20\} already infected by day 20.) & cumulative infections at day $T$ minus the fixed displayed count at day 20; distribution over fixed-state continuations \\
    intervention & \{Region\} will run a vaccination campaign on day 21 (95\% efficacy, \{c\}\% coverage): on day 21, \{c\}\% of \{Region\}'s population is vaccinated, and vaccination reduces a vaccinated person's probability of infection per exposure by 95\%. & appended to the report as ``PLANNED INTERVENTION''; the two-region form adds ``\{Other\} has no planned intervention.'' \\
    \bottomrule
  \end{tabular}
\end{table}

\subsection{Elicitation protocol by simulation}
\label{app:protocol}

\Cref{tab:protocol} lists how the three runs differ in role, named scoring rule, questions per prompt, reasoning effort, repeated calls and answer format. The shared structure, with the world report before the questions and an instruction to end with a probability or a set of quantiles, is described in \cref{sec:design-protocol}. For the excess-bits score defined in \cref{sec:design}, we bound forecasts to $[0.001, 0.999]$ before the logarithm, so at $q=0$ or $1$ the score has a floor of $-\log_2(0.999) \approx 0.00144$ bits. Micropolis and FreeCiv use the same positive-tail rule, $0<q\le0.05$, and intend $0.05<q<0.95$ for mid-range, but the retained FreeCiv bank includes one $q=0.95$ boundary item; its original 750-item scores are preserved (\cref{app:micropolis-robustness}). They sample the sets differently. FreeCiv draws its mid-range set evenly from five bands of $q$, while the Micropolis mid-range set keeps every instance that falls in the range, so its mean $q$, \MPDMeanQBinary, is lower.

\begin{table}[htbp]
  \caption{Elicitation protocol by simulation. Every prompt places the world report before the question; sampling is at provider defaults. \Cref{app:prompts} gives prompt templates and report excerpts; \cref{sec:design-protocol} summarizes the shared structure.}
  \label{tab:protocol}
  \centering
  \small
  \setlength{\tabcolsep}{4pt}
  \begin{tabular}{@{}>{\raggedright\arraybackslash}p{2.55cm}>{\raggedright\arraybackslash}p{3.35cm}>{\raggedright\arraybackslash}p{3.5cm}>{\raggedright\arraybackslash}p{3.35cm}@{}}
    \toprule
    & Micropolis & FreeCiv & Starsim \\
    \midrule
    Questions per prompt & \MPDPerPromptBinary\ binary, \MPDPerPromptContinuous\ continuous & 50 binary, 20 continuous; 1 for natural conditionals & 1 \\
    Repeated calls per item & 1 & 1 & 3 continuous, 5 binary planned; median of available responses \\
    Reasoning effort & lowest supported requested where controllable; budget-only exceptions & same nominal requests & same nominal requests \\
    Role assigned to the model & ``expert superforecaster, familiar with the work of Tetlock and others'' & none & none \\
    Scoring rule named & Brier score (binary); CRPS (continuous) & ``a proper scoring rule'', not named & none; the opening statement names the task as probabilistic forecasting of simulator output \\
    Continuous answer & p10, p25, p50, p75, p90 & p5, p25, p50, p75, p95 & p10, p25, p50, p75, p90 \\
    Conditional questions & none & second turn of the same conversation after one revealed fact & intervention described in the report: one sentence after the report, in a separate prompt from the unconditional question \\
    \bottomrule
  \end{tabular}
\end{table}

\subsection{Model roster, reasoning settings, hosting and cost}
\label{app:roster}

\Cref{tab:roster} gives the 24 models, ECI \citep{epoch2026eci,ho2025rosetta}, available ForecastBench scores \citep{karger2024forecastbench}, requested reasoning settings and mean observed reasoning-token counts in Starsim. Seven models lack ForecastBench scores: Fable, Opus 5, GPT-5.6 Sol, GPT-5.6 Luna, Gemini 3.7 Flash, DeepSeek V4 Flash and Qwen3.5 Flash. Exact OpenRouter identifiers, requested-host and per-item billing tables are \changednew{in the data release} (\cref{app:release}).

FreeCiv and Micropolis request low effort for 11 models, minimal for five, a 1,024-token budget for four, and no control for four. Starsim used the same lowest-effort rule on September 16. The same requested setting does not imply the same amount of computation: DeepSeek V4 Flash ignores the request, and observed reasoning use varies widely (\cref{tab:roster}). FreeCiv's requested hosts on September 20 follow the committed Micropolis registry; the two DeepSeek conditional sets were rerun after their requested hosts changed. Micropolis's exported provider field comes from the model-ID prefix, not records of the host that actually served each call. Starsim recorded 14 providers for V4 Flash; V3 used DeepInfra (135 responses) and StreamLake (81), Llama DeepInfra (169) and Novita (47), and Fable Anthropic (213) and Azure (3). Routing differs within and across runs; host names do not establish the numerical precision or quantization used.

FreeCiv's scored inputs cost \$174.15: \$26.10 for 1,056 grouped prompts on September 20 plus \$148.05 for conditionals, including \$2.93 for the two reruns. Thus \$29.03 is the additional September 20 expenditure represented in these records; most conditional calls are reused from September 9. The historical ungrouped run cost \$393.83. Grouped-prompt costs are allocated across questions, and the same prompt can contribute to more than one set's prompt count. The conditional arm has 355 distinct turn-1 prompts, 400 reveal cells and 99 no-news prompts. Starsim's \$40.99 covers 5,045 successful scored calls, not every failed or unrecorded attempt. Micropolis's recorded total is \$\MPDCostTotal\ over \MPDCallsTotal\ calls, with \MPDPromptsBinary\ binary and \MPDPromptsContinuous\ continuous prompts per model. These costs describe the recorded runs; they do not estimate the cost of refreshing the benchmark for all models.

\paragraph{Exploratory effort check.}
\label{app:ablation-effort}
On 200 FreeCiv mid-range questions, raising seven models to the next reasoning level (medium or a 2,048-token budget) cost \$4.83; five improved, with excess-Brier changes from $-0.022$ to $+0.014$. Every model remained worse than constant 0.5 (excess 0.081), with negative mean forecast bias at both levels ($-0.27$ to $-0.10$). Correlations with ECI, signed so that higher values mean better forecasting, were $-0.46$ and $-0.39$. Reasoning requests and batch composition both changed, so the comparison does not isolate the causal effect of reasoning effort.

\begin{table}[htbp]
  \caption{The 24 models with their Epoch Capabilities Index (ECI) and ForecastBench overall score (FB; -- where none exists); the reasoning setting in FreeCiv and Micropolis, which is the same in both, where ``budget 1,024'' is a 1,024-token reasoning budget; and the mean number of reasoning tokens per call that each model returned in Starsim at the setting of the September 16 run (all five sets, weighted by calls).}
  \label{tab:roster}
  \centering
  \small
  \setlength{\tabcolsep}{4pt}
\input{data/roster_table}
\end{table}

\subsection{Scoring details}
\label{app:scoring}

We score forecasts against the replay ground truth using proper scoring rules or their quantile approximations, and we report the \emph{excess score}, the expected score of the forecast under the replay distribution minus the expected score of the ground truth itself. For a strictly proper rule on full distributions, excess is nonnegative and zero only at the ground-truth distribution \citep{gneiting2007proper}. The finite-quantile approximation instead assesses the elicited quantiles. For a binary question with forecast $f$, the expected Brier score \citep{brier1950verification} over the replays is $(f-q)^2 + q(1-q)$. The second term is the uncertainty component of \citet{murphy1973vector}, and the \emph{excess Brier score} $(f-q)^2$ is the binary metric on mid-range questions in every simulation.

For a continuous question we approximate the CRPS \citep{matheson1976scoring} with pinball losses at the elicited quantiles. The \emph{floor} is the same score evaluated at the replay distribution's own quantiles, and the \emph{excess CRPS} is the expected score over replays minus this floor. Matching the five elicited quantiles need not match the full distribution, and discrete replay distributions can have nonunique quantile minimizers. Because continuous quantities differ in units, each simulation divides excess CRPS by a fixed scale, which gives the \emph{excess nCRPS}: per city and metric in Micropolis and per horizon in Starsim.

Squared error does little to distinguish tail forecasts, since for any forecast in $[0, 0.05]$ the excess Brier score on a tail question is at most $0.0025$. FreeCiv therefore also scores tail questions in \emph{excess bits}, $\mathrm{KL}(q\,\|\,f)$, the excess of the logarithmic score, which penalizes probability errors logarithmically, with finite endpoint values after clipping. We bound forecasts to $[0.001, 0.999]$ first (\cref{app:protocol}). Starsim scores its binary questions in excess bits as well. Micropolis scores its tail questions in excess bits, as FreeCiv does.

\clearpage

%% file: data/freeciv_items_per_anchor.tex
\begin{tabular}{@{}l rrrrrrrr r@{}}
\toprule
Set & 7001 & 7003 & 7005 & 7008 & 7010 & 7011 & 7014 & 7022 & Total \\
\midrule
Binary bank & 66 & 116 & 112 & 112 & 99 & 109 & 99 & 37 & 750 \\
Tail & 32 & 52 & 38 & 51 & 28 & 36 & 32 & 31 & 300 \\
Mirror & 1 & 8 & 13 & 9 & 2 & 8 & 5 & 4 & 50 \\
Continuous & 34 & 48 & 41 & 37 & 39 & 38 & 37 & 26 & 300 \\
Natural-conditional cells & 33 & 68 & 45 & 65 & 63 & 65 & 38 & 23 & 400 \\
Extra turn-1 questions & 9 & 18 & 13 & 23 & 22 & 19 & 8 & 12 & 124 \\
\bottomrule
\end{tabular}

%% file: data/freeciv_binary_families.tex
\begin{tabular}{@{}l >{\raggedright\arraybackslash}p{5.4cm} r r r r@{}}
\toprule
Family & Resolves YES if \ldots & Mid-range & Tail & Mirror & Nat.\ cond. \\
\midrule
\texttt{EX\_comparative} & civilization A's quantity (score, population, cities, territory, treasury, military units or technologies) at $T$ is strictly greater than B's & 53 & 9 & 2 & 82 \\
\texttt{EX\_government\_at} & the civilization's government at $T$ is the named form; anarchy counts as a form & 46 & 22 & 4 & 6 \\
\texttt{EX\_tech\_discovered} & the civilization discovers the named technology in the window & 50 & 22 & 2 & 1 \\
\texttt{NB1\_threshold} & the civilization's quantity at $T$ is at least the stated value & 48 & 0 & 4 & 2 \\
\texttt{NB1\_value\_threshold} & the civilization's quantity at $T$ is at least the stated value & 51 & 20 & 3 & 83 \\
\texttt{NB4\_drawdown} & the civilization's quantity falls below a stated share of its turn-60 value at the end of any turn in the window & 44 & 20 & 2 & 80 \\
\texttt{NB6\_event} & any city is captured (or at least $k$ captures occur in total) in the window & 7 & 3 & 3 & 1 \\
\texttt{NEW\_civil\_war} & the civilization splits in a civil war in the window & 17 & 20 & 0 & 19 \\
\texttt{NW1\_war\_at} & the pair is at war at $T$ & 49 & 15 & 3 & 25 \\
\texttt{NW2\_diplo} & the pair is in the named state (peace, alliance, or cease-fire or armistice) at the end of some turn in the window & 44 & 21 & 1 & 26 \\
\texttt{NW4\_survival} & the civilization is eliminated in the window & 0 & 16 & 0 & 0 \\
\texttt{NW5\_wonder} & any civilization completes the named wonder in the window & 34 & 21 & 3 & 4 \\
\texttt{S3\_wars\_at\_T} & at least $k$ of the ten pairs are at war at $T$ & 3 & 0 & 1 & 0 \\
\texttt{S4\_gov\_change\_count} & the civilization's government changes at least $k$ times in the window, counting changes into and out of anarchy & 32 & 6 & 4 & 3 \\
\texttt{S5\_civ\_wonders} & the civilization completes at least $k$ great wonders in the window & 43 & 21 & 2 & 2 \\
\texttt{S6\_city\_founding} & the civilization founds at least $k$ new cities in the window & 34 & 2 & 2 & 2 \\
\texttt{S7\_any\_destroyed} & any city is destroyed in the window & 2 & 0 & 0 & 0 \\
\texttt{W1\_wonder\_race} & the named civilization completes the named wonder in the window & 36 & 21 & 2 & 6 \\
\texttt{W2\_directed\_conquest} & civilization A captures at least one city held by B in the window & 49 & 20 & 3 & 22 \\
\texttt{W3\_capture\_k} & the civilization captures at least $k$ cities in the window & 38 & 3 & 4 & 2 \\
\texttt{W4\_lose\_k} & the civilization loses at least $k$ cities to capture in the window & 35 & 3 & 5 & 20 \\
\texttt{W5\_tech\_lead} & the civilization knows strictly more technologies than each other civilization at $T$ & 23 & 15 & 0 & 5 \\
\texttt{W6\_peace\_at} & the pair is at peace at $T$ & 12 & 20 & 0 & 9 \\
\midrule
Total (23 families) & & 750 & 300 & 50 & 400 \\
\bottomrule
\end{tabular}

%% file: data/freeciv_continuous_families.tex
\begin{tabular}{@{}l >{\raggedright\arraybackslash}p{5.8cm} r >{\raggedright\arraybackslash}p{3.1cm}@{}}
\toprule
Family & Resolves to \ldots & Items & Constant \\
\midrule
\texttt{NC1\_value\_at\_T} & the civilization's score, population, number of cities or territory size at $T$ & 60 & cities 30; population 70; score 70; territory 300 \\
\texttt{P5\_techs\_at\_T} & the number of technologies the civilization knows at $T$ & 31 & 9 \\
\texttt{P6\_world\_techs} & the sum over the five civilizations of technologies known at $T$ & 20 & 20 \\
\texttt{P1\_civ\_conquests} & the number of cities the civilization captures in the window & 31 & 10 \\
\texttt{P2\_civ\_losses} & the number of cities the civilization loses to capture in the window & 31 & 10 \\
\texttt{P3\_civ\_founds} & the number of new cities the civilization founds in the window & 31 & 20 \\
\texttt{NC5\_world\_captures} & the total number of city captures by all players in the window & 32 & 50 \\
\texttt{NC14\_world\_wonders} & the number of great wonders completed by all civilizations in the window & 31 & 4 \\
\texttt{S7\_world\_razings} & the number of cities destroyed in the window & 33 & 20 \\
\midrule
Total (9 families) & & 300 & \\
\bottomrule
\end{tabular}

%% file: data/roster_table.tex
\begin{tabular}{@{}l r r l r@{}}
\toprule
Model & ECI & FB & FreeCiv/Micropolis & Starsim tokens \\
\midrule
Claude Fable & 162.49 & -- & low & 424 \\
Claude Opus 5 & 161.54 & -- & low & 342 \\
GPT-5.6 Sol & 161.08 & -- & low & 1,168 \\
GPT-5.5 & 158.63 & 61.1 & low & 1,262 \\
Gemini 3.7 Flash & 156.85 & -- & low & 651 \\
GPT-5.6 Luna & 156.22 & -- & low & 1,198 \\
Claude Sonnet 5 & 155.90 & 59.9 & low & 435 \\
DeepSeek V4 Flash 0731 & 154.43 & -- & low & 18,044 \\
Gemini 3 Flash Preview & 150.71 & 59.1 & minimal & 0 \\
GPT-5 & 150.00 & 60.5 & minimal & 0 \\
o3 & 147.10 & 60.6 & low & 2,112 \\
GPT-5.4 Nano & 146.54 & 55.9 & low & 1,727 \\
o4 Mini & 146.53 & 60.9 & low & 827 \\
GPT-5 Mini & 145.83 & 59.7 & minimal & 0 \\
Gemini 3.1 Flash Lite Preview & 144.66 & 54.6 & minimal & 0 \\
Qwen3.5-Flash & 144.12 & -- & budget 1,024 & 1,003 \\
Claude Haiku 4.5 & 142.91 & 59.7 & budget 1,024 & 684 \\
GPT-5 Nano & 140.85 & 58.0 & minimal & 0 \\
Kimi K2 & 140.42 & 57.6 & none & 3,263 \\
Gemini 2.5 Flash & 140.31 & 58.2 & budget 1,024 & 754 \\
Qwen3 235B A22B & 139.61 & 58.6 & budget 1,024 & 1,020 \\
GPT-4.1 & 137.52 & 58.5 & none & 0 \\
DeepSeek V3 & 133.03 & 57.6 & none & 0 \\
Llama 4 Scout & 130.52 & 56.6 & none & 0 \\
\bottomrule
\end{tabular}

%% file: appendix/B_prompts_and_templates.tex
\begingroup
\renewcommand{\ttdefault}{lmtt}

\section{Prompts, question templates and world reports}
\label{app:prompts}

This appendix reproduces the forecasting instructions, question templates and report excerpts for Micropolis and FreeCiv, followed by complete Starsim examples. In both simulations a prompt has the same structure: an opening instruction, the world report, the question or questions, and the answer format. The administered instructions and question templates are retained below; the availability of the source files is described in \cref{app:release}. We wrapped lines longer than 80 characters at a space and indented the continuation by four spaces, and we truncated the two world reports where stated. Nothing else was changed.\footnote{One exception: the continuous Micropolis preamble writes its inequalities with the \emph{less than or equal} sign, which the typewriter font lacks, so the four signs in that block are printed as \texttt{<=}.}

\subsection{Micropolis}
\label{app:prompts-micropolis}

Our harness combines a preamble, the world report under the heading \texttt{\#\# Game report}, the numbered questions under \texttt{\#\# Questions}, and an epilogue in which \texttt{\{n\}} is replaced by the number of questions. The binary and continuous runs each have their own preamble and epilogue, reproduced below. A binary prompt carries \MPDPerPromptBinary\ questions, and the model answers them in one reply. A continuous prompt carries one snapshot's questions for the five metrics, ordered metric by metric (\MPDPerPromptContinuous\ questions per prompt). The report omits the city's funds and no question asks about them. Both prompts open with the same role, ``an expert superforecaster, familiar with the work of Tetlock and others'', and name the scoring rule.

\paragraph{Binary preamble}
\begin{small}
\begin{verbatim}
You are an expert superforecaster, familiar with the work of
Tetlock and others. Your task is to forecast, for each of several yes/no
    questions about a
city in Micropolis, the open-source release of the original SimCity Classic
    simulation engine, the probability that the question resolves YES.

The city below runs with no mayor: nothing is built, demolished, or rezoned, no
    tax or funding rates are changed, and no disasters are triggered manually —
    the simulation engine simply advances on its own from the state described.
    Whether random disasters are enabled is stated in the game report.

Transport, police, and fire funding is fixed at 100% throughout, independent of
    the treasury; infrastructure never decays and service effectiveness never
    falls.

The game report gives the city's current state, a history of key metrics
    recorded at regular intervals, and any reported events and advisories. Form
your forecasts from what the report shows, combined with your knowledge of
the Micropolis engine.

The turns in the questions are counted in the same unit as the game report's,
and 48 turns make one game year. Every question asks only about what happens
strictly after the report's current turn, up to and including the turn it
names: events that already occurred by the current turn — including those
listed in the report — do not count toward any question. City
classifications, in increasing order: Village (<2k population), Town
(2k–10k), City (10k–50k), Capital (50k–100k), Metropolis (100k–500k),
Megalopolis (>500k).

For each question, give a probability that honestly reflects your uncertainty
about whether it resolves YES. Forecasts are scored with the Brier score
against the realized outcome: probabilities that match your true uncertainty
score best, and overconfidence in either direction scores worse.
\end{verbatim}
\end{small}

\paragraph{Binary epilogue}
\begin{small}
\begin{verbatim}
You MUST provide a probability for every question UNDER ALL CIRCUMSTANCES.
    Always return a number between 0 and 1.
Avoid responding with exactly 0 or 1, unless you are ABSOLUTELY SURE that the
    event is impossible or certain.

You may analyze the data, but you MUST end your response with your probabilities
    in this exact format:
<<<PROBABILITIES>>>
Q1: 0.65
Q2: 0.03
<<<END>>>

Provide one such line for each of the {n} questions, in order, replacing the
    example values with your probability that the answer is YES, as a decimal
    between 0 and 1.
\end{verbatim}
\end{small}

\paragraph{Continuous preamble}
\begin{small}
\begin{verbatim}
You are an expert superforecaster, familiar with the work of
Tetlock and others. Your task is to forecast future values of several metrics
    for a
city in Micropolis, the open-source release of the original SimCity Classic
    simulation engine.

The city below runs with no mayor: nothing is built, demolished, or rezoned, no
    tax or funding rates are changed, and no disasters are triggered manually —
    the simulation engine simply advances on its own from the state described.
    Whether random disasters are enabled is stated in the game report.

Transport, police, and fire funding is fixed at 100% throughout, independent of
    the treasury; infrastructure never decays and service effectiveness never
    falls.

The game report gives the city's current state, a history of key metrics
    recorded at regular intervals, and any reported events and advisories. Form
your forecasts from what the report shows, combined with your knowledge of
the Micropolis engine.

For each question, give percentiles that honestly reflect your uncertainty
about that metric's value at that turn. Forecasts are scored with CRPS
against the realized values: intervals that are too narrow and intervals that
are too wide both score worse than intervals that match your true
uncertainty. In every answer line the five values must be non-decreasing:
p10 <= p25 <= p50 <= p75 <= p90, with p10 always the numerically lowest value.
\end{verbatim}
\end{small}

\paragraph{Continuous epilogue}
\begin{small}
\begin{verbatim}
You MUST provide percentile estimates for every question UNDER ALL
    CIRCUMSTANCES. If for some reason you can't answer, provide reasonable
    mid-range estimates, but always return numeric percentile values.

You may analyze the data, but you MUST end your response with your percentile
    estimates in this exact format:
<<<PERCENTILES>>>
Q1: p10=5, p25=10, p50=15, p75=20, p90=25
Q2: p10=100, p25=200, p50=300, p75=400, p90=500
<<<END>>>

Provide one such line for each of the {n} questions, in order, replacing the
    example values with your actual percentile estimates.
- p10 means you estimate there's a 10% chance the true value is below this
    number
- p25 means you estimate there's a 25% chance the true value is below this
    number
- p50 (median) means you estimate there's a 50% chance the true value is below
    this number
- p75 means you estimate there's a 75% chance the true value is below this
    number
- p90 means you estimate there's a 90% chance the true value is below this
    number
\end{verbatim}
\end{small}

\paragraph{Example world report} This binary example uses city \texttt{bruce} at turn 960 with disasters enabled. The excerpt shows the current-state fields and the first two history rows; the remaining history and event log are omitted. Events in the full example include three disasters and standing pollution, traffic and treasury advisories. Questions follow the report, numbered 1--\MPDPerPromptBinary: the 25 templates below repeat for turns 1104 and 1200; the second prompt covers turns 1296 and 1440.
\begin{small}
\begin{verbatim}
## Game report
City: bruce
Disasters: enabled

CURRENT STATE (turn 960)

  Difficulty: Easy
  City class: City
  city score: 682
  population: 23520
  average traffic: 43
  average pollution: 65
  average crime: 79
  average land value: 77
  Tax rate: 0%

  Infrastructure — Roads 140, Rail 84, Police stations 2, Fire stations 1,
    Seaports 0, Airports 0, Hospitals 1, Stadiums 0, Coal plants 1, Nuclear
    plants 1, Powered zones 48, Unpowered zones 9
  Ground survey — rubble tiles 74, tiles on fire 0, road tiles 125

HISTORY (every 24 turns)

Turn,city score,population,average traffic,average pollution,average
    crime,average land value
0,755,14440,2,59,8,71
24,755,14440,2,68,40,101
[remaining history rows and EVENTS omitted]
\end{verbatim}
\end{small}

\paragraph{Binary question templates} The 25 templates, in the order they appear in the prompt, with \texttt{\{T\}} the horizon turn. Templates A1 to A16 target mid-range probabilities and B1 to B9 target tail probabilities. The analysis assigns every question to the mid-range or tail set by its ground-truth probability, not by this label.
\begin{small}
\begin{description}[leftmargin=2.2em,labelwidth=1.6em,itemsep=0pt,parsep=0pt,topsep=2pt,font=\normalfont]
\item[A1] Will at least one earthquake be reported between the current turn and turn \{T\}?
\item[A2] Will at least one tornado be sighted between the current turn and turn \{T\}?
\item[A3] Will at least one flood be reported between the current turn and turn \{T\}?
\item[A4] Will a monster be sighted between the current turn and turn \{T\}?
\item[A5] Will an airplane crash between the current turn and turn \{T\}?
\item[A6] Will a shipwreck be reported between the current turn and turn \{T\}?
\item[A7] Will the city's population at turn \{T\} be lower than it is at the current turn?
\item[A8] Will the city's population at turn \{T\} be less than half of its current value?
\item[A9] Will the city's population read zero at any yearly checkpoint between the current turn and turn \{T\}?
\item[A10] Will the city's classification at turn \{T\} be lower than it is now (e.g. Metropolis $\rightarrow$ Capital)?
\item[A11] Will a ``Blackouts reported'' advisory appear between the current turn and turn \{T\}?
\item[A12] Will the citywide average pollution level exceed 60 at turn \{T\}?
\item[A13] Will the map hold at least 50 more rubble tiles at turn \{T\} than it does now?
\item[A14] Will at least one map tile be actively burning at turn \{T\}?
\item[A15] Will the map contain fewer road tiles at turn \{T\} than it does now?
\item[A16] Will the city's evaluation score at turn \{T\} be higher than it is now?
\item[B1] Will a nuclear meltdown occur between the current turn and turn \{T\}?
\item[B2] Will two or more earthquakes be reported between the current turn and turn \{T\}?
\item[B3] Will two or more tornadoes be sighted between the current turn and turn \{T\}?
\item[B4] Will a ``Fire reported!'' disaster strike between the current turn and turn \{T\}?
\item[B5] Will a train crash between the current turn and turn \{T\}?
\item[B6] Will two or more separate floods be reported between the current turn and turn \{T\}?
\item[B7] Will the monster be sighted two or more times between the current turn and turn \{T\}?
\item[B8] Will the city's population reach a new all-time high at any yearly checkpoint between the current turn and turn \{T\}?
\item[B9] Will the city's classification at turn \{T\} be higher than it is now?
\end{description}
\end{small}

\paragraph{Continuous question templates} Each continuous question reads ``What will the \emph{metric} be at turn \emph{T}?'' The \emph{metric} is one of population, average traffic, average pollution, average crime and average land value. A sixth template for city funds exists in the code and is not asked, because the funds are censored from the report.

\subsection{FreeCiv}
\label{app:prompts-freeciv}

Our harness sends one user message and no system message. The user message has two parts: the opening instruction, question background and full world report; then the questions and answer instructions. First-party hosts receive the first part with a cache breakpoint, other hosts receive the same text as one string. In the run the paper reports, a prompt for the mid-range set, tail, mirror and continuous sets carries up to 50 binary or 20 continuous questions of one game, numbered, each with its resolution criteria, and asks for one answer line per question inside a delimited block; the instructions follow the Micropolis prompts above, with the FreeCiv background line and no persona. Natural conditionals retain the one-question prompt from September 9, with the same first part, resolution criteria and an answer in tags. Turn 2 of a cell continues that conversation with three messages: the turn-1 message, the model's own turn-1 reply, and a third message that reveals one fact and asks the question again. The no-news control replaces the third message with a statement that nothing new is known. The wording generated by our harness follows; the item names are those of the released question sets.

\paragraph{Grouped binary prompt, opening of the first part} The world report follows the third line.
\begin{small}
\begin{verbatim}
You will be asked several binary forecasting questions about a simulation in
    progress. For each question, your task is to estimate the probability that
    the event described in the question occurs. Please provide your reasoning
    before stating your final probabilities.

Question Background: This is a partial report on a FreeCiv game simulation in
    progress, observed at turn 60. Five AI civilizations are competing.
\end{verbatim}
\end{small}

\paragraph{Second part, grouped binary} Prompt \texttt{seed7008\_bin\_c1of4}, 43 questions; the first two are shown.
\begin{small}
\begin{verbatim}
Questions (43 in total):

1. Will Peruvian have a higher population than Latvian at turn 90?
   Resolution Criteria: Resolves YES if Peruvian's population as recorded at the
    end of turn 90 is strictly greater than Latvian's. Equal values resolve NO.
    The quantity is the same quantity the world report calls Population.

2. Will Latvian have discovered University by turn 90?
   Resolution Criteria: Resolves YES if Latvian discovers University on any turn
    from 61 through 90 inclusive (an event on turn 90 itself counts). Resolves
    NO if it is not discovered by the end of turn 90.

[questions 3 to 43]

Think step by step about the information provided, reason about uncertainty, and
    then state your final confidence that each question resolves YES. Each
    probability should be a number between 0 and 1.

Your answers will be evaluated with a proper scoring rule, so your best strategy
    is to report your honest probability estimate for every question. You MUST
    provide a probability for every question.

Your response SHOULD STRICTLY END with your probabilities in this exact format,
    one line per question, in order:
<<<PROBABILITIES>>>
Q1: 0.65
Q2: 0.03
<<<END>>>
Provide one such line for each of the 43 questions, replacing the example values
    with your probability that the question resolves YES.
\end{verbatim}
\end{small}

\paragraph{Grouped continuous prompt} The first part opens with the paragraph below instead; the background line and the report are the same as in the binary prompt.
\begin{small}
\begin{verbatim}
You will be asked several numeric forecasting questions about a simulation in
    progress. For each question, your task is to estimate the full range of
    plausible values for the quantity described, expressed as percentiles of
    your forecast distribution. Please provide your reasoning before stating
    your final percentiles.
\end{verbatim}
\end{small}

\paragraph{Second part, grouped continuous} Prompt \texttt{seed7003\_cont\_c1of3}, 16 questions; the first is shown.
\begin{small}
\begin{verbatim}
Questions (16 in total):

1. How many wonders will be completed (across all civilizations) between turn 61
    and turn 90?
   Resolution Criteria: Resolves to the number of great wonders completed by all
    civilizations on any turn from 61 through 90 inclusive (an event on turn 90
    itself counts). Small wonders such as a palace do not count.

[questions 2 to 16]

Think step by step about the information provided, reason about uncertainty, and
    give your final answer to each question as five percentiles of your forecast
    distribution: p5, p25, p50, p75, p95. "p5" means you estimate a 5% chance
    the true value falls below that number; "p50" is your median. Your
    percentiles must be non-decreasing (p5 <= p25 <= p50 <= p75 <= p95). Every
    answer is a whole number.

Your answers will be evaluated with a proper scoring rule, so your best strategy
    is to report your honest estimate of each distribution - percentiles that
    are too narrow and percentiles that are too wide will both cost you. You
    MUST provide percentiles for every question.

Your response SHOULD STRICTLY END with your percentiles in this exact format, one
    line per question, in order:
<<<PERCENTILES>>>
Q1: p5=5, p25=10, p50=15, p75=20, p95=25
Q2: p5=100, p25=200, p50=300, p75=400, p95=500
<<<END>>>
Provide one such line for each of the 16 questions, replacing the example values
    with your actual percentile estimates.
\end{verbatim}
\end{small}

\paragraph{Natural-conditional turn-1 prompt, opening of the first part} One question per prompt, as in the run of September 9. The world report follows the third line.
\begin{small}
\begin{verbatim}
You will be asked a binary forecasting question about a simulation in progress.
    Your task is to estimate the probability that the event described in the
    question occurs. Please provide your reasoning before stating your final
    probability.

Question Background: This is a partial report on a FreeCiv game simulation in
    progress, observed at turn 60. Five AI civilizations are competing.
\end{verbatim}
\end{small}

\paragraph{Second part, natural-conditional turn 1} Item \texttt{seed7003:b687}.
\begin{small}
\begin{verbatim}
Question Title: Will Ghana and Sammarinese reach a cease-fire or armistice at
    any point between turns 61 and 120?

Resolution Criteria: Resolves YES if the diplomatic state between Ghana and
    Sammarinese is Cease-fire or Armistice at the end of any turn from 61
    through 120 inclusive. Resolves NO if the state never occurs in that window
    (war, peace and alliance do not count).

Think step by step about the information provided, reason about uncertainty, and
    put your final confidence that the question resolves YES in <probability>
    </probability> tags. The probability should be a number between 0 and 1.

Your answer will be evaluated with a proper scoring rule, so your best strategy
    is to report your honest probability estimate.

Your final answer should be the probability that the event resolves YES and your
    response SHOULD STRICTLY END with <probability> </probability> tags.
\end{verbatim}
\end{small}

\paragraph{Turn-2 message of a natural-conditional cell} Cell \texttt{seed7003:b687|seed7003:r30}. The first sentence is the same in every cell up to the revealed fact; the turn on which the fact occurred is never disclosed.
\begin{small}
\begin{verbatim}
One fact about turns 61 through 90 has been revealed to you: Rhenish had at
    least four more cities at turn 90 than at turn 60.

Given this, please answer the same question again. Reason about what, if
    anything, this changes, and end your response with your updated probability
    in <probability> </probability> tags.
\end{verbatim}
\end{small}

\paragraph{No-news control message}
\begin{small}
\begin{verbatim}
The game has continued. No new information about turns 61 through 90 is
    available. If you wish, revise your forecast.

Please answer the same question again and end your response with your
    probability in <probability> </probability> tags.
\end{verbatim}
\end{small}

\paragraph{Reading the answer} For a grouped prompt we read the last delimited block in the reply and match each line to its question by the stated number. When the block is missing we accept numbered lines outside it and, for percentiles, five numbers on one line or on the line after a restated question, and mark the row as leniently parsed; each grouped prompt was sent once, a missing binary answer scores as 0.5, and a missing continuous answer is dropped. For a natural-conditional prompt the answer is the number inside the last \texttt{<probability>} tags; when the tags were missing we accepted a trailing ``probability \ldots 0.xx'' or a bare \texttt{<0.xx>} at the end and marked the row as leniently parsed, and when neither could be read we asked once more.

\paragraph{World report} The turn-60 report of anchor \texttt{seed7001} has 319 lines (19.2~KB). This excerpt retains the header, civilization list, current-state column headings and first civilization row. The omitted sections contain the other current states, diplomacy and borders, government changes, five-turn economic and military series, event definitions and 145 chronological events from turns 2--60.
\begin{small}
\begin{verbatim}
WORLD REPORT TXT v1
Game ID: seed7001
Snapshot turn: 60
Turns analyzed: 1..60
Barbarians: default settings (barbarian and pirate raids begin around turn 60;
    none exist at turn 60)
Map size: 78x52

CIVILIZATIONS
ID | Name
---+---------------
0  | Brazzaville-Congolese
1  | Santomean
2  | Estonian
3  | Guinean
4  | Sahrawi

CURRENT STATE (TURN 60)
ID | Name                  | Score | Rank | Government | Treasury | Population |
    Techs | Wonders | Cities | Territory | Units | Military | Science | Trade |
    Food  | Shields
---+-----------------------+-------+------+------------+----------+------------+
    -------+---------+--------+-----------+-------+----------+---------+-------+
    -------+--------
0  | Brazzaville-Congolese | 38    | 3    | Despotism  | 449      | 21         |
    8     | 0       | 14     | 188       | 8     | 1        | 60      | 21.0  |
    42.0  | 39.0
[remaining civilization rows and report sections omitted]
\end{verbatim}
\end{small}

\paragraph{Question families} We generated the FreeCiv questions from 23 binary and nine continuous families. Each family is a template with parameters (a civilization, a pair, a threshold, a count or a turn), and its resolution criteria state how we read the quantity from the saved game. The generator is released with the question sets; the tables below give each family's identifier, its number of questions, and one question drawn from it. \Cref{tab:freeciv-binary-families} covers the mid-range set and the tail set, \cref{tab:freeciv-continuous-families} the continuous set, and \cref{tab:freeciv-reveals} the kinds of fact revealed in the natural-conditional cells.

\begin{center}
\small
\setlength{\tabcolsep}{4pt}
\begin{tabular}{@{}l r r >{\raggedright\arraybackslash}p{8.1cm}@{}}
\toprule
Family & Mid-range & Tails & Example question \\
\midrule
\texttt{EX\_tech\_discovered} & 50 & 22 & Will Brazzaville-Congolese have discovered Seafaring by turn 90? \\
\texttt{NB1\_value\_threshold} & 51 & 20 & Will Chinese's territory size be at least 405 at turn 90? \\
\texttt{W2\_directed\_conquest} & 49 & 20 & Will Guyanese capture at least one city from Libyan by turn 90? \\
\texttt{EX\_government\_at} & 46 & 22 & Will Abyssinian's government be Anarchy at turn 90? \\
\texttt{NW2\_diplo} & 44 & 21 & Will Central American and Spanish be at formal Peace at any point between turns 61 and 90? \\
\texttt{NW1\_war\_at} & 49 & 15 & Will Spanish and Peruvian be at war at turn 90? \\
\texttt{NB4\_drawdown} & 44 & 20 & Will Ghana's number of cities drop below 99\% of its turn-60 level at any point up to turn 90? \\
\texttt{S5\_civ\_wonders} & 43 & 21 & Will Beninese complete at least 1 wonder by turn 90? \\
\texttt{EX\_comparative} & 53 & 9 & Will Bissau-Guinean have a higher territory size than Tajik at turn 90? \\
\texttt{W1\_wonder\_race} & 36 & 21 & Will Greek complete the Great Wall by turn 90? \\
\texttt{NW5\_wonder} & 34 & 21 & Will Marco Polo's Embassy be completed by any civilization by turn 90? \\
\texttt{NB1\_threshold} & 48 & 0 & Will Cañari's number of technologies be at least 22 at turn 90? \\
\texttt{W3\_capture\_k} & 38 & 3 & Will American capture at least 2 cities by turn 90? \\
\texttt{W5\_tech\_lead} & 23 & 15 & Will Crimean Tatar know strictly more technologies than every other civilization at turn 90? \\
\texttt{S4\_gov\_change\_count} & 32 & 6 & Will Guinean's government change at least 1 time by turn 90? (Anarchy counts as a form of government, for this question.) \\
\texttt{W4\_lose\_k} & 35 & 3 & Will Saka lose at least 5 cities by turn 90? \\
\texttt{NEW\_civil\_war} & 17 & 20 & Will Rhenish suffer a civil war (a breakaway civilization splitting off) by turn 120? \\
\texttt{S6\_city\_founding} & 34 & 2 & Will Ghana found at least 8 new cities by turn 90? \\
\texttt{W6\_peace\_at} & 12 & 20 & Will Santomean and Guinean have a formal peace in effect at turn 90? \\
\texttt{NW4\_survival} & 0 & 16 & Will Western Roman be eliminated (cease to exist as a civilization) by turn 210? \\
\texttt{NB6\_event} & 7 & 3 & Will at least 10 cities be conquered (in total) by turn 90? \\
\texttt{S3\_wars\_at\_T} & 3 & 0 & Will at least 4 pairs of civilizations be at war at exactly turn 90? \\
\texttt{S7\_any\_destroyed} & 2 & 0 & Will any city be destroyed by turn 90? \\
\midrule
Total & 750 & 300 & \\
\bottomrule
\end{tabular}
\captionof{table}{The 23 binary question families of FreeCiv: the questions each contributes to the 750-question mid-range set and to the 300-question tail set, and one example from each. 22 families appear in the mid-range set and 20 in the tail set.}
\label{tab:freeciv-binary-families}
\end{center}

\begin{center}
\small
\setlength{\tabcolsep}{4pt}
\begin{tabular}{@{}l r >{\raggedright\arraybackslash}p{8.8cm}@{}}
\toprule
Family & Questions & Example question \\
\midrule
\texttt{NC1\_value\_at\_T} & 60 & What will Peruvian's number of cities be at turn 90? \\
\texttt{S7\_world\_razings} & 33 & How many cities will be destroyed in total by turn 90? \\
\texttt{NC5\_world\_captures} & 32 & How many cities will be captured (in total, across all civilizations) by turn 90? \\
\texttt{NC14\_world\_wonders} & 31 & How many wonders will be completed (across all civilizations) between turn 61 and turn 90? \\
\texttt{P1\_civ\_conquests} & 31 & How many cities will Ostrogothic capture by turn 90? \\
\texttt{P2\_civ\_losses} & 31 & How many cities will Otomí lose by turn 90? \\
\texttt{P3\_civ\_founds} & 31 & How many new cities will Fijian found by turn 90? \\
\texttt{P5\_techs\_at\_T} & 31 & How many technologies will Corsican know at turn 90? \\
\texttt{P6\_world\_techs} & 20 & What will the combined number of technologies known by all five civilizations be at turn 90? \\
\midrule
Total & 300 & \\
\bottomrule
\end{tabular}
\captionof{table}{The nine continuous question families of FreeCiv. The answer is five percentiles; every quantity is a whole number.}
\label{tab:freeciv-continuous-families}
\end{center}

\begin{center}
\small
\setlength{\tabcolsep}{4pt}
\begin{tabular}{@{}l r >{\raggedright\arraybackslash}p{8.8cm}@{}}
\toprule
Revealed fact & Cells & Example \\
\midrule
\texttt{civ\_lost3} & 191 & American lost at least three cities. \\
\texttt{pair\_war\_began} & 90 & Pelasgian and Abyssinian went to war. \\
\texttt{civ\_lost} & 39 & Trinidadian and Tobagonian lost a city. \\
\texttt{civ\_score\_leader} & 13 & American had the highest score of any civilization at turn 90. \\
\texttt{civ\_govchange} & 12 & Greek changed its form of government (not counting a spell of anarchy). \\
\texttt{pair\_ceasefire} & 12 & Sammarinese and Corsican reached a cease-fire or armistice. \\
\texttt{civ\_cities\_grew4} & 10 & Rhenish had at least four more cities at turn 90 than at turn 60. \\
\texttt{civ\_cities\_fell2} & 8 & Saka had at least two fewer cities at turn 90 than at turn 60. \\
\texttt{pair\_peace} & 5 & American and Spanish signed a peace treaty. \\
\texttt{civ\_captured} & 5 & Peruvian captured a city. \\
\texttt{civ\_wonder} & 5 & Spanish completed a wonder. \\
\texttt{civ\_techs\_K} & 5 & Belgian knew at least 21 technologies at turn 90. \\
\texttt{civ\_anarchy} & 4 & American fell into anarchy at some point. \\
\texttt{pair\_contact} & 1 & Estonian and Sahrawi made first contact. \\
\midrule
Total & 400 & \\
\bottomrule
\end{tabular}
\captionof{table}{The 14 kinds of fact revealed at turn 2 in the 400 natural-conditional cells, with one example of each. The reveal sentence is ``One fact about turns 61 through 90 has been revealed to you:'' followed by the fact.}
\label{tab:freeciv-reveals}
\end{center}

\subsection{Starsim}
\label{app:prompts-starsim}

Every Starsim prompt combines the same four parts in one text: a statement of provenance, the world report, the question, and the answer format. The interventional items add one line, \texttt{PLANNED INTERVENTION}, between the report and the question; nothing else changes, and the unconditional and interventional versions of an item go out as separate prompts. No role is assigned and no scoring rule is named. The first block is the complete prompt of one continuous interventional item (world w0, day 40, 50 percent coverage); the unconditional item is the same prompt without the intervention line, and the 25 and 90 percent items change only the two coverage figures in that line. The second block is the complete prompt of one binary unconditional item (world w0, day 40); its interventional and hold items insert the two-region form of the intervention line, which ends ``Southbay has no planned intervention.'' The prompts use the symbols $\approx$ and $\geq$, written here as \texttt{\textasciitilde=} and \texttt{>=}. Lines are wrapped for the page. The institution named in the statement of provenance is redacted for review.

\paragraph{Continuous item, interventional (world w0, day 40, 50 percent coverage).}
\begin{small}
\begin{verbatim}
This is a question from ForecastBench-Sim, a forecasting-calibration benchmark
    built by [redacted for anonymous review]. The data below comes from a
    toy SIR simulation (the open-source starsim package): one synthetic
    population of 5,000 agents with fixed, stated parameters, no behavior
    change, no travel, and no interventions unless stated. Your task is
    probabilistic forecasting of simulator output.

PANDEMIC WORLD REPORT - Day 20

One isolated region, Riverton: 5,000 people, no travel in or out.
Disease model: SIR (susceptible -> infectious -> recovered; no deaths).
    Infected people are infectious for 10 days on average, then recover with
    permanent immunity. Each person has about 6 contacts per day. The
    simulation is agent-based and stochastic.
    Transmission probability per contact per day 0.030 (basic reproduction
      number R0 ~= 1.8); 28 people infectious at day 0.

Observed cumulative infections at the end of each day (people ever infected,
    including those infectious at day 0):
    Riverton: d0:32 -> d5:66 -> d10:118 -> d15:225 -> d20:431   (day 20: 327
      currently infectious)

PLANNED INTERVENTION: Riverton will run a vaccination campaign on day 21 (95%
    efficacy, 50% coverage): on day 21, 50% of Riverton's population is
    vaccinated, and vaccination reduces a vaccinated person's probability of
    infection per exposure by 95%.

QUESTION: How many people in Riverton will be newly infected between the end
    of day 20 and the end of day 40? (That is: cumulative infections at day 40
    minus the 431 already infected by day 20.)

The report ends at day 20. The simulation is stochastic, so give a
    distribution: your 10th, 25th, 50th, 75th and 90th percentiles for the
    number of people (integers >= 0). You may reason briefly first. Then end
    your reply with one JSON object, alone on its final line, with exactly
    these keys:
    "p10", "p25", "p50", "p75", "p90"  (non-decreasing numbers of people)
\end{verbatim}
\end{small}

\paragraph{Binary item, unconditional (world w0, day 40).}
\begin{small}
\begin{verbatim}
This is a question from ForecastBench-Sim, a forecasting-calibration benchmark
    built by [redacted for anonymous review]. The data below comes from a
    toy SIR simulation (the open-source starsim package): two synthetic
    populations of 5,000 agents each, with fixed, stated parameters, no
    behavior change, no travel between regions, and no interventions unless
    stated. Your task is probability calibration on simulator output.

PANDEMIC WORLD REPORT - Day 20

Two isolated regions, Riverton and Southbay: 5,000 people each, no travel
    between them.
Disease model: SIR (susceptible -> infectious -> recovered; no deaths).
    Infected people are infectious for 10 days on average, then recover with
    permanent immunity. Each person has about 6 contacts per day. The
    simulation is agent-based and stochastic.
    Riverton: transmission probability per contact per day 0.040 (basic
      reproduction number R0 ~= 2.4); 23 people infectious at day 0.
    Southbay: transmission probability per contact per day 0.030 (R0 ~= 1.8);
      28 people infectious at day 0.

Observed cumulative infections at the end of each day (people ever infected,
    including those infectious at day 0):
    Riverton: d0:28 -> d5:88 -> d10:227 -> d15:548 -> d20:1129   (day 20: 911
      currently infectious)
    Southbay: d0:32 -> d5:66 -> d10:118 -> d15:225 -> d20:431   (day 20: 327
      currently infectious)

QUESTION: Will Riverton have more cumulative infections than Southbay at day
    40?

The report ends at day 20. You may reason briefly first. Then end your reply
    with one JSON object, alone on its final line, with exactly this key:
    "p": probability that the answer is YES, a number in [0,1]
\end{verbatim}
\end{small}

\paragraph{Two-region intervention line (world w0, 90 percent coverage).}
\begin{small}
\begin{verbatim}
PLANNED INTERVENTION: Riverton will run a vaccination campaign on day 21 (95%
    efficacy, 90% coverage): on day 21, 90% of Riverton's population is
    vaccinated, and vaccination reduces a vaccinated person's probability of
    infection per exposure by 95%. Southbay has no planned intervention.
\end{verbatim}
\end{small}

\endgroup

%% file: appendix/C_full_results.tex
\section{Full results}
\label{app:full-results}

This appendix gives the per-model scores behind the figures of \cref{sec:micropolis,sec:freeciv,sec:starsim} and the correlations of \cref{sec:validation}. Every table lists models in descending order of the Epoch Capabilities Index (ECI, \cref{sec:design-models}) and uses the short model names of the figures. Scores follow the definitions of \cref{sec:design-scoring}. Excess Brier score, excess bits and excess normalized CRPS (nCRPS) are lower-is-better. Required artifacts are described in \cref{app:release}.

\subsection{Baselines}
\label{app:baselines}

\paragraph{Base rates.}
A binary question's base rate (\cref{sec:design-baselines}) is the mean $q$ of the questions with the same template at the same horizon in the other worlds of its simulation. It is what a forecaster would know after seeing the same questions resolve in other worlds, which is what publishing a benchmark reveals (\cref{sec:discussion}). A Micropolis question uses both snapshots of each of the other 14 cities; a FreeCiv question uses the other 7 games. The pool is every question the generator produced. Micropolis asks all 25 of its templates at every snapshot and horizon, so its pool is the 3,000 asked questions, including those with $q=0$ or $q\ge 0.95$. FreeCiv's is the \changed{\BLFCCandidates{} candidates} with $0<q<1$ from which the draw selected its questions (\cref{app:worlds-freeciv}). A Micropolis template is one of the 25 questions of \cref{app:prompts-micropolis}. A FreeCiv template is a family with the parameters its text states, such as a count, a metric, a government, a technology or a wonder, but not the civilizations. A threshold, as in \emph{will a civilization's score be at least 180 at turn 150?}, is matched by its ratio to the civilization's turn-60 value in the world report, in bins of width 0.5 on a log scale, so that thresholds equally far above or below each civilization's current value share a base rate. Military units, often zero at turn 60, are matched by the threshold itself. When no other game has the template at that horizon, we fall back to the same template at any horizon, then the family at the same horizon, and finally the family at any horizon. Each Micropolis base rate averages 28 questions, and each FreeCiv one a median of \changed{\BLFCQuestionsPerRate}. Note that the mean is the best constant forecast for a group of questions under both the excess Brier score and excess bits.

\Cref{tab:binary-baselines} compares this choice with alternatives. \changed{Coarser base rates, one per simulation or per simulation and horizon, ignore that questions differ widely in how likely they are, and score much worse on tail questions; pooling over horizons changes little. A base rate that includes the forecast world uses the question's own $q$, and scores better for that reason. So do base rates computed within the question's set, and the constant 0.05 on tail questions, since the sets are defined by $q$: a tail question's base rate is then at most 0.05. In FreeCiv, a family without its parameters mixes questions such as ``at least one wonder'' and ``at least three'', and does much worse on tail questions. Pooling only the asked questions instead of all candidates leads to the same conclusions (\BLFCMidAsked{} on mid-range and \BLFCTailAsked{} on tail questions), but each rate then rests on a median of four questions, selected on $q$.}

\changed{Why no model beats the base rate on FreeCiv's mid-range set, while \BLFCTailBeat{} do on the tail set, lies in the base rate, not the models. Every binary prompt mixes mid-range, tail and mirror questions, and the models' forecasts barely follow $q$.} On the mid-range set they average \BLFCMidMeanForecast{} against a mean $q$ of \BLFCMidMeanQ, \changed{while the base rate averages \BLFCMidMeanRate{} and correlates more closely with $q$ (\cref{sec:freeciv-results}).} On the tail set the forecasts average \BLFCTailMeanForecast{} \changed{and the base rate \BLFCTailMeanRate, both far above the mean $q$ of \BLFCTailMeanQ: a template's base rate is pulled up by its mid-range questions.} \changed{On the mid-range set the base rate is thus much closer to $q$ than the forecasts are; on the tail set both are about equally far above it, and the models that beat it are those that forecast lowest overall.}

\paragraph{Climatological change.}
The climatological change forecasts a continuous quantity as its current value plus a typical change, taken from the other worlds of the same simulation. It is built in three steps. First, for every replay of the same quantity at the same horizon in another world, we compute the change: the replay's outcome minus that world's value at the snapshot. Second, we take the percentiles of all these changes that the question asks for (the 10th to the 90th; in FreeCiv the 5th to the 95th). Third, we add them to the question's own value at the snapshot, from its world report. For example, if the other worlds' changes have a median of $+3$, the forecast's median is the current value plus 3.

The other worlds are the other 28 snapshots in Micropolis, the other games' candidates of the same family or metric in FreeCiv, among the \changed{\BLFCContCandidates{} eligible ones} (\cref{app:worlds-freeciv}), and the other three worlds in Starsim. Micropolis's cities differ in size by two orders of magnitude, so there we measure each change relative to its city's scale (\cref{sec:design-scoring}) and convert it back with the forecast city's scale. When a question counts events after the snapshot, as most FreeCiv questions and all Starsim ones do, the current value is zero and the forecast is simply the other worlds' counts. Finally, we clip the forecast to the quantity's possible range: nonnegative, within the engine's bounds for Micropolis's averages, and at most the people not yet infected in Starsim.

\Cref{tab:continuous-baselines} compares this choice with alternatives. We initially considered a simpler \emph{persistence} baseline: every quantile at the latest value of the forecast quantity. \changed{It forecasts no events for counts after the snapshot and misses the strong growth of FreeCiv's quantities after turn 60 and of Starsim's epidemics; in those two simulations it scores worse than every model.} In Micropolis, where the quantities move less, it scores much closer to the models, but 23 of 24 still beat it. Pooling the other worlds' values rather than their changes fails in Micropolis, whose cities differ in scale; \changed{in the other two simulations most questions count events after the snapshot, so the two coincide or nearly so.}

\paragraph{SIR forecast (Starsim).}
The world report states that the disease follows an SIR model \citep{hethcote2000mathematics} and gives its parameters: 5,000 people with about six contacts a day, the transmission probability $p$ per contact and day, and a ten-day mean infectious period. It also gives the cumulative infections $C_{20}$ and the number $I_{20}$ infectious on day 20. The SIR forecast integrates $\dot S=-\beta SI/N$, $\dot I=\beta SI/N-I/10$ with $\beta=6p$ and $N=5{,}000$ from $S=N-C_{20}$, $I=I_{20}$ on day 20 to the horizon, and sets every quantile to the new infections $S(20)-S(T)$. It uses only numbers printed in the report, and lands within \BLSSSIRErrMin--\BLSSSIRErrMax\% of the true median for every item. The simulator uses stochastic agent-level transmission, a dynamic contact network and lognormally distributed infectious periods. Started instead from the day-0 infections alone, the same equations get the epidemic's timing wrong and score far worse. This deterministic forecast does not represent outcome uncertainty, yet only \BLSSContSIRBeat{} models beat it, against \BLSSContBeat{} for the climatological change.

\paragraph{Independence (conditionals).}
A conditional question is forecast as if the condition made no difference, by the unconditional baseline of the same question. A FreeCiv natural conditional gets the base rate of $Y$, computed as above, whatever fact is revealed; its forecasts before and after the fact are the same, so its gain is zero. A Starsim interventional question gets the climatological change of new infections without vaccination at the same horizon. A model can beat independence by forecasting the unconditional quantity more accurately, by using the condition, or both; two diagnostics help distinguish these effects. In FreeCiv, ignoring the fact but using the true $p(Y)$ scores \BLFCNatcondTrueP, better than every model: the revealed facts move $p(Y)$ little compared with the models' errors on $p(Y)$ itself. In Starsim, the true no-vaccine distribution scores \BLSSIntTrueNoVax, worse than the baseline's \BLSSInt{} only because the other worlds' outcomes are sometimes smaller and so happen to lie closer to the vaccinated ones; and \BLSSIntOwnBeat{} of \BLSSIntOwnN{} models beat their own no-vaccine forecast reused under vaccination.

\begin{table}[htbp]
  \caption{Binary base rates, the one used and alternatives, scored on the mid-range (excess Brier) and tail (excess bits) sets; in parentheses, the number of the 24 models with a lower mean score. Indented rows vary the base rate used. $^\dagger$Uses the question's set, which is defined by $q$.}
  \label{tab:binary-baselines}
  \centering
  \small
\input{data/appendix_tables/binary_baselines}
\end{table}

\begin{table}[htbp]
  \caption{Continuous baselines, the one used and alternatives, scored in excess nCRPS on the unconditional continuous questions; in parentheses, the number of the 24 models with a lower mean score. Persistence: every quantile at the quantity's value at the snapshot, zero for counts after it. In Starsim, which counts new infections from day 20, the other worlds' values are their changes. The SIR forecast applies only to Starsim.}
  \label{tab:continuous-baselines}
  \centering
  \small
\input{data/appendix_tables/continuous_baselines}
\end{table}

\clearpage
\subsection{Micropolis}
\label{app:full-micropolis}

\Cref{tab:micropolis-models} lists all 24 models of the Micropolis run, pooled over the four horizons. Every model returned a forecast that we could parse on at least \MPDMinParseBinary\% of its 3,000 binary questions and \MPDMinParseContinuous\% of its \MPDNQuestionsContinuous\ continuous questions. \Cref{tab:micropolis-horizon} gives the two binary scores at each horizon, and \cref{tab:micropolis-continuous} the excess nCRPS by horizon and by metric.

\begin{table}[htbp]
  \caption{Micropolis scores per model, pooled over the four horizons; lower is better. Excess Brier score on mid-range questions ($5\% < q < 95\%$, \MPDNQuestionsBinary\ per model), excess bits on tail questions ($0 < q \le 5\%$, \MPDNQuestionsTail), and excess nCRPS on the \MPDNQuestionsContinuous\ continuous questions. Parsed: percentage of questions with a readable forecast. The last row is the baseline, the base rate for binary scores and the climatological change for continuous ones (\cref{app:baselines}); scores lower than the baseline's are in bold.
  }
  \label{tab:micropolis-models}
  \centering
  \small
\input{data/appendix_tables/micropolis_models}
\end{table}

\begin{table}[htbp]
  \caption{Micropolis scores by horizon (game years after the snapshot); lower is better. Mid-range questions ($5\% < q < 95\%$) are scored in excess Brier score and tail questions ($0 < q \le 5\%$) in excess bits. Each cell averages forecast scores for one model at one horizon. The last row is the base rate at each horizon (\cref{app:baselines}); scores lower than the base rate's are in bold.}
  \label{tab:micropolis-horizon}
  \centering
  \scriptsize
\input{data/appendix_tables/micropolis_horizon}
\end{table}

\begin{table}[htbp]
  \caption{Micropolis excess nCRPS on the \MPDNQuestionsContinuous\ continuous questions, by horizon and by metric; lower is better. The CRPS of each forecast is scored against the replay distribution and divided by a scale computed per city and metric. The last row is the climatological change for each horizon and metric (\cref{app:baselines}); scores lower than the climatological change's are in bold.}
  \label{tab:micropolis-continuous}
  \centering
  \scriptsize
\input{data/appendix_tables/micropolis_continuous}
\end{table}

\begin{figure}[htbp]
  \centering
  \includegraphics[width=\linewidth]{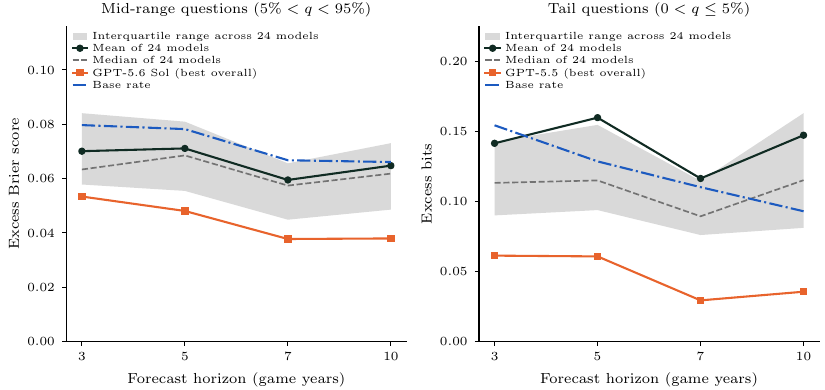}
  \caption{Micropolis forecasting scores by forecast horizon (\MPDHorizonYears\ game years); lower is better in both panels. Left: mid-range questions ($5\% < q < 95\%$), excess Brier score. Right: tail questions ($0 < q \le 5\%$), excess bits. Solid line: mean over the \MPDNModelsBinary\ models; dashed: median; band: interquartile range across models; orange: the best model pooled over horizons (\MPDBestBinaryModel\ for mid-range, \MPDBestTailModel\ for the tail). Blue dash-dotted: the base rate at each horizon (\cref{app:baselines}).}
  \label{fig:micropolis-horizon}
\end{figure}

\clearpage
\Cref{fig:micropolis-horizon} separates mid-range and tail scores.

\begin{figure}[htbp]
  \centering
  \includegraphics[width=\linewidth]{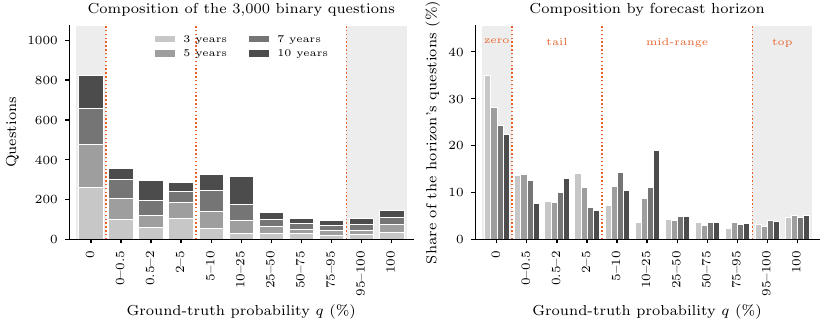}
  \caption{Composition of the Micropolis binary question set by ground-truth probability $q$, over the \MPDBandNQuestions\ questions each model is asked. Left: questions per probability band, stacked by forecast horizon. Right: the same counts as each horizon's share of its own questions. The bands contain unequal numbers of questions; a template's intended band does not determine which band its replay probability falls in. Each band includes its upper edge, so that $q=5\%$ is a tail question, except that the top set starts at $q=95\%$, and $q=0$ and $q=1$ have bars of their own. Dotted lines separate the four sets: zero ($q=0$, \MPDBandZeroShare\% of the questions), tail ($0<q\le 5\%$, \MPDBandTailShare\%), mid-range ($5\%<q<95\%$, \MPDBandMidShare\%) and top ($q\ge 95\%$, \MPDBandTopShare\%). The shaded zero and top sets are in neither scored set.}
  \label{fig:micropolis-bands}
\end{figure}

\Cref{fig:micropolis-bands} shows the realized base-rate composition. Most mid-range questions lie below $q=25\%$, rather than near one half.

As horizons lengthen, the same templates can move between probability bands. Horizon comparisons therefore change question composition and position within horizon-ordered prompts (\cref{app:micropolis-robustness}).

\paragraph{Selection and uncertainty checks.}\label{app:micropolis-robustness}
The current Micropolis split has 940 positive-tail, 983 mid-range, 826 zero and 251 top instances. Selection and scoring use the same empirical replay probabilities. Replay estimation error near a cutoff can change which questions enter a set, and excluding zero estimates selects positive estimates of very rare events. Zero successes in 1,000 independent Bernoulli replays would imply a one-sided 95\% upper bound of 0.003, not a known probability of zero; this is neither a joint bound over items nor a verification of replay independence. Varying the positive-tail cutoff through 1, 2.5, 4, 5 and 6 percent gives ECI correlations 0.708, 0.729, 0.757, 0.750 and 0.763. These checks vary the composition of the question set; they do not correct for selection bias or finite-replay error.

The existing whole-city bootstrap intervals are \MPDRhoBinaryCICities\ (mid-range), \MPDRhoTailCICities\ (tail), and \MPDRhoContinuousCICities\ (continuous). Each resampled city contributes both snapshots and all its questions together. They measure variation from reweighting the 15 selected cities while holding forecasts and empirical truths fixed; they do not combine uncertainty from models and replays.

For the 21 missing mid-range and 10 missing tail answer entries, a sensitivity check that assigns 0.5 to missing answers leaves ECI correlations unchanged. The reported Micropolis scores still exclude unparsed forecasts. Continuous excess is unbounded, and the exported scores lack the original quantiles, so we cannot establish a finite worst-case bound or rescore the forecasts against new replay samples from these files alone. Audit code, exact inputs and sensitivities are retained separately \changednew{and available on request}.

\subsection{Starsim}
\label{app:full-starsim}

\paragraph{Historical binary diagnostic, excluded from headline scores.}
The report-visible rule ``the region ahead at day 20 remains ahead'' solves all eight unconditional binary questions and attains the clipped minimum, $-\log_2(0.999)=0.0014434$ bits per item. Reversing that rule solves all eight 90 percent items. Ignoring intervention gets the correct side on all eight hold items, although their non-endpoint probabilities still require calibration. These are findings about the selected cached sets, not guarantees on new worlds. The unconditional binary ECI association ($\rho=0.64$ in skill orientation) describes this ranking diagnostic; it is excluded from headline validation and the combined score because of what the task measures, independently of that correlation.

\paragraph{Historical hold set.}
\label{app:ablation-hold}
The unchanged, appendix-only binary \texttt{hold} set has eight items, four per horizon, with five responses planned per item. The vaccinated region is the one ahead at day 20; coverage is the largest 5-percent step preserving the ranking at both horizons (20, 10, 15 and 20 percent). Historical probabilities are fractions over all combinations of regional draws, rather than matched pairs across regions; day-60 probabilities are 0.97, 0.98, 0.01 and 0.03. Neither these truths nor the other binary sets were repaired by the continuous rescore.
The historical pooled skill-oriented correlation $\rho(\mathrm{ECI},-\mathrm{loss})$ is $0.56$ ($n=23$, $p=0.005$), with $0.64$/$0.41$ by horizon; this is not a headline validation result. The set records 905 calls costing \$6.48. DeepSeek covers only three day-40 and one day-60 items at repetition 0, so its mean over answered items (0.0171203 bits) is omitted from tables comparing the full set; it ranks first on those same four items, including a repetition-0-only comparison. Kimi covers all eight items with 21 of 40 planned responses. The original answers and historical details are preserved.

\begin{figure}[htbp]
  \centering
  \includegraphics[width=\linewidth]{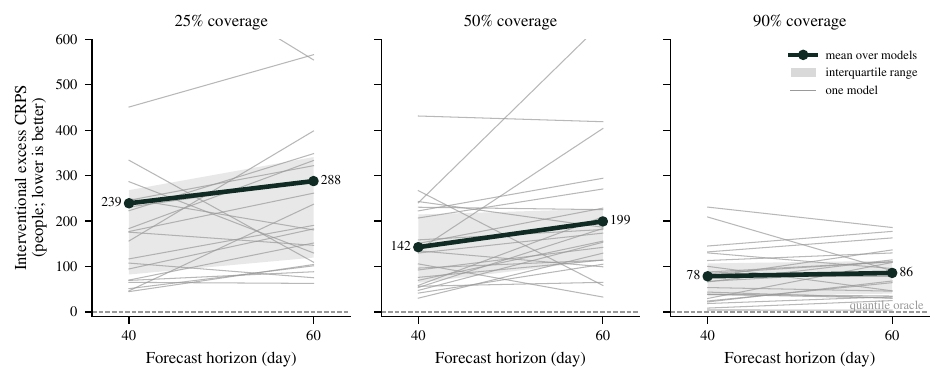}
  \caption{Interventional excess CRPS, in people, at day 40 and day 60 by vaccination coverage; lower is better. Thick line: mean over the 23 complete models; band: interquartile range; thin lines: one model each, three of which rise above the panel at 25 percent coverage and one at 50. Excess rises with horizon at every coverage: the mean goes from 239 to 288 at 25 percent, 142 to 199 at 50 and 78 to 86 at 90, and 16, 17 and 13 of 23 models do worse at day 60 than at day 40. DeepSeek V4 Flash is absent from all three panels.}
  \label{fig:starsim-horizon}
\end{figure}

\paragraph{Coverage, unconditional and horizon results.}
The revised skill-oriented interventional correlations with ECI at 25, 50 and 90 percent coverage are $\rho=0.70$ [$0.40,0.86$], $0.48$ [$0.07,0.76$] and $0.25$ [$-0.18,0.63$], using the model-panel intervals of \cref{app:starsim-rescore}.
In people, mean excess rises from 153 at day 40 to 191 at day 60, pooled over coverage levels, and 19 of 23 models do worse at day 60. At 50 percent coverage it rises from 142 to 199, with 17 of 23 worse (\cref{fig:starsim-horizon}). Unconditional continuous excess falls from 272 to 245 people, with 9 of 24 worse; normalized model means range from 0.227 to 4.363, and the skill-oriented ECI correlation is $0.75$ [$0.49,0.88$]. Original binary scores are unchanged: 14 of 24 models are within 0.05 bit of their original truth and the largest excess is 0.34 bit. These binary targets were not repaired by the continuous production run.

\Cref{tab:starsim-models,tab:starsim-horizon,tab:starsim-rung-horizon} report pooled scores and horizon breakdowns. Calls and costs count each retained response once (\cref{app:worlds-starsim}). Partial continuous scores carry a dagger; partial hold means are omitted from full-set comparisons (\cref{app:ablation-hold}). Mean interventional excess rises in people but falls after horizon normalization.

\begin{table}[htbp]
  \caption{Starsim scores per model, pooled over day 40 and day 60; lower is better in every score column. Binary bits: excess bits $\mathrm{KL}(P^*\,\|\,f)$ on the unconditional comparison of the two regions' cases (lower is better). Cont.\ excess: excess CRPS on unconditional new infections, divided by the horizon's constant (0 attains the floor at the scored quantiles). Interventional excess CRPS: the same for new infections under a vaccination of 25, 50 or 90 percent of Riverton, and pooled. Binary c90 bits: excess bits on the binary interventional set at 90 percent coverage. Hold bits: excess bits on the hold set (\cref{app:ablation-hold}). Calls and cost (OpenRouter USD) cover all five sets. $\dagger$: DeepSeek V4 Flash completed 21 of 24 continuous intervention items; its hold score is omitted because it covers only 4 of 8 items (3/4 at day 40, 1/4 at day 60). The September 16 forecasts, rescored for continuous outcomes only; binary truths are unchanged. The last two rows are the baseline and the SIR forecast (\cref{app:baselines}). The baseline is the climatological change in the unconditional continuous column and independence in the interventional columns. Scores lower than the baseline's are in bold, and continuous scores also lower than the SIR forecast's carry an asterisk.}
  \label{tab:starsim-models}
  \centering
  \scriptsize
\input{data/appendix_tables/starsim_models}
\end{table}

\begin{table}[htbp]
  \caption{Starsim scores per model at day 40 and day 60; lower is better. Columns as in \cref{tab:starsim-models}; Interv.\ excess pools the three coverage levels. The hold set is defined in \cref{app:ablation-hold}. The last two rows are the baseline and the SIR forecast at each horizon (\cref{app:baselines}). The baseline is the climatological change in the Cont.\ excess columns and independence in the Interv.\ excess columns. Scores lower than the baseline's are in bold, and continuous scores also lower than the SIR forecast's carry an asterisk.}
  \label{tab:starsim-horizon}
  \centering
  \scriptsize
\input{data/appendix_tables/starsim_horizon}
\end{table}

\begin{table}[htbp]
  \caption{Starsim interventional excess CRPS per model by coverage level and horizon, divided by the horizon's constant; lower is better. Each cell contains four items with three repetitions planned per item. We score the median of the available saved answers and report the mean score over items. $\dagger$: DeepSeek V4 Flash completed 3 of 4 items in three cells: day 60 at 25 percent coverage, day 40 at 50 percent and day 60 at 90 percent. The last row is the baseline, independence, at each coverage and horizon (\cref{app:baselines}); scores lower than the baseline's are in bold.}
  \label{tab:starsim-rung-horizon}
  \centering
  \small
\input{data/appendix_tables/starsim_rung_horizon}
\end{table}

\begin{figure}[htbp]
  \centering
  \includegraphics[width=\linewidth]{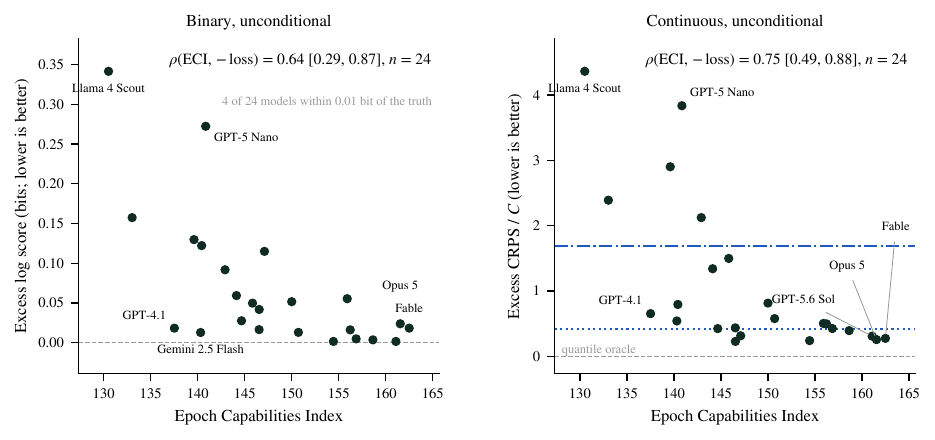}
  \caption{Unconditional Starsim questions against the Epoch Capabilities Index, $n=24$, pooled over horizons; lower is better in both panels. Left: binary comparison of the two regions' cumulative cases, excess bits (lower is better); 14 of 24 models are within 0.05 bit of the empirical truth; the visible ranking rule attains the clipped minimum. Right: continuous new-infections question, excess CRPS divided by the horizon's constant (0 attains the floor at the scored quantiles, dashed line); four models are above 2. The blue dash-dotted line is the climatological change and the blue dotted line the SIR forecast (\cref{app:baselines}). The continuous panel is retrospectively rescored; the binary panel is unrepaired. Spearman $\rho(\mathrm{ECI},-\mathrm{loss})$ has model-panel intervals (2,000 draws for continuous; the original 10,000 for binary); a positive $\rho$ means that the more capable models have the lower excess.}
  \label{fig:starsim-unconditional}
\end{figure}

\clearpage
\paragraph{Additional figures.}
\Cref{fig:starsim-unconditional} shows the two unconditional Starsim questions against ECI and \cref{fig:starsim-horizon} interventional excess at day 40 and day 60 by coverage; \cref{sec:starsim} quotes their numbers.

\Cref{tab:starsim-items} lists the 56 items, their transmission parameters, coverage, horizon, truth, empirical five-quantile floor and fixed normalization scale, plus the median, lowest and highest model excess on complete-item panels. Continuous truths and floors use the retrospective fixed-state rescore; binary values retain their original matched-seed construction. Median item difficulty is 0.55 at 25 percent coverage, 0.45 at 50 percent, 0.44 for unconditional continuous outcomes and 0.20 at 90 percent. These losses measure predictions of outcomes under each condition; they do not isolate skill in estimating the intervention effect.

\begin{table}[htbp]
  \caption{Starsim items and their difficulty using saved September 16 forecasts. Truth: original matched-seed probability (binary, unrepaired) or median new infections from 1,000 fixed-state continuations (continuous rescore). Floor: the CRPS of the ground-truth distribution's own quantiles, in people; $C$: the horizon's constant that divides the excess. Excess: excess bits (binary) or excess CRPS divided by $C$ (continuous), per item, over the models that answered every item of the set (24, or 23 where DeepSeek V4 Flash has missing items).}
  \label{tab:starsim-items}
  \centering
  \scriptsize
  \setlength{\tabcolsep}{3pt}
  \input{data/starsim/per_item_table}
\end{table}

\begin{table}[htbp]
  \caption{Retrospective Starsim continuous rescore and current combined-score sensitivity. Positive $\rho$ means better scores accompany higher external scores. Intervals from resampling models and replays are reported separately and hold different quantities fixed (\cref{app:starsim-rescore}); neither combines both sources of uncertainty. The nine-cell combined fit excludes all binary Starsim scores.}
  \label{tab:starsim-rescore-uncertainty}
  \centering\small
  \input{data/starsim/fixed_state_uncertainty}
\end{table}

\subsection{FreeCiv}
\label{app:full-freeciv}

FreeCiv uses the grouped run of September 20, 2026, and September 9 conditionals reused for 22 models, with two conditional reruns on September 20 (\cref{app:worlds}).

\Cref{tab:freeciv-models} lists the four main-text sets for the 24 models, pooled over horizons. It adds the excess nCRPS, which subtracts the five-quantile score at the replay distribution's own quantiles, and the natural-conditional gain, the turn-1 excess Brier score against $p(Y \mid X)$ minus the turn-2 score. The parsed share counts the answers we could read over the mid-range set, tail, mirror, continuous and natural-conditional sets, 1,800 items per model; it is 99.1\% or more for every model. The cost column is each model's OpenRouter charge for the calls behind its scored rows, \$174.15 over the 24 models (\cref{app:roster}). The gain is positive for 7 of 24 models, with mean $-0.002$ and range $-0.018$ to $0.016$. Gains compare forecasts against the same conditional truth; panel means weight models equally after averaging available cells. With forecasts and replay estimates fixed, 10,000 paired bootstrap resamples of the eight source worlds give a nominal 95\% interval $[-0.016,0.009]$ for mean gain; resampling whole model vectors separately gives $[-0.005,0.002]$. These summaries describe variation across a few selected worlds and a deliberately chosen model panel. They do not combine replay and world uncertainty or measure variation across repeated model responses.

\begin{table}[htbp]
  \caption{FreeCiv scores per model, pooled over horizons; lower is better in every score column except gain. Mid-range: excess Brier score $(f-q)^2$ on 750 questions. Tails: excess bits $\mathrm{KL}(q\,\|\,f)$ on 300 questions with $0 < q \le 0.05$. Continuous: nCRPS on 300 questions and its excess over the score at the replay distribution's own five quantiles. Natural conditional: turn-2 excess Brier score against $p(Y \mid X)$ on 400 cells, and the gain over keeping the turn-1 forecast. Parsed: readable answers over the five scored sets (1,800 items). Cost: OpenRouter USD for all of the model's calls. The last row is the baseline, the base rate for binary scores, the climatological change for continuous ones and independence for natural conditionals (\cref{app:baselines}); scores lower than the baseline's are in bold.}
  \label{tab:freeciv-models}
  \centering
  \scriptsize
\input{data/appendix_tables/freeciv_models}
\end{table}

\paragraph{Mirrors.}
The mirror set is a check on the top end of the probability scale that the main text does not use. It holds 50 binary questions with $0.95 \le q < 1$, ten per horizon, drawn from the binary templates; the mean $q$ is \FCMirMeanQ. 
\Cref{tab:freeciv-mirrors} gives each model's excess Brier score, pooled and by horizon. Excess Brier runs from \FCMirBest{} (\FCMirBestModel) to \FCMirWorst (\FCMirWorstModel). A constant forecast of 0.5 has an excess of \FCMirFlat{} on this set, and no model scores below it: models given grouped prompts assign low probabilities to these near-certain events, as they do on the mid-range set (\cref{app:ablation-grouping}). The skill-oriented Spearman correlation $\rho(\mathrm{ECI},-\mathrm{loss})$ is \FCMirRho{} [\FCMirRhoLo, \FCMirRhoHi], $p = \FCMirRhoP$, with the model bootstrap of \cref{sec:validation}.

\begin{table}[htbp]
  \caption{FreeCiv mirror set: excess Brier score per model on 50 questions with $0.95 \le q < 1$, pooled and by resolution turn (10 questions each); lower is better. A constant 0.5 scores \FCMirFlat. The last row is the base rate (\cref{app:baselines}); scores lower than the base rate's are in bold.}
  \label{tab:freeciv-mirrors}
  \centering
  \scriptsize
\input{data/appendix_tables/freeciv_mirrors}
\end{table}

\begin{figure}[htbp]
  \centering
  \includegraphics[width=\linewidth]{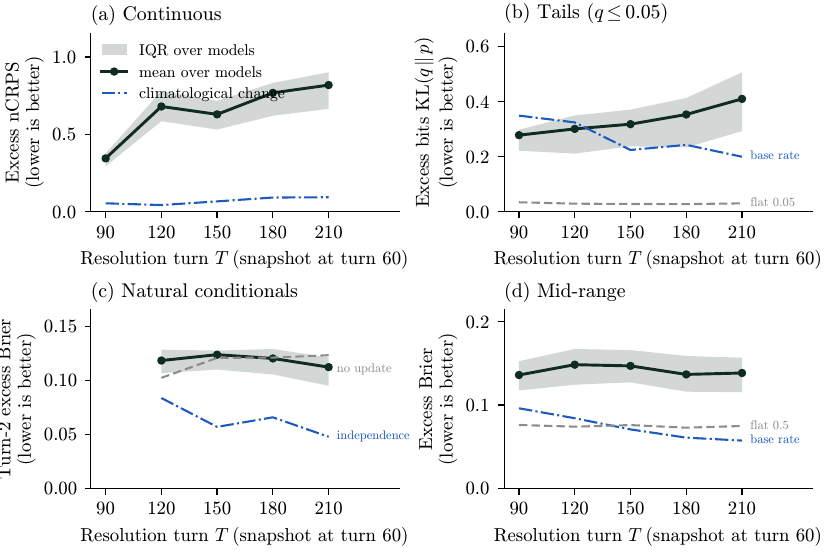}
  \caption{FreeCiv scores by resolution turn $T$ from the turn-60 snapshot. The line is the mean over the 24 models and the band the interquartile range across models; lower is better. Dashed gray lines are reference forecasters: a constant 0.05 on tail questions, the unchanged turn-1 forecast on natural conditionals, and a constant 0.5 on the mid-range set. Mean excess nCRPS goes from 0.35 at $T = 90$ to 0.82 at $T = 210$, mean tail bits from \FCTailHorizonFirst{} to \FCTailHorizonLast, the mid-range excess Brier score from \FCMidHorizonFirst{} to \FCMidHorizonLast, and the natural-conditional excess Brier score from 0.118 at $T = 120$ to 0.112. Natural conditionals exist only for $T \ge 120$. The blue dash-dotted line is the baseline: the base rate on mid-range and tail questions, the climatological change on continuous ones and independence on natural conditionals (\cref{app:baselines}).}
  \label{fig:freeciv-horizon}
\end{figure}

\paragraph{By horizon.}
\Cref{tab:freeciv-horizon-binary} gives the mid-range excess Brier score and the tail excess bits at each resolution turn, and \cref{tab:freeciv-horizon-cont} the continuous nCRPS and the natural-conditional turn-2 excess Brier score. Across the 24 models the mean mid-range score stays between \FCMidHorizonLo{} and \FCMidHorizonHi{} over the five horizons, and the mean natural-conditional score between 0.112 and 0.124 over its four. The mean tail bits rise from \FCTailHorizonFirst{} at turn 90 to \FCTailHorizonLast{} at turn 210, and the mean nCRPS from 0.41 to 1.03. \Cref{fig:freeciv-horizon} plots the excess nCRPS instead, which subtracts the floor and rises from 0.35 to 0.82 over the same turns.

\begin{table}[htbp]
  \caption{FreeCiv mid-range excess Brier score (150 questions per turn) and tail excess bits (60 per turn) per model by resolution turn; lower is better. The last row is the base rate at each turn (\cref{app:baselines}); scores lower than the base rate's are in bold.}
  \label{tab:freeciv-horizon-binary}
  \centering
  \scriptsize
\input{data/appendix_tables/freeciv_horizon_binary}
\end{table}

\begin{table}[htbp]
  \caption{FreeCiv continuous nCRPS (60 questions per turn) and natural-conditional turn-2 excess Brier score (100 cells per turn) per model by resolution turn; lower is better. Natural conditionals exist only for turns 120 to 210. The last row is the baseline at each turn (\cref{app:baselines}): the climatological change, in nCRPS, \changed{which no model beats}, and independence for natural conditionals; scores lower than the baseline's are in bold.}
  \label{tab:freeciv-horizon-cont}
  \centering
  \scriptsize
\input{data/appendix_tables/freeciv_horizon_cont}
\end{table}

\paragraph{Natural conditionals by structural group.}
\Cref{tab:freeciv-natcond-blocks} splits the 400 cells by the five structural groups of \cref{sec:freeciv-setup}: A, irrelevant news (36 cells); B, related weak news (36); C1, mechanical same-series (56); C2, persistence across series (172); and D, inferential (100). The gain over not updating is negative for 23 of 24 models in group C1, where the mean gain is $-0.053$. In the other four groups the mean gain lies between 0.003 and 0.010, and it is negative for 3, 9, 9 and 4 models in groups A, B, C2 and D. The other 344 cells have mean gain $+0.006$. The mean gain in C1 remains negative when any one world is left out; the source-world interval for the non-C1 mean includes zero. Positive means are descriptive.

Using an export of the aligned replays \changednew{in the data release}, we computed 95\% percentile intervals from 100,000 resamples of 1,003 clusters defined by random-number-generator (RNG) tags shared across worlds. A sampled tag is counted the same number of times wherever it occurs. We match tags to each world's replay columns by identity, not by column position. The resampling includes all outcomes, including $X=\mathrm{false}$; unconditional probabilities divide by the number of sampled replays in that world, and conditional probabilities divide by the number in which $X$ is true. All conditional denominators stayed positive. We keep the 9,584 forecast pairs, 16 exclusions for missing pairs and score weights fixed: we average the available cells within each model and group, then give the 24 models equal weight. Replay-only gain intervals are $[-0.004,-0.000]$ overall, $[-0.059,-0.047]$ for C1 and $[+0.005,+0.008]$ for the other 344 cells. Independently resampling 1,000 columns within each world (also 100,000 draws) gives similar results. The overall upper endpoint is extremely close to zero; its nominal exclusion of zero should not determine the substantive conclusion. These intervals condition on the selected cells, worlds and observed forecasts; they do not account for variation across source worlds, model panels or repeated elicitation, or uncertainty from question selection. Distinct tags are treated as independent; dependence beyond shared tags is not established. We do not repeat the original selection by probability band or the substitutions used to construct the corpus.

Replay uncertainty further limits interpretation of movement. Reconstructing each cell's joint reveal/outcome counts and resampling that table 20,000 times preserves the dependence between its unconditional and conditional estimates. Cellwise paired 95\% percentile intervals exclude zero for 228 of 400 shifts; 152 of the 174 estimated absolute shifts below 0.03 remain unresolved. These are unadjusted cellwise intervals, and comparisons based on the estimated shift or on whether its interval excludes zero are exploratory. C1 losses persist among cells whose intervals exclude zero. Aligned replay identities now permit the joint gain analysis above; the cellwise movement intervals remain a separate diagnostic. Movement relative to the estimated shift is also distinct from gain: writing $d=f_2-f_1$ and $\Delta=q_c-p$, where $q_c=p(Y\mid X)$ and $p=p(Y)$, gain equals $2d(p-f_1)+2d\Delta-d^2$. Thus a movement poorly matched to the shift can still improve the score by correcting initial forecast error. With one elicitation per condition, imprecision across cells does not tell us how much responses would vary if the model were asked again.

\begin{table}[htbp]
  \caption{FreeCiv natural conditionals per model by structural group: turn-2 excess Brier score against $p(Y \mid X)$ (lower is better) and gain over keeping the turn-1 forecast (higher is better). Groups: A irrelevant news (36 cells), B related weak news (36), C1 mechanical same-series (56), C2 persistence across series (172), D inferential (100). The last row is independence, the base rate of $Y$, in each group (\cref{app:baselines}); its gain is zero. Scores lower than independence's are in bold.}
  \label{tab:freeciv-natcond-blocks}
  \centering
  \scriptsize
\input{data/appendix_tables/freeciv_natcond_blocks}
\end{table}

\begin{figure}[htbp]
  \centering
  \includegraphics[width=3.2in]{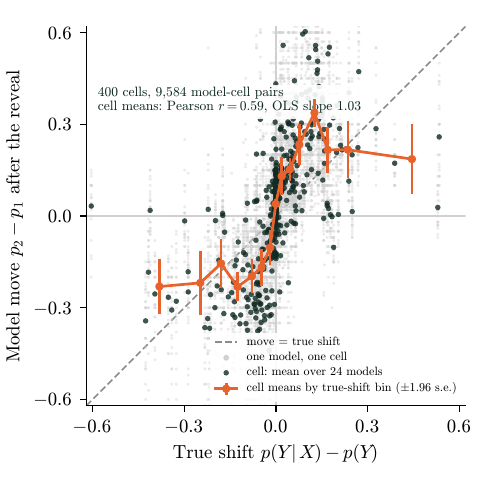}
  \caption{Natural conditionals: how forecasts move after one revealed fact. Each green point is one of the 400 cells: $x$ is the estimated replay shift $p(Y \mid X) - p(Y)$, with the conditional probability using only replays satisfying the reveal, and $y$ is the mean move $p_2 - p_1$ over available models. Gray points are the 9,584 model--cell pairs. Orange is the mean of cell means by bin of estimated shift, with 1.96 standard-error bars; these bars do not represent paired replay uncertainty. The dashed line marks a move equal to the estimated shift. Cell means track the estimated shift (Pearson $r = 0.59$, least-squares slope 1.03). Mean absolute movement over the 400 cell means is 0.17, 2.3 times the mean absolute estimated shift of 0.07; over model--cell pairs it is 0.18. The 2.3 ratio compares mean absolute magnitudes; it does not estimate a coefficient of proportional overshooting. Pooling the groups obscures their differences, and the replay shifts are themselves uncertain: C1 has negative mean gain, while the other groups have positive descriptive means (\cref{tab:freeciv-natcond-blocks}).}
  \label{fig:freeciv-natcond}
\end{figure}

\paragraph{Additional figures.}
\Cref{fig:freeciv-horizon} shows each FreeCiv score by resolution turn and \cref{fig:freeciv-natcond} the turn-2 moves on the natural-conditional cells against the estimated replay shift.

\begin{figure}[htbp]
  \centering
  \includegraphics[width=3.0in]{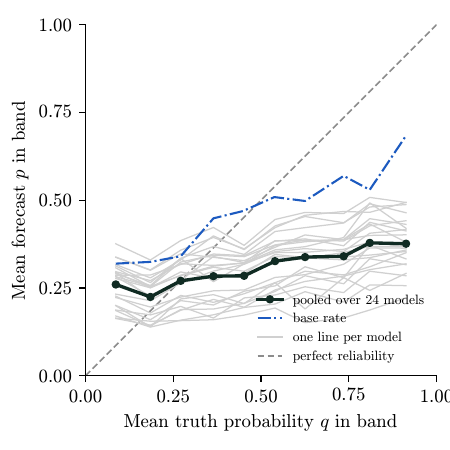}
  \caption{Mean FreeCiv mid-range forecast by replay-truth bin over 750 questions: mean forecast $f$ against mean ground-truth probability $q$ in ten equal-width intervals of $q$ on $[0.05, 0.95]$. One gray line per model, green the pool over 24 models, dashed the identity line; its distance from a truth-binned curve is not a calibration decomposition. Pooled forecasts run from \FCMidRelPLow{} in the lowest interval ($q = \FCMidRelQLow$) to \FCMidRelPHigh{} in the highest ($q = \FCMidRelQHigh$); \FCMidRelWhere. The question-level mean of $f - q$ is negative for \FCMidNegBias{} of 24 models (range \FCMidBiasMin{} to \FCMidBiasMax). Blue dash-dotted: the mean base rate of the questions in each interval (\cref{app:baselines}).}
  \label{fig:freeciv-reliability}
\end{figure}

\paragraph{Truth-conditioned forecasts and updating.}
\Cref{fig:freeciv-reliability} plots mean mid-range forecasts by replay-truth bin; it is not a forecast-binned reliability diagram, and \cref{fig:freeciv-natcond} above plots the move of each natural-conditional forecast against the estimated replay shift. Discrimination, the Pearson correlation between a model's forecasts and the truth over the 750 mid-range questions, runs from \FCMidDiscMin{} (\FCMidDiscMinModel) to \FCMidDiscMax{} (\FCMidDiscMaxModel) and correlates with ECI at $\rho = \FCMidDiscRho$ [\FCMidDiscLo, \FCMidDiscHi] (model bootstrap, 10,000 resamples); the signed question-level mean of $f - q$ correlates with ECI at \FCMidBiasEciRho, and the excess Brier score correlates with the absolute bias at \FCMidExcessAbsBiasRho{} and with discrimination at \FCMidExcessDiscRho. Comparative questions, which ask whether one civilization exceeds another, show no bias (mean $f - q$ of \FCMidCompBias) and the highest discrimination (\FCMidCompDisc); event and threshold questions are under-forecast by \FCMidUnderLo{} to \FCMidUnderHi (\cref{tab:freeciv-family-difficulty}). On the natural conditionals the turn-2 excess Brier score of a model correlates at 0.91 with the standard deviation of its turn-2 forecasts (0.21 for o3 to 0.35 for Gemini 2.5 Flash) and at -0.26 with its discrimination at turn 2 (0.16 to 0.33), which itself rises with ECI at 0.73.

\paragraph{Joint binary references and composition.}\label{app:freeciv-baselines}
Across the 1,100 selected binary questions, equally weighted per item (750 mid-range, 300 tail, 50 mirror), mean excess Brier is 0.135. Universal 0.5 scores 0.125, and nine of 24 models beat it. A leave-one-world-out ridge reference using only question-text features scores 0.095; it learns from replay labels in the other seven worlds, information the LLMs were not supplied. It is a supervised reference, not a matched zero-shot baseline or report-conditional oracle. All 24 model point estimates are worse under these archive weights. The fixed penalty is 10; no held-out band labels are inputs. \changed{The base rate of \cref{app:baselines}, which also learns from the other games' replays, scores \BLFCSelected{} on the 1,100 questions; no model beats it (best \BLFCSelectedBest).}

Under archive weights, grouping changes mean excess Brier from 0.138 to 0.135; equal mass for each band reverses the change (0.186 to 0.207). The two weighting schemes measure different averages. Neither describes the full population of candidate events before selection, and the improvement does not hold under both weighting schemes. Reused worlds, overlapping training folds, replay estimation and model generation limit uncertainty claims about the supervised reference.

Forecast-binned diagnostics on the selected mid-range set also differ from \cref{fig:freeciv-reliability}: pooled rows with $f<0.1$ average $f=\FCMidLowF$, $q=\FCMidLowQ$ (\FCMidLowRows{} rows), while $f\ge0.9$ averages $f=\FCMidHighF$, $q=\FCMidHighQ$ (\FCMidHighRows{} rows). These repeated observations of models on the same questions describe discrepancies in the selected archive; they are not independent evidence of miscalibration in deployment. Mean mid-range excess decomposes as $\FCMidDecTotal=\FCMidDecBias+\FCMidDecVarF+\FCMidDecVarQ-\FCMidDecCov$ (squared bias, forecast variance, truth variance, minus twice covariance); squared bias is about \FCMidDecBiasShare, not all of the error.

\paragraph{Selection weights.}\label{app:freeciv-weights}
The draw takes 30 mid-range questions from each band of $q$ at each horizon (\cref{app:worlds-freeciv}), although the candidates cluster near the ends of the mid-range. Selecting on $q$, the quantity being forecast, could penalize good forecasts. Consider a toy model: suppose each world's event probability $q$ is drawn from a density $p(q)$ and the world report says nothing about $q$, so the best forecast is the mean $\bar q$ under $p$. If the draw selects worlds so as to spread $q$ evenly over $[0,1]$, a forecaster that knows $p$ but not the selection has expected excess Brier score $1/12+(\bar q-1/2)^2$ and loses to a constant $1/2$ unless $\bar q=1/2$. Note that selecting on parameters that are given to the forecaster with the question, such as family or horizon, would not have such an effect. \changed{The base rate over all candidates (\cref{app:baselines}) is such a forecaster for each template and horizon, but it slightly outperforms a constant 0.5 on the mid-range set (\BLFCMid{} against \BLFCMidConst).}

To assess sensitivity to selection within each set, we weight scores by inverse group selection fractions. A mid-range question's weight is the number of candidates in its band, horizon and family divided by the number selected, rescaled so that each family keeps its share of the questions at each horizon. This rebalances the sampled probability bands within each family and horizon while retaining the draw's family mix. Because the draw also uses caps, these fractions need not equal individual inclusion probabilities. We do not reweight the tail and mirror sets, whose within-family draws are not stratified by $q$. The split into sets by $q$ is itself a selection that no weighting within a set undoes, so each set's scores are conditional on it.

The weights leave the mean score, the correlation with ECI and the ranking of models nearly unchanged (\cref{tab:freeciv-selection-weights}). Statistics that depend on the mix of $q$ move more: the constant 0.5, which gains from the even spread of $q$, scores worse but still beats every model; squared bias is a smaller share of the excess; and discrimination correlates less with ECI. We therefore weight questions equally elsewhere.

\begin{table}[htbp]
  \caption{FreeCiv mid-range results over the 750 questions and 24 models with equal weights, as in the main text, and with the selection weights of \cref{app:freeciv-weights}. $\rho$ is sign-adjusted Spearman with its 95\% percentile bootstrap over models.}
  \label{tab:freeciv-selection-weights}
  \centering
  \small
\input{data/appendix_tables/freeciv_selection_weights}
\end{table}

\paragraph{Difficulty by question family.}
\Cref{tab:freeciv-family-difficulty} gives the mean score of each family at each horizon, pooled over the 24 models. In the mid-range set the three worst families (any city destroyed, the number of government changes, and cities lost) hold \FCMidWorstItems{} of 750 items and \FCMidWorstShare{} of the pooled excess Brier score. In the tail set the three worst (a government in place at the horizon, a peace in effect, a wonder completed) hold \FCTailWorstItems{} of 300 items and \FCTailWorstShare{} of the pooled bits. In the continuous set the two technology-count families and world wonders hold 82 of 300 items and 58\% of the pooled raw normalized CRPS. This is not their share of excess loss or evidence that these families drive its ECI association.

\begin{table}[htbp]
  \caption{FreeCiv difficulty by question family and horizon, pooled over the 24 models: mean excess Brier score (mid-range), mean excess bits (tail set) and mean normalized CRPS (continuous set); lower is easier. Families are sorted from hardest to easiest by the item-weighted mean over horizons.}
  \label{tab:freeciv-family-difficulty}
  \centering
  \scriptsize
  \setlength{\tabcolsep}{3.5pt}
\input{data/appendix_tables/freeciv_family_difficulty}
\end{table}

\subsection{Combined score}
\label{app:combined}

\begin{figure}[htbp]
  \centering
  \includegraphics[width=0.6\linewidth]{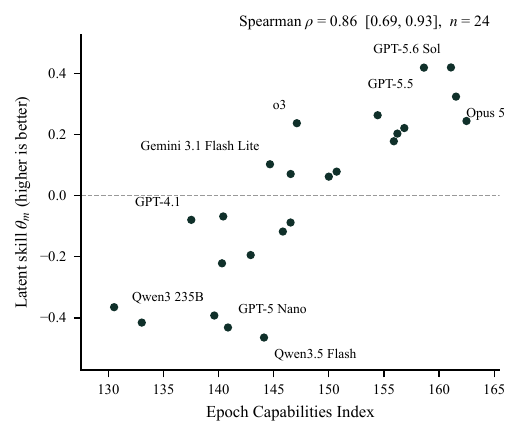}
  \caption{Latent skill $\theta_m$ against the Epoch Capabilities Index. Brackets: 95\% percentile bootstrap over models, 10,000 resamples, seed 2026.}
  \label{fig:combined-score}
\end{figure}

\Cref{tab:combined-score} lists the latent skill $\theta_m$ of \cref{sec:validation} for every model, and \cref{fig:combined-score} plots it against ECI. The logarithm lets cells with different loss scales share one axis. The correlation with ECI is \CombinedRho{} [\CombinedLo, \CombinedHi]. ECI is itself a latent-ability fit across benchmarks \citep{ho2025rosetta}, and this fit gives each observed log-loss cell equal nominal weight, not equal influence, so it is a descriptive summary of the nine score sets, not evidence of a single common latent skill. The fit is $\log \ell_{mc} = \delta_c - \theta_m + \varepsilon_{mc}$ by alternating least squares, omitting the one missing cell (DeepSeek V4 Flash on the Starsim interventional set) from the sums and $\theta_m$ centered at zero. Losses are floored at $10^{-4}$ before the logarithm; no cell reaches the floor. Every cell is an excess score. The residual standard deviation on the log scale is \CombinedResidual{}. The agreement between $\theta_m$ and each cell's own ranking, as a Spearman correlation, is \CombinedCoreAgreementLo{} to \CombinedCoreAgreementHi{} for the Micropolis and Starsim cells and \CombinedFreeAgreementLo{} to \CombinedFreeAgreementHi{} for the FreeCiv cells, across sets. Leaving out any one cell moves the correlation with ECI between \CombinedLooLo{} and \CombinedLooHi{}. This fit excludes historical Starsim binary. \Cref{tab:starsim-rescore-uncertainty} reports its ForecastBench association on 17 models, with separate model-panel and replay intervals.

\paragraph{Binary exclusion and panel sensitivity.}
The primary nine-cell fit keeps the 24-model roster and omits the one missing intervention cell from its sums. Using repaired continuous scores throughout, the ECI association is \CombinedRhoThree; refitting the 23 models with all cells gives \CombinedCompleteRho, and restricting to the 22 models with all planned Starsim repetitions gives \CombinedRepsRho. The ten-cell historical-binary sensitivity gives \CombinedBinaryRho{} on 24 models. Exclusion is motivated by the visible fixed rule, not by the direction of these associations. These fits use different model panels and are not interchangeable; missing responses are not imputed. Paired bootstrap intervals over models for the change after removing binary scores include zero in the archived audit and do not include missing-response or replay uncertainty. Audit inputs and sensitivities are retained separately \changednew{and available on request}.

\paragraph{Matched external panel and influence.}
On the same 17 models with ForecastBench scores, the ECI association is \CombinedMatchedEciRho{} rather than the 24-model \CombinedRhoThree. Six of the eight highest-ECI models lack ForecastBench scores. \Cref{tab:combined-matched} reports both external associations and the partial correlation with ForecastBench given ECI, computed as in \cref{app:validation-forecastbench}, using 10,000 shared model resamples, seed 2026, with fitted scores held fixed. Once more, if ECI is accounted for, the combined score shows no association with ForecastBench. Its interval is wide, so this is insufficient evidence of an association, not evidence of its absence.

Leaving one cell out keeps aggregate ECI correlation within \CombinedLooLo--\CombinedLooHi\ but moves individual ranks by up to \CombinedLooMaxMove{} places. Cell log-loss standard deviations range from \CombinedCellSdLo{} to \CombinedCellSdHi, so nominally equal weights do not give equal influence. Equal total weight per simulation gives ECI $\rho=\CombinedEqualSimRho$, rank agreement \CombinedEqualSimAgreement{} with the primary fit and maximum movement of \CombinedEqualSimMaxMove{} places. These are exploratory aggregate sensitivities. FreeCiv mid-range and tail ranks correlate \FreeCivMidTailRankCorr, underscoring the limits of a common-skill interpretation.

\begin{table}[htbp]
  \caption{Nine-cell combined score for the same 17 models with both external benchmark scores. Intervals resample the same model indices for all three correlations; ``ForecastBench, given ECI'' is the partial Spearman correlation with ECI partialled out of both sides; they omit replay and external-score measurement uncertainty.}
  \label{tab:combined-matched}
  \centering\small
  \input{data/combined_matched_panel}
\end{table}

\begin{table}[htbp]
  \caption{Latent skill $\theta_m$ per model from the additive log-loss model, centered at zero, higher is better. Sorted by $\theta_m$.}
  \label{tab:combined-score}
  \centering
  \small
\input{data/combined_score_table}
\end{table}

\begin{table}[htbp]
  \caption{Latent skill $\theta_m$ from the combined score and each model's rank on every cell, where rank 1 is the lowest loss and tied models share the average rank. ``Avg Rank'' ranks the mean of a model's cell ranks. Cell shading runs from green for the best value in a column, through white, to purple for the worst. DeepSeek V4 Flash has no Starsim interventional score; the other 23 models are ranked on that set. Its ``Avg Rank'' uses eight available cell ranks; its $\theta_m$ is fitted using eight available loss cells. Sorted by $\theta_m$.}
  \label{tab:rank-table}
  \centering
  \footnotesize
\input{data/rank_table}
\end{table}

\begin{table}[htbp]
  \caption{Spearman correlation between the model rankings on $\theta_m$ and on each cell of \cref{tab:rank-table}, and between every pair of cells. Every ranking puts the best model first, so a positive value means two rankings agree. Correlations with the Starsim interventional set use the 23 models that have it.}
  \label{tab:rank-correlation}
  \centering
  \footnotesize
\input{data/rank_correlation_table}
\end{table}

\paragraph{Simulator-level combined scores.}
Fitting the same additive model to the cells of one simulator at a time gives a combined score for each simulator. \Cref{tab:simulator-correlation} gives the Spearman correlations of the three scores with each other and with ECI, and the partial correlation of each pair of simulators given ECI, computed as in \cref{app:validation-forecastbench}, each with a 95\% percentile bootstrap over the 24 models. The three correlate with one another at \SimScoreRhoLo{} to \SimScoreRhoHi{} showing a high degree of consistency between them; no interval includes zero. Given ECI, the pairwise correlations fall to \SimScorePartialLo{} to \SimScorePartialHi{} and every interval includes zero: most of the agreement between simulators is shared with ECI.

\begin{table}[htbp]
  \caption{Spearman correlations between ECI and the three simulator-level combined scores, and partial Spearman correlations between the simulator-level scores given ECI. CI: 95\% percentile bootstrap over models.}
  \label{tab:simulator-correlation}
  \centering
  \small
\input{data/simulator_correlation_table}
\end{table}

\subsection{Held-out worlds and simulators}
\label{app:generalization}
All replays come from 27 worlds (15 Micropolis cities, 8 FreeCiv games, 4 Starsim scenarios). Replays of a world estimate the outcome distribution of its questions; they are not independent evaluation settings, and our intervals resample models, not replays. We therefore test whether the model ranking depends on which worlds and simulators we used. \emph{Half vs half}: we split the worlds into two disjoint halves, compute every cell from each half's questions alone, fit $\theta_m$ on each half, and correlate the two fits (\cref{tab:generalization}a). Over \GenHalfSplits{} random splits of all three simulators' worlds, the two halves agree at $\rho=\GenHalfRho$ on average and at least \GenHalfMin{} in every split. Within each simulator, over every distinct split, the simulator-level scores of \cref{tab:simulator-correlation} agree at \GenHalfMicropolis{} (Micropolis), \GenHalfFreeCiv{} (FreeCiv) and \GenHalfStarsim{} (Starsim). Each fit uses only the questions from its half of the worlds; this check does not establish agreement on new worlds. \emph{Leave one simulator out}: $\theta_m$ fitted on the cells of two simulators agrees with the held-out simulator's combined score at $\rho=\GenLosoLo$--$\GenLosoHi$ (\cref{tab:generalization}b).

\begin{table}[htbp]
  \caption{Generalisation across worlds and simulators, Spearman $\rho$ over 24 models. (a) $\theta_m$ fitted separately on two disjoint halves of the worlds. ``All three'' splits every simulator's worlds at once (Micropolis + FreeCiv + Starsim), fits all nine cells, and draws \GenHalfSplits{} random splits, seed 2026; the per-simulator rows fit that simulator's cells and use every distinct split. (b) $\theta_m$ fitted on two simulators' cells against the held-out simulator's combined score; 95\% CI: percentile bootstrap over models, 10,000 resamples, seed 2026, fitted scores held fixed.}
  \label{tab:generalization}
  \centering
  \small
\input{data/generalization_table}
\end{table}

%% file: data/appendix_tables/binary_baselines.tex
\setlength{\tabcolsep}{4pt}
\begin{tabular}{lrrrr}
\toprule
 & \multicolumn{2}{c}{Micropolis} & \multicolumn{2}{c}{FreeCiv} \\ \cmidrule(lr){2-3} \cmidrule(lr){4-5}
Base rate & Mid-range & Tail & Mid-range & Tail \\
\midrule
Constant 0.5 (mid-range) or 0.05 (tail)$^\dagger$ & 0.120 (23) & 0.035 (0) & 0.075 (0) & 0.030 (0) \\
One rate per simulation & 0.079 (20) & 0.214 (20) & 0.075 (0) & 0.816 (24) \\
Per simulation and horizon & 0.079 (20) & 0.212 (20) & 0.076 (0) & 0.820 (24) \\
Per question & 0.070 (16) & 0.128 (17) & 0.079 (0) & 0.340 (12) \\
\textbf{Per question and horizon (used)} & 0.071 (17) & 0.124 (16) & 0.074 (0) & 0.268 (9) \\
\quad including the forecast world & 0.065 (12) & 0.115 (14) & 0.068 (0) & 0.254 (8) \\
\quad within the question's set$^\dagger$ & 0.050 (5) & 0.005 (0) & 0.063 (0) & 0.008 (0) \\
\quad FreeCiv family instead of template & -- & -- & 0.108 (1) & 0.475 (23) \\
\quad FreeCiv asked questions instead of candidates & -- & -- & 0.086 (0) & 0.304 (11) \\
\bottomrule
\end{tabular}

%% file: data/appendix_tables/continuous_baselines.tex
\setlength{\tabcolsep}{4pt}
\begin{tabular}{lrrr}
\toprule
Forecast & Micropolis & FreeCiv & Starsim \\
\midrule
\textbf{Climatological change (used)} & 0.037 (11) & 0.070 (0) & 1.68 (19) \\
Persistence & 0.059 (23) & 1.36 (24) & 8.49 (24) \\
Other worlds' values instead of changes & 0.115 (24) & 0.071 (0) & -- \\
SIR forecast & -- & -- & 0.414 (7) \\
\bottomrule
\end{tabular}

%% file: data/appendix_tables/micropolis_models.tex
\setlength{\tabcolsep}{3pt}
\begin{tabular}{lrrrrrr}
\toprule
Model & ECI & Excess Brier & Excess bits & Excess nCRPS & \multicolumn{2}{c}{Parsed (\%)} \\
\cmidrule(lr){6-7}
 & & Mid-range & Tail & & Binary & Continuous \\
\midrule
Claude Fable & 162.5 & \textbf{0.0521} & \textbf{0.0837} & \textbf{0.0268} & 100.0 & 100.0 \\
Claude Opus 5 & 161.5 & \textbf{0.0447} & \textbf{0.0747} & \textbf{0.0289} & 100.0 & 100.0 \\
GPT-5.6 Sol & 161.1 & \textbf{0.0428} & \textbf{0.0664} & \textbf{0.0286} & 100.0 & 100.0 \\
GPT-5.5 & 158.6 & \textbf{0.0451} & \textbf{0.0481} & \textbf{0.0284} & 100.0 & 100.0 \\
Gemini 3.7 Flash & 156.8 & \textbf{0.0455} & \textbf{0.0494} & \textbf{0.0303} & 100.0 & 100.0 \\
GPT-5.6 Luna & 156.2 & \textbf{0.0555} & \textbf{0.1055} & \textbf{0.0334} & 100.0 & 100.0 \\
Claude Sonnet 5 & 155.9 & \textbf{0.0522} & \textbf{0.1048} & \textbf{0.0300} & 100.0 & 100.0 \\
DeepSeek V4 Flash 0731 & 154.4 & \textbf{0.0441} & \textbf{0.0698} & 0.0374 & 100.0 & 100.0 \\
Gemini 3 Flash Preview & 150.7 & \textbf{0.0518} & \textbf{0.1214} & \textbf{0.0369} & 100.0 & 100.0 \\
GPT-5 & 150.0 & \textbf{0.0522} & \textbf{0.0980} & \textbf{0.0327} & 100.0 & 100.0 \\
o3 & 147.1 & \textbf{0.0502} & \textbf{0.1017} & \textbf{0.0358} & 100.0 & 100.0 \\
GPT-5.4 Nano & 146.5 & \textbf{0.0683} & \textbf{0.1224} & 0.0411 & 100.0 & 100.0 \\
o4 Mini & 146.5 & 0.0817 & 0.1250 & 0.0419 & 100.0 & 100.0 \\
GPT-5 Mini & 145.8 & \textbf{0.0711} & \textbf{0.0939} & 0.0420 & 100.0 & 99.8 \\
Gemini 3.1 Flash Lite Preview & 144.7 & \textbf{0.0637} & \textbf{0.1045} & \textbf{0.0368} & 100.0 & 100.0 \\
Qwen3.5-Flash & 144.1 & 0.1345 & 0.5989 & 0.0514 & 100.0 & 96.7 \\
Claude Haiku 4.5 & 142.9 & \textbf{0.0682} & 0.1508 & 0.0412 & 100.0 & 100.0 \\
GPT-5 Nano & 140.8 & 0.0875 & \textbf{0.1069} & 0.0711 & 100.0 & 92.0 \\
Kimi K2 & 140.4 & \textbf{0.0661} & 0.1683 & 0.0417 & 98.4 & 98.3 \\
Gemini 2.5 Flash & 140.3 & 0.1020 & 0.2673 & 0.0411 & 100.0 & 99.8 \\
Qwen3 235B A22B & 139.6 & \textbf{0.0696} & 0.2331 & 0.0464 & 100.0 & 100.0 \\
GPT-4.1 & 137.5 & 0.0764 & \textbf{0.1007} & 0.0415 & 100.0 & 100.0 \\
DeepSeek V3 & 133.0 & 0.0727 & 0.2738 & 0.0379 & 100.0 & 99.8 \\
Llama 4 Scout & 130.5 & 0.0762 & 0.1321 & 0.0428 & 100.0 & 100.0 \\
\midrule
Baseline & & 0.0713 & 0.1240 & 0.0370 & & \\
\bottomrule
\end{tabular}

%% file: data/appendix_tables/micropolis_horizon.tex
\setlength{\tabcolsep}{3pt}
\begin{tabular}{lrrrrrrrrr}
\toprule
Model & ECI & \multicolumn{4}{c}{Mid-range excess Brier} & \multicolumn{4}{c}{Tail excess bits} \\
\cmidrule(lr){3-6}\cmidrule(lr){7-10}
 & & 3 & 5 & 7 & 10 & 3 & 5 & 7 & 10 \\
\midrule
Claude Fable & 162.5 & \textbf{0.0583} & \textbf{0.0572} & \textbf{0.0467} & \textbf{0.0499} & \textbf{0.0871} & \textbf{0.0874} & \textbf{0.0743} & \textbf{0.0849} \\
Claude Opus 5 & 161.5 & \textbf{0.0489} & \textbf{0.0489} & \textbf{0.0404} & \textbf{0.0434} & \textbf{0.0913} & \textbf{0.0890} & \textbf{0.0553} & \textbf{0.0567} \\
GPT-5.6 Sol & 161.1 & \textbf{0.0534} & \textbf{0.0480} & \textbf{0.0377} & \textbf{0.0379} & \textbf{0.0708} & \textbf{0.0816} & \textbf{0.0518} & \textbf{0.0578} \\
GPT-5.5 & 158.6 & \textbf{0.0563} & \textbf{0.0477} & \textbf{0.0421} & \textbf{0.0399} & \textbf{0.0614} & \textbf{0.0608} & \textbf{0.0294} & \textbf{0.0356} \\
Gemini 3.7 Flash & 156.8 & \textbf{0.0571} & \textbf{0.0535} & \textbf{0.0399} & \textbf{0.0385} & \textbf{0.0677} & \textbf{0.0522} & \textbf{0.0356} & \textbf{0.0368} \\
GPT-5.6 Luna & 156.2 & \textbf{0.0602} & \textbf{0.0561} & \textbf{0.0512} & \textbf{0.0564} & \textbf{0.0965} & \textbf{0.1207} & \textbf{0.0882} & 0.1182 \\
Claude Sonnet 5 & 155.9 & \textbf{0.0563} & \textbf{0.0575} & \textbf{0.0487} & \textbf{0.0492} & \textbf{0.1151} & \textbf{0.1172} & \textbf{0.0880} & 0.0946 \\
DeepSeek V4 Flash 0731 & 154.4 & \textbf{0.0514} & \textbf{0.0485} & \textbf{0.0395} & \textbf{0.0413} & \textbf{0.0769} & \textbf{0.0717} & \textbf{0.0588} & \textbf{0.0703} \\
Gemini 3 Flash Preview & 150.7 & \textbf{0.0616} & \textbf{0.0580} & \textbf{0.0429} & \textbf{0.0501} & \textbf{0.1170} & 0.1406 & 0.1130 & 0.1133 \\
GPT-5 & 150.0 & \textbf{0.0645} & \textbf{0.0574} & \textbf{0.0471} & \textbf{0.0466} & \textbf{0.1087} & \textbf{0.1130} & \textbf{0.0767} & \textbf{0.0893} \\
o3 & 147.1 & \textbf{0.0580} & \textbf{0.0504} & \textbf{0.0455} & \textbf{0.0504} & \textbf{0.0941} & \textbf{0.0955} & \textbf{0.1036} & 0.1173 \\
GPT-5.4 Nano & 146.5 & \textbf{0.0614} & \textbf{0.0759} & \textbf{0.0632} & 0.0709 & \textbf{0.1292} & 0.1458 & \textbf{0.0902} & 0.1201 \\
o4 Mini & 146.5 & 0.0996 & 0.0868 & 0.0721 & 0.0774 & \textbf{0.1134} & 0.1296 & \textbf{0.1035} & 0.1585 \\
GPT-5 Mini & 145.8 & 0.0863 & \textbf{0.0763} & \textbf{0.0650} & \textbf{0.0650} & \textbf{0.0911} & \textbf{0.0994} & \textbf{0.0770} & 0.1095 \\
Gemini 3.1 Flash Lite Preview & 144.7 & \textbf{0.0773} & \textbf{0.0675} & \textbf{0.0583} & \textbf{0.0586} & \textbf{0.0851} & \textbf{0.1018} & \textbf{0.1087} & 0.1289 \\
Qwen3.5-Flash & 144.1 & 0.0964 & 0.1502 & 0.1240 & 0.1519 & 0.5180 & 0.6602 & 0.4940 & 0.7472 \\
Claude Haiku 4.5 & 142.9 & 0.0803 & \textbf{0.0694} & \textbf{0.0633} & \textbf{0.0655} & \textbf{0.1378} & 0.1711 & 0.1201 & 0.1771 \\
GPT-5 Nano & 140.8 & 0.0866 & 0.0963 & 0.0836 & 0.0850 & 0.1639 & \textbf{0.0969} & \textbf{0.0887} & \textbf{0.0636} \\
Kimi K2 & 140.4 & \textbf{0.0697} & \textbf{0.0741} & \textbf{0.0565} & 0.0668 & 0.1577 & 0.1940 & 0.1372 & 0.1848 \\
Gemini 2.5 Flash & 140.3 & 0.0876 & 0.0983 & 0.0971 & 0.1164 & 0.2743 & 0.3604 & 0.1905 & 0.2288 \\
Qwen3 235B A22B & 139.6 & \textbf{0.0690} & \textbf{0.0737} & \textbf{0.0619} & 0.0739 & 0.2148 & 0.2749 & 0.1994 & 0.2435 \\
GPT-4.1 & 137.5 & 0.0954 & 0.0794 & 0.0671 & 0.0727 & \textbf{0.1133} & \textbf{0.1077} & \textbf{0.0806} & 0.0977 \\
DeepSeek V3 & 133.0 & \textbf{0.0621} & 0.0856 & \textbf{0.0624} & 0.0776 & 0.2876 & 0.3179 & 0.2213 & 0.2596 \\
Llama 4 Scout & 130.5 & 0.0833 & 0.0887 & 0.0708 & 0.0682 & \textbf{0.1260} & 0.1493 & \textbf{0.1099} & 0.1438 \\
\midrule
Baseline &  & 0.0796 & 0.0782 & 0.0667 & 0.0661 & 0.1545 & 0.1287 & 0.1103 & 0.0931 \\
\bottomrule
\end{tabular}

%% file: data/appendix_tables/micropolis_continuous.tex
\setlength{\tabcolsep}{3pt}
\begin{tabular}{lrrrrrrrrrr}
\toprule
Model & ECI & \multicolumn{4}{c}{By horizon (years)} & \multicolumn{5}{c}{By metric} \\
\cmidrule(lr){3-6}\cmidrule(lr){7-11}
 & & 3 & 5 & 7 & 10 & Popul. & Traffic & Pollut. & Crime & Land \\
\midrule
Claude Fable & 162.5 & \textbf{0.019} & \textbf{0.025} & \textbf{0.029} & \textbf{0.034} & \textbf{0.031} & \textbf{0.034} & \textbf{0.021} & \textbf{0.029} & \textbf{0.019} \\
Claude Opus 5 & 161.5 & \textbf{0.020} & \textbf{0.026} & \textbf{0.031} & \textbf{0.039} & \textbf{0.033} & \textbf{0.032} & 0.028 & \textbf{0.031} & \textbf{0.022} \\
GPT-5.6 Sol & 161.1 & \textbf{0.021} & \textbf{0.027} & \textbf{0.030} & \textbf{0.037} & \textbf{0.031} & \textbf{0.033} & 0.025 & \textbf{0.032} & \textbf{0.022} \\
GPT-5.5 & 158.6 & \textbf{0.020} & \textbf{0.026} & \textbf{0.030} & \textbf{0.037} & \textbf{0.030} & \textbf{0.031} & 0.025 & \textbf{0.035} & \textbf{0.020} \\
Gemini 3.7 Flash & 156.8 & \textbf{0.021} & \textbf{0.028} & \textbf{0.032} & \textbf{0.040} & \textbf{0.036} & \textbf{0.035} & 0.026 & \textbf{0.031} & \textbf{0.023} \\
GPT-5.6 Luna & 156.2 & \textbf{0.024} & \textbf{0.031} & \textbf{0.036} & 0.043 & \textbf{0.031} & \textbf{0.038} & 0.035 & 0.039 & \textbf{0.026} \\
Claude Sonnet 5 & 155.9 & \textbf{0.022} & \textbf{0.028} & \textbf{0.032} & \textbf{0.038} & \textbf{0.036} & \textbf{0.043} & \textbf{0.016} & \textbf{0.034} & \textbf{0.020} \\
DeepSeek V4 Flash 0731 & 154.4 & \textbf{0.029} & \textbf{0.035} & 0.039 & 0.047 & 0.056 & \textbf{0.040} & 0.027 & \textbf{0.032} & 0.032 \\
Gemini 3 Flash Preview & 150.7 & \textbf{0.023} & \textbf{0.032} & 0.041 & 0.052 & \textbf{0.032} & \textbf{0.031} & 0.050 & \textbf{0.035} & 0.036 \\
GPT-5 & 150.0 & \textbf{0.025} & \textbf{0.031} & \textbf{0.035} & \textbf{0.040} & \textbf{0.035} & \textbf{0.047} & 0.027 & \textbf{0.031} & \textbf{0.024} \\
o3 & 147.1 & \textbf{0.026} & \textbf{0.033} & \textbf{0.038} & 0.047 & \textbf{0.032} & \textbf{0.047} & 0.038 & 0.036 & \textbf{0.026} \\
GPT-5.4 Nano & 146.5 & \textbf{0.029} & 0.037 & 0.044 & 0.054 & 0.043 & \textbf{0.058} & 0.027 & 0.036 & 0.041 \\
o4 Mini & 146.5 & 0.034 & 0.039 & 0.045 & 0.050 & 0.053 & \textbf{0.054} & 0.033 & 0.039 & 0.031 \\
GPT-5 Mini & 145.8 & 0.033 & 0.039 & 0.044 & 0.051 & 0.040 & \textbf{0.049} & 0.038 & 0.041 & 0.043 \\
Gemini 3.1 Flash Lite Preview & 144.7 & \textbf{0.025} & \textbf{0.032} & 0.039 & 0.052 & 0.040 & \textbf{0.039} & 0.036 & \textbf{0.033} & 0.036 \\
Qwen3.5-Flash & 144.1 & 0.037 & 0.047 & 0.055 & 0.065 & 0.049 & 0.100 & 0.027 & 0.041 & 0.039 \\
Claude Haiku 4.5 & 142.9 & \textbf{0.031} & 0.039 & 0.044 & 0.051 & \textbf{0.034} & \textbf{0.046} & 0.030 & 0.037 & 0.060 \\
GPT-5 Nano & 140.8 & 0.058 & 0.068 & 0.075 & 0.083 & 0.148 & 0.079 & \textbf{0.024} & 0.056 & 0.047 \\
Kimi K2 & 140.4 & \textbf{0.030} & 0.036 & 0.043 & 0.058 & \textbf{0.034} & \textbf{0.041} & 0.039 & 0.047 & 0.048 \\
Gemini 2.5 Flash & 140.3 & \textbf{0.030} & 0.036 & 0.043 & 0.055 & \textbf{0.039} & \textbf{0.049} & 0.036 & 0.044 & 0.038 \\
Qwen3 235B A22B & 139.6 & \textbf{0.030} & 0.039 & 0.051 & 0.066 & 0.042 & \textbf{0.052} & 0.051 & 0.043 & 0.044 \\
GPT-4.1 & 137.5 & \textbf{0.030} & 0.036 & 0.044 & 0.056 & \textbf{0.029} & \textbf{0.036} & 0.067 & 0.042 & 0.033 \\
DeepSeek V3 & 133.0 & \textbf{0.026} & \textbf{0.033} & 0.040 & 0.052 & \textbf{0.036} & \textbf{0.051} & 0.030 & 0.041 & 0.031 \\
Llama 4 Scout & 130.5 & \textbf{0.031} & 0.037 & 0.045 & 0.058 & 0.057 & \textbf{0.059} & 0.031 & \textbf{0.031} & 0.037 \\
\midrule
Baseline &  & 0.032 & 0.036 & 0.039 & 0.042 & 0.040 & 0.059 & 0.024 & 0.036 & 0.027 \\
\bottomrule
\end{tabular}

%% file: data/appendix_tables/starsim_models.tex
\setlength{\tabcolsep}{2.2pt}
\begin{tabular}{lrrrrrrrrrrr}
\toprule
Model & ECI & Binary & Cont. & \multicolumn{4}{c}{Interventional} & Binary & Hold & Calls & Cost \\
\cmidrule(lr){5-8}
 & & bits & excess & 25\% & 50\% & 90\% & pooled & c90 bits & bits & & (\$) \\
\midrule
Fable & 162.5 & 0.018 & \textbf{0.274}$^{*}$ & \textbf{0.428} & \textbf{0.440} & \textbf{0.658} & \textbf{0.509} & 0.103 & 0.097 & 216 & 8.15 \\
Opus 5 & 161.5 & 0.024 & \textbf{0.255}$^{*}$ & \textbf{0.541} & \textbf{0.278} & \textbf{0.168} & \textbf{0.329} & 0.125 & 0.047 & 216 & 3.01 \\
GPT-5.6 Sol & 161.1 & 0.001 & \textbf{0.308}$^{*}$ & \textbf{0.267} & \textbf{0.250} & \textbf{0.017} & \textbf{0.178} & 0.004 & 0.038 & 216 & 2.88 \\
GPT-5.5 & 158.6 & 0.003 & \textbf{0.393}$^{*}$ & \textbf{0.297} & \textbf{0.380} & \textbf{0.102} & \textbf{0.260} & 0.052 & 0.159 & 216 & 9.49 \\
Gemini 3.7 Flash & 156.8 & 0.005 & \textbf{0.424}\phantom{$^{*}$} & \textbf{1.058} & \textbf{0.918} & \textbf{0.327} & \textbf{0.768} & 0.069 & 0.018 & 216 & 0.81 \\
GPT-5.6 Luna & 156.2 & 0.016 & \textbf{0.496}\phantom{$^{*}$} & \textbf{0.285} & \textbf{0.291} & \textbf{0.063} & \textbf{0.213} & 0.060 & 0.471 & 216 & 0.35 \\
Sonnet 5 & 155.9 & 0.055 & \textbf{0.502}\phantom{$^{*}$} & \textbf{0.321} & \textbf{0.613} & \textbf{0.318} & \textbf{0.417} & 1.628 & 0.161 & 216 & 1.64 \\
DeepSeek V4 Flash$^{\dagger}$ & 154.4 & 0.001 & \textbf{0.239}$^{*}$ & \textbf{0.246} & \textbf{0.298} & \textbf{0.186} & \textbf{0.243} & 0.004 & -- & 111 & 0.45 \\
Gemini 3 Flash & 150.7 & 0.013 & \textbf{0.577}\phantom{$^{*}$} & \textbf{0.924} & \textbf{0.764} & \textbf{0.356} & \textbf{0.681} & 0.133 & 0.046 & 216 & 0.45 \\
GPT-5 & 150.0 & 0.051 & \textbf{0.814}\phantom{$^{*}$} & \textbf{0.878} & \textbf{0.501} & \textbf{0.167} & \textbf{0.516} & 1.619 & 0.086 & 216 & 0.81 \\
o3 & 147.1 & 0.115 & \textbf{0.314}$^{*}$ & \textbf{0.511} & \textbf{0.374} & \textbf{0.328} & \textbf{0.404} & 0.373 & 0.703 & 216 & 4.16 \\
GPT-5.4 Nano & 146.5 & 0.042 & \textbf{0.227}$^{*}$ & \textbf{1.014} & \textbf{1.012} & \textbf{0.403} & \textbf{0.810} & 0.144 & 0.219 & 216 & 0.53 \\
o4-mini & 146.5 & 0.016 & \textbf{0.435}\phantom{$^{*}$} & \textbf{1.150} & \textbf{1.663} & \textbf{0.499} & \textbf{1.104} & 0.052 & 1.077 & 216 & 1.07 \\
GPT-5 Mini & 145.8 & 0.050 & \textbf{1.498}\phantom{$^{*}$} & \textbf{1.054} & \textbf{0.431} & \textbf{0.155} & \textbf{0.547} & 0.716 & 0.244 & 216 & 0.24 \\
Gemini 3.1 Flash Lite & 144.7 & 0.027 & \textbf{0.424}\phantom{$^{*}$} & \textbf{0.394} & \textbf{0.467} & \textbf{0.601} & \textbf{0.488} & 0.114 & 0.376 & 216 & 0.20 \\
Qwen3.5 Flash & 144.1 & 0.059 & \textbf{1.340}\phantom{$^{*}$} & \textbf{0.683} & \textbf{0.699} & \textbf{0.255} & \textbf{0.546} & 0.186 & 0.916 & 216 & 0.26 \\
Haiku 4.5 & 142.9 & 0.092 & 2.123\phantom{$^{*}$} & \textbf{0.933} & \textbf{0.433} & \textbf{0.673} & \textbf{0.680} & 1.931 & 0.281 & 216 & 1.35 \\
GPT-5 Nano & 140.8 & 0.272 & 3.836\phantom{$^{*}$} & 2.987 & \textbf{0.951} & \textbf{0.886} & \textbf{1.608} & 1.001 & 0.497 & 216 & 0.04 \\
Kimi K2 & 140.4 & 0.122 & \textbf{0.793}\phantom{$^{*}$} & \textbf{0.609} & \textbf{1.046} & \textbf{0.484} & \textbf{0.713} & 0.668 & 0.282 & 182 & 1.67 \\
Gemini 2.5 Flash & 140.3 & 0.013 & \textbf{0.540}\phantom{$^{*}$} & \textbf{1.137} & \textbf{0.319} & \textbf{0.148} & \textbf{0.535} & 0.054 & 0.991 & 216 & 1.56 \\
Qwen3 235B & 139.6 & 0.130 & 2.901\phantom{$^{*}$} & 2.932 & \textbf{0.967} & \textbf{0.160} & \textbf{1.353} & 1.486 & 0.636 & 216 & 0.49 \\
GPT-4.1 & 137.5 & 0.018 & \textbf{0.652}\phantom{$^{*}$} & \textbf{1.017} & \textbf{0.494} & \textbf{0.421} & \textbf{0.644} & 1.530 & 0.036 & 216 & 1.20 \\
DeepSeek V3 & 133.0 & 0.157 & 2.390\phantom{$^{*}$} & 3.287 & \textbf{1.775} & \textbf{0.395} & \textbf{1.819} & 0.449 & 0.833 & 216 & 0.13 \\
Llama 4 Scout & 130.5 & 0.342 & 4.363\phantom{$^{*}$} & 2.070 & \textbf{0.757} & \textbf{0.210} & \textbf{1.013} & 1.404 & 0.513 & 216 & 0.06 \\
\midrule
Baseline & & & 1.681\phantom{$^{*}$} & 1.460 & 2.459 & 4.672 & 2.864 & & & & \\
SIR forecast & & & 0.414\phantom{$^{*}$} & & & & & & & & \\
\bottomrule
\end{tabular}

%% file: data/appendix_tables/starsim_horizon.tex
\setlength{\tabcolsep}{2.2pt}
\begin{tabular}{lrrrrrrrrrrr}
\toprule
Model & ECI & \multicolumn{2}{c}{Binary bits} & \multicolumn{2}{c}{Cont. excess} & \multicolumn{2}{c}{Interv. excess} & \multicolumn{2}{c}{Binary c90 bits} & \multicolumn{2}{c}{Hold bits} \\
\cmidrule(lr){3-4}\cmidrule(lr){5-6}\cmidrule(lr){7-8}\cmidrule(lr){9-10}\cmidrule(lr){11-12}
 & & d40 & d60 & d40 & d60 & d40 & d60 & d40 & d60 & d40 & d60 \\
\midrule
Fable & 162.5 & 0.018 & 0.018 & \textbf{0.304}$^{*}$ & \textbf{0.244}$^{*}$ & \textbf{0.448} & \textbf{0.569} & 0.125 & 0.082 & 0.058 & 0.136 \\
Opus 5 & 161.5 & 0.026 & 0.022 & \textbf{0.284}$^{*}$ & \textbf{0.226}$^{*}$ & \textbf{0.263} & \textbf{0.395} & 0.144 & 0.105 & 0.054 & 0.040 \\
GPT-5.6 Sol & 161.1 & 0.001 & 0.001 & \textbf{0.398}$^{*}$ & \textbf{0.217}$^{*}$ & \textbf{0.209} & \textbf{0.146} & 0.006 & 0.001 & 0.021 & 0.056 \\
GPT-5.5 & 158.6 & 0.002 & 0.005 & \textbf{0.542}\phantom{$^{*}$} & \textbf{0.244}$^{*}$ & \textbf{0.243} & \textbf{0.277} & 0.095 & 0.010 & 0.077 & 0.241 \\
Gemini 3.7 Flash & 156.8 & 0.001 & 0.008 & \textbf{0.531}\phantom{$^{*}$} & \textbf{0.318}$^{*}$ & \textbf{0.921} & \textbf{0.615} & 0.089 & 0.048 & 0.020 & 0.016 \\
GPT-5.6 Luna & 156.2 & 0.018 & 0.014 & \textbf{0.354}$^{*}$ & \textbf{0.639}\phantom{$^{*}$} & \textbf{0.139} & \textbf{0.287} & 0.095 & 0.026 & 0.158 & 0.784 \\
Sonnet 5 & 155.9 & 0.048 & 0.063 & \textbf{0.700}\phantom{$^{*}$} & \textbf{0.304}$^{*}$ & \textbf{0.443} & \textbf{0.391} & 1.416 & 1.840 & 0.260 & 0.062 \\
DeepSeek V4 Flash$^{\dagger}$ & 154.4 & 0.001 & 0.001 & \textbf{0.268}$^{*}$ & \textbf{0.210}$^{*}$ & \textbf{0.187} & \textbf{0.317} & 0.005 & 0.003 & -- & -- \\
Gemini 3 Flash & 150.7 & 0.011 & 0.014 & \textbf{0.906}\phantom{$^{*}$} & \textbf{0.248}$^{*}$ & \textbf{0.970} & \textbf{0.393} & 0.152 & 0.114 & 0.088 & 0.003 \\
GPT-5 & 150.0 & 0.052 & 0.051 & \textbf{1.228}\phantom{$^{*}$} & \textbf{0.400}\phantom{$^{*}$} & \textbf{0.559} & \textbf{0.472} & 2.127 & 1.110 & 0.103 & 0.069 \\
o3 & 147.1 & 0.105 & 0.125 & \textbf{0.296}$^{*}$ & \textbf{0.331}$^{*}$ & \textbf{0.323} & \textbf{0.485} & 0.423 & 0.322 & 0.358 & 1.047 \\
GPT-5.4 Nano & 146.5 & 0.065 & 0.018 & \textbf{0.286}$^{*}$ & \textbf{0.168}$^{*}$ & \textbf{0.678} & \textbf{0.941} & 0.192 & 0.095 & 0.318 & 0.120 \\
o4-mini & 146.5 & 0.018 & 0.014 & \textbf{0.501}\phantom{$^{*}$} & \textbf{0.369}$^{*}$ & \textbf{0.998} & \textbf{1.210} & 0.048 & 0.056 & 1.348 & 0.806 \\
GPT-5 Mini & 145.8 & 0.048 & 0.051 & \textbf{1.471}\phantom{$^{*}$} & 1.525\phantom{$^{*}$} & \textbf{0.485} & \textbf{0.608} & 0.696 & 0.736 & 0.479 & 0.009 \\
Gemini 3.1 Flash Lite & 144.7 & 0.033 & 0.022 & \textbf{0.559}\phantom{$^{*}$} & \textbf{0.288}$^{*}$ & \textbf{0.553} & \textbf{0.422} & 0.154 & 0.074 & 0.084 & 0.668 \\
Qwen3.5 Flash & 144.1 & 0.052 & 0.067 & 2.340\phantom{$^{*}$} & \textbf{0.340}$^{*}$ & \textbf{0.606} & \textbf{0.485} & 0.238 & 0.135 & 0.986 & 0.846 \\
Haiku 4.5 & 142.9 & 0.098 & 0.086 & 3.050\phantom{$^{*}$} & \textbf{1.196}\phantom{$^{*}$} & \textbf{0.942} & \textbf{0.418} & 1.996 & 1.867 & 0.349 & 0.213 \\
GPT-5 Nano & 140.8 & 0.257 & 0.288 & 3.684\phantom{$^{*}$} & 3.987\phantom{$^{*}$} & \textbf{1.918} & \textbf{1.298} & 1.152 & 0.851 & 0.525 & 0.470 \\
Kimi K2 & 140.4 & 0.096 & 0.148 & \textbf{1.019}\phantom{$^{*}$} & \textbf{0.567}\phantom{$^{*}$} & \textbf{0.781} & \textbf{0.644} & 0.837 & 0.499 & 0.408 & 0.155 \\
Gemini 2.5 Flash & 140.3 & 0.013 & 0.013 & \textbf{0.580}\phantom{$^{*}$} & \textbf{0.501}\phantom{$^{*}$} & \textbf{0.612} & \textbf{0.457} & 0.063 & 0.044 & 0.911 & 1.072 \\
Qwen3 235B & 139.6 & 0.145 & 0.114 & 3.594\phantom{$^{*}$} & 2.208\phantom{$^{*}$} & \textbf{1.756} & \textbf{0.951} & 1.486 & 1.486 & 0.699 & 0.574 \\
GPT-4.1 & 137.5 & 0.018 & 0.018 & \textbf{0.908}\phantom{$^{*}$} & \textbf{0.397}\phantom{$^{*}$} & \textbf{0.782} & \textbf{0.505} & 2.300 & 0.761 & 0.042 & 0.030 \\
DeepSeek V3 & 133.0 & 0.074 & 0.241 & 3.192\phantom{$^{*}$} & 1.588\phantom{$^{*}$} & \textbf{1.904} & \textbf{1.734} & 0.673 & 0.224 & 0.856 & 0.809 \\
Llama 4 Scout & 130.5 & 0.370 & 0.313 & 5.634\phantom{$^{*}$} & 3.092\phantom{$^{*}$} & \textbf{1.092} & \textbf{0.933} & 1.397 & 1.411 & 0.416 & 0.610 \\
\midrule
Baseline & & & & 2.127\phantom{$^{*}$} & 1.236\phantom{$^{*}$} & 3.276 & 2.451 & & & & \\
SIR forecast & & & & 0.440\phantom{$^{*}$} & 0.388\phantom{$^{*}$} & & & & & & \\
\bottomrule
\end{tabular}

%% file: data/appendix_tables/starsim_rung_horizon.tex
\setlength{\tabcolsep}{3pt}
\begin{tabular}{lrrrrrrr}
\toprule
Model & ECI & \multicolumn{2}{c}{25\% coverage} & \multicolumn{2}{c}{50\% coverage} & \multicolumn{2}{c}{90\% coverage} \\
\cmidrule(lr){3-4}\cmidrule(lr){5-6}\cmidrule(lr){7-8}
 & & d40 & d60 & d40 & d60 & d40 & d60 \\
\midrule
Fable & 162.5 & \textbf{0.351} & \textbf{0.505} & \textbf{0.268} & \textbf{0.612} & \textbf{0.725} & \textbf{0.591} \\
Opus 5 & 161.5 & \textbf{0.473} & \textbf{0.609} & \textbf{0.207} & \textbf{0.349} & \textbf{0.110} & \textbf{0.227} \\
GPT-5.6 Sol & 161.1 & \textbf{0.325} & \textbf{0.209} & \textbf{0.282} & \textbf{0.217} & \textbf{0.021} & \textbf{0.013} \\
GPT-5.5 & 158.6 & \textbf{0.256} & \textbf{0.338} & \textbf{0.380} & \textbf{0.380} & \textbf{0.092} & \textbf{0.112} \\
Gemini 3.7 Flash & 156.8 & \textbf{1.154} & \textbf{0.963} & \textbf{1.214} & \textbf{0.621} & \textbf{0.394} & \textbf{0.260} \\
GPT-5.6 Luna & 156.2 & \textbf{0.223} & \textbf{0.348} & \textbf{0.151} & \textbf{0.431} & \textbf{0.042} & \textbf{0.083} \\
Sonnet 5 & 155.9 & \textbf{0.346} & \textbf{0.295} & \textbf{0.648} & \textbf{0.578} & \textbf{0.336} & \textbf{0.301} \\
DeepSeek V4 Flash$^{\dagger}$ & 154.4 & \textbf{0.141} & \textbf{0.385} & \textbf{0.292} & \textbf{0.302} & \textbf{0.128} & \textbf{0.264} \\
Gemini 3 Flash & 150.7 & \textbf{1.246} & \textbf{0.601} & \textbf{1.334} & \textbf{0.194} & \textbf{0.331} & \textbf{0.382} \\
GPT-5 & 150.0 & \textbf{0.885} & \textbf{0.871} & \textbf{0.672} & \textbf{0.331} & \textbf{0.120} & \textbf{0.214} \\
o3 & 147.1 & \textbf{0.232} & \textbf{0.790} & \textbf{0.237} & \textbf{0.511} & \textbf{0.501} & \textbf{0.155} \\
GPT-5.4 Nano & 146.5 & \textbf{0.917} & \textbf{1.111} & \textbf{0.678} & \textbf{1.345} & \textbf{0.439} & \textbf{0.368} \\
o4-mini & 146.5 & \textbf{1.228} & \textbf{1.073} & \textbf{1.201} & 2.125 & \textbf{0.565} & \textbf{0.432} \\
GPT-5 Mini & 145.8 & \textbf{0.780} & 1.327 & \textbf{0.482} & \textbf{0.379} & \textbf{0.193} & \textbf{0.116} \\
Gemini 3.1 Flash Lite & 144.7 & \textbf{0.537} & \textbf{0.251} & \textbf{0.461} & \textbf{0.474} & \textbf{0.660} & \textbf{0.543} \\
Qwen3.5 Flash & 144.1 & \textbf{0.877} & \textbf{0.488} & \textbf{0.793} & \textbf{0.605} & \textbf{0.148} & \textbf{0.363} \\
Haiku 4.5 & 142.9 & \textbf{1.433} & \textbf{0.433} & \textbf{0.348} & \textbf{0.519} & \textbf{1.044} & \textbf{0.302} \\
GPT-5 Nano & 140.8 & 3.449 & 2.526 & \textbf{1.152} & \textbf{0.750} & \textbf{1.153} & \textbf{0.619} \\
Kimi K2 & 140.4 & \textbf{0.583} & \textbf{0.635} & \textbf{1.111} & \textbf{0.980} & \textbf{0.649} & \textbf{0.318} \\
Gemini 2.5 Flash & 140.3 & \textbf{1.114} & \textbf{1.161} & \textbf{0.528} & \textbf{0.109} & \textbf{0.195} & \textbf{0.101} \\
Qwen3 235B & 139.6 & 4.016 & 1.849 & \textbf{1.033} & \textbf{0.902} & \textbf{0.218} & \textbf{0.102} \\
GPT-4.1 & 137.5 & 1.669 & \textbf{0.365} & \textbf{0.292} & \textbf{0.697} & \textbf{0.386} & \textbf{0.455} \\
DeepSeek V3 & 133.0 & 3.134 & 3.440 & \textbf{2.155} & \textbf{1.395} & \textbf{0.422} & \textbf{0.368} \\
Llama 4 Scout & 130.5 & 2.254 & 1.887 & \textbf{0.754} & \textbf{0.761} & \textbf{0.268} & \textbf{0.152} \\
\midrule
Baseline & & 1.653 & 1.267 & 2.795 & 2.123 & 5.381 & 3.963 \\
\bottomrule
\end{tabular}

%% file: data/starsim/per_item_table.tex
\begin{tabular}{@{}l l l r r r r r r r r@{}}
\toprule
Set & Item & $\beta$ ($R_0$) or $\beta_R$ / $\beta_S$ & Cov.\ \% & Day & Truth & Floor & $C$ & Excess med.\ & Min & Max \\
\midrule
Binary, unconditional & \texttt{w0\_Rlead\_d40} & 0.040 / 0.030 & -- & 40 & 1.00 & -- & -- & 0.01 & 0.00 & 0.51 \\
 & \texttt{w0\_Rlead\_d60} & 0.040 / 0.030 & -- & 60 & 1.00 & -- & -- & 0.03 & 0.00 & 0.51 \\
 & \texttt{w1\_Rlead\_d40} & 0.050 / 0.040 & -- & 40 & 1.00 & -- & -- & 0.03 & 0.00 & 0.36 \\
 & \texttt{w1\_Rlead\_d60} & 0.050 / 0.040 & -- & 60 & 1.00 & -- & -- & 0.03 & 0.00 & 0.42 \\
 & \texttt{w2\_Slead\_d40} & 0.050 / 0.065 & -- & 40 & 0.00 & -- & -- & 0.05 & 0.00 & 0.32 \\
 & \texttt{w2\_Slead\_d60} & 0.050 / 0.065 & -- & 60 & 0.00 & -- & -- & 0.02 & 0.00 & 0.79 \\
 & \texttt{w3\_Slead\_d40} & 0.030 / 0.040 & -- & 40 & 0.00 & -- & -- & 0.01 & 0.00 & 0.32 \\
 & \texttt{w3\_Slead\_d60} & 0.030 / 0.040 & -- & 60 & 0.00 & -- & -- & 0.02 & 0.00 & 0.32 \\
\addlinespace
Continuous, unconditional & \texttt{w0\_d40} & 0.030 (1.8) & -- & 40 & 1669 & 52 & 200 & 0.68 & 0.20 & 6.77 \\
 & \texttt{w0\_d60} & 0.030 (1.8) & -- & 60 & 2824 & 45 & 300 & 0.52 & 0.07 & 7.54 \\
 & \texttt{w1\_d40} & 0.040 (2.4) & -- & 40 & 2845 & 26 & 200 & 0.86 & 0.08 & 9.28 \\
 & \texttt{w1\_d60} & 0.040 (2.4) & -- & 60 & 3128 & 21 & 300 & 0.37 & 0.16 & 5.85 \\
 & \texttt{w2\_d40} & 0.050 (3.0) & -- & 40 & 2276 & 13 & 200 & 0.51 & 0.10 & 5.69 \\
 & \texttt{w2\_d60} & 0.050 (3.0) & -- & 60 & 2308 & 12 & 300 & 0.37 & 0.07 & 2.33 \\
 & \texttt{w3\_d40} & 0.065 (3.9) & -- & 40 & 865 & 7 & 200 & 0.29 & 0.08 & 3.04 \\
 & \texttt{w3\_d60} & 0.065 (3.9) & -- & 60 & 866 & 7 & 300 & 0.22 & 0.04 & 2.32 \\
\addlinespace
Continuous, 25\% coverage & \texttt{w0\_d40} & 0.030 (1.8) & 25 & 40 & 863 & 37 & 200 & 0.32 & 0.01 & 3.26 \\
 & \texttt{w0\_d60} & 0.030 (1.8) & 25 & 60 & 1383 & 52 & 300 & 0.44 & 0.04 & 3.29 \\
 & \texttt{w1\_d40} & 0.040 (2.4) & 25 & 40 & 1813 & 30 & 200 & 1.50 & 0.11 & 6.68 \\
 & \texttt{w1\_d60} & 0.040 (2.4) & 25 & 60 & 2051 & 30 & 300 & 0.69 & 0.10 & 4.44 \\
 & \texttt{w2\_d40} & 0.050 (3.0) & 25 & 40 & 1629 & 19 & 200 & 0.96 & 0.11 & 4.64 \\
 & \texttt{w2\_d60} & 0.050 (3.0) & 25 & 60 & 1656 & 19 & 300 & 0.65 & 0.13 & 4.33 \\
 & \texttt{w3\_d40} & 0.065 (3.9) & 25 & 40 & 655 & 9 & 200 & 0.37 & 0.08 & 2.95 \\
 & \texttt{w3\_d60} & 0.065 (3.9) & 25 & 60 & 656 & 9 & 300 & 0.30 & 0.04 & 2.14 \\
\addlinespace
Continuous, 50\% coverage & \texttt{w0\_d40} & 0.030 (1.8) & 50 & 40 & 388 & 21 & 200 & 0.24 & 0.06 & 2.06 \\
 & \texttt{w0\_d60} & 0.030 (1.8) & 50 & 60 & 494 & 29 & 300 & 0.41 & 0.04 & 4.61 \\
 & \texttt{w1\_d40} & 0.040 (2.4) & 50 & 40 & 964 & 26 & 200 & 0.74 & 0.15 & 3.60 \\
 & \texttt{w1\_d60} & 0.040 (2.4) & 50 & 60 & 1065 & 29 & 300 & 0.71 & 0.08 & 2.65 \\
 & \texttt{w2\_d40} & 0.050 (3.0) & 50 & 40 & 1023 & 19 & 200 & 0.54 & 0.19 & 3.59 \\
 & \texttt{w2\_d60} & 0.050 (3.0) & 50 & 60 & 1039 & 20 & 300 & 0.49 & 0.07 & 3.08 \\
 & \texttt{w3\_d40} & 0.065 (3.9) & 50 & 40 & 449 & 10 & 200 & 0.30 & 0.02 & 1.11 \\
 & \texttt{w3\_d60} & 0.065 (3.9) & 50 & 60 & 450 & 10 & 300 & 0.19 & 0.03 & 1.34 \\
\addlinespace
Continuous, 90\% coverage & \texttt{w0\_d40} & 0.030 (1.8) & 90 & 40 & 55 & 5 & 200 & 0.11 & 0.01 & 0.90 \\
 & \texttt{w0\_d60} & 0.030 (1.8) & 90 & 60 & 55 & 5 & 300 & 0.12 & 0.00 & 1.06 \\
 & \texttt{w1\_d40} & 0.040 (2.4) & 90 & 40 & 161 & 8 & 200 & 0.35 & 0.02 & 1.53 \\
 & \texttt{w1\_d60} & 0.040 (2.4) & 90 & 60 & 162 & 9 & 300 & 0.35 & 0.02 & 1.18 \\
 & \texttt{w2\_d40} & 0.050 (3.0) & 90 & 40 & 242 & 9 & 200 & 0.38 & 0.04 & 1.76 \\
 & \texttt{w2\_d60} & 0.050 (3.0) & 90 & 60 & 243 & 9 & 300 & 0.26 & 0.02 & 1.08 \\
 & \texttt{w3\_d40} & 0.065 (3.9) & 90 & 40 & 139 & 6 & 200 & 0.14 & 0.00 & 2.82 \\
 & \texttt{w3\_d60} & 0.065 (3.9) & 90 & 60 & 139 & 6 & 300 & 0.13 & 0.00 & 1.21 \\
\addlinespace
Binary, 90\% coverage & \texttt{w0\_Rlead\_d40} & 0.040 / 0.030 & 90 & 40 & 0.00 & -- & -- & 0.23 & 0.00 & 3.64 \\
 & \texttt{w0\_Rlead\_d60} & 0.040 / 0.030 & 90 & 60 & 0.00 & -- & -- & 0.07 & 0.00 & 2.74 \\
 & \texttt{w1\_Rlead\_d40} & 0.050 / 0.040 & 90 & 40 & 0.00 & -- & -- & 0.12 & 0.00 & 2.74 \\
 & \texttt{w1\_Rlead\_d60} & 0.050 / 0.040 & 90 & 60 & 0.00 & -- & -- & 0.08 & 0.00 & 3.64 \\
 & \texttt{w2\_Slead\_d40} & 0.050 / 0.065 & 90 & 40 & 1.00 & -- & -- & 0.23 & 0.00 & 2.74 \\
 & \texttt{w2\_Slead\_d60} & 0.050 / 0.065 & 90 & 60 & 1.00 & -- & -- & 0.22 & 0.00 & 2.74 \\
 & \texttt{w3\_Slead\_d40} & 0.030 / 0.040 & 90 & 40 & 1.00 & -- & -- & 0.19 & 0.00 & 4.64 \\
 & \texttt{w3\_Slead\_d60} & 0.030 / 0.040 & 90 & 60 & 1.00 & -- & -- & 0.14 & 0.00 & 3.47 \\
\addlinespace
Binary, hold & \texttt{w0\_Rlead\_d40} & 0.040 / 0.030 & 20 & 40 & 1.00 & -- & -- & 0.07 & 0.00 & 0.51 \\
 & \texttt{w0\_Rlead\_d60} & 0.040 / 0.030 & 20 & 60 & 0.97 & -- & -- & 0.08 & 0.00 & 1.51 \\
 & \texttt{w1\_Rlead\_d40} & 0.050 / 0.040 & 10 & 40 & 1.00 & -- & -- & 0.10 & 0.01 & 2.00 \\
 & \texttt{w1\_Rlead\_d60} & 0.050 / 0.040 & 10 & 60 & 0.98 & -- & -- & 0.07 & 0.00 & 2.27 \\
 & \texttt{w2\_Slead\_d40} & 0.050 / 0.065 & 15 & 40 & 0.00 & -- & -- & 0.35 & 0.02 & 3.63 \\
 & \texttt{w2\_Slead\_d60} & 0.050 / 0.065 & 15 & 60 & 0.01 & -- & -- & 0.27 & 0.01 & 4.21 \\
 & \texttt{w3\_Slead\_d40} & 0.030 / 0.040 & 20 & 40 & 0.00 & -- & -- & 0.07 & 0.00 & 0.47 \\
 & \texttt{w3\_Slead\_d60} & 0.030 / 0.040 & 20 & 60 & 0.03 & -- & -- & 0.03 & 0.00 & 2.47 \\
\bottomrule
\end{tabular}

%% file: data/starsim/fixed_state_uncertainty.tex
\begin{tabular}{llrrrr}
\toprule
Score & Comparator & $n$ & $\rho$ & Models, 95\% & Replays, 95\% \\
\midrule
Continuous baseline & ECI & 24 & $0.75$ & $[0.49, 0.88]$ & $[0.75, 0.76]$ \\
Continuous baseline & ForecastBench & 17 & $0.24$ & $[-0.35, 0.82]$ & $[0.24, 0.24]$ \\
Interventional outcomes & ECI & 23 & $0.67$ & $[0.34, 0.85]$ & $[0.67, 0.68]$ \\
Interventional outcomes & ForecastBench & 17 & $0.45$ & $[-0.14, 0.88]$ & $[0.45, 0.45]$ \\
Combined nine cells & ECI & 24 & $0.86$ & $[0.69, 0.93]$ & $[0.86, 0.86]$ \\
Combined nine cells & ForecastBench & 17 & $0.37$ & $[-0.22, 0.81]$ & $[0.37, 0.37]$ \\
\bottomrule
\end{tabular}

%% file: data/appendix_tables/freeciv_models.tex
\setlength{\tabcolsep}{3pt}
\begin{tabular}{lrrrrrrrrr}
\toprule
Model & ECI & Mid-range & Tails & \multicolumn{2}{c}{Continuous} & \multicolumn{2}{c}{Natural conditional} & Parsed & Cost \\
\cmidrule(lr){5-6}\cmidrule(lr){7-8}
 & & excess Brier & bits & nCRPS & excess & excess Brier & gain & (\%) & (\$) \\
\midrule
Fable & 162.5 & 0.1659 & \textbf{0.206} & 0.792 & 0.641 & 0.1138 & $-$0.0151 & 100.0 & 37.48 \\
Opus 5 & 161.5 & 0.1527 & \textbf{0.223} & 0.789 & 0.638 & 0.1120 & $-$0.0056 & 99.9 & 17.13 \\
GPT-5.6 Sol & 161.1 & 0.1171 & 0.398 & 0.552 & 0.402 & 0.1002 & 0.0159 & 100.0 & 7.24 \\
GPT-5.5 & 158.6 & 0.1304 & \textbf{0.248} & 0.589 & 0.439 & 0.0944 & 0.0163 & 100.0 & 29.80 \\
Gemini 3.7 Flash & 156.8 & 0.1077 & 0.393 & 0.588 & 0.438 & 0.1214 & $-$0.0021 & 100.0 & 3.40 \\
GPT-5.6 Luna & 156.2 & 0.1294 & 0.361 & 0.724 & 0.574 & 0.1058 & $-$0.0023 & 99.9 & 0.66 \\
Sonnet 5 & 155.9 & 0.1733 & \textbf{0.151} & 0.875 & 0.725 & 0.1128 & $-$0.0018 & 100.0 & 5.23 \\
DeepSeek V4 Flash & 154.4 & 0.1614 & \textbf{0.237} & 0.688 & 0.537 & 0.1335 & $-$0.0063 & 99.4 & 0.74 \\
Gemini 3 Flash & 150.7 & 0.1034 & 0.397 & 0.572 & 0.422 & 0.1138 & $-$0.0095 & 100.0 & 2.24 \\
GPT-5 & 150.0 & 0.1146 & 0.390 & 0.776 & 0.625 & 0.1063 & $-$0.0006 & 99.9 & 6.64 \\
o3 & 147.1 & 0.1169 & 0.317 & 0.715 & 0.565 & 0.0889 & 0.0067 & 99.9 & 12.32 \\
GPT-5.4 Nano & 146.5 & 0.1644 & \textbf{0.256} & 0.796 & 0.646 & 0.1123 & $-$0.0071 & 100.0 & 0.97 \\
o4-mini & 146.5 & 0.1679 & \textbf{0.231} & 0.848 & 0.696 & 0.1455 & 0.0054 & 99.2 & 5.32 \\
GPT-5 Mini & 145.8 & 0.1525 & 0.289 & 1.040 & 0.889 & 0.1170 & $-$0.0011 & 100.0 & 1.58 \\
Gemini 3.1 Flash Lite & 144.7 & 0.1316 & 0.303 & 0.743 & 0.593 & 0.1213 & 0.0009 & 100.0 & 0.99 \\
Qwen3.5 Flash & 144.1 & 0.1298 & 0.704 & 0.712 & 0.563 & 0.1544 & $-$0.0106 & 99.8 & 1.15 \\
Haiku 4.5 & 142.9 & 0.1110 & 0.442 & 0.833 & 0.683 & 0.1031 & $-$0.0076 & 100.0 & 6.83 \\
GPT-5 Nano & 140.8 & 0.1917 & \textbf{0.180} & 1.280 & 1.130 & 0.1115 & $-$0.0016 & 100.0 & 0.23 \\
Kimi K2 & 140.4 & 0.1206 & 0.405 & 0.618 & 0.468 & 0.1124 & $-$0.0054 & 99.7 & 6.82 \\
Gemini 2.5 Flash & 140.3 & 0.1542 & 0.453 & 0.827 & 0.676 & 0.1760 & $-$0.0178 & 99.9 & 5.26 \\
Qwen3 235B & 139.6 & 0.1380 & 0.342 & 0.955 & 0.806 & 0.1115 & 0.0146 & 99.1 & 5.89 \\
GPT-4.1 & 137.5 & 0.1361 & 0.413 & 0.866 & 0.716 & 0.1384 & 0.0011 & 100.0 & 12.92 \\
DeepSeek V3 & 133.0 & 0.1310 & 0.384 & 0.928 & 0.777 & 0.1204 & $-$0.0005 & 100.0 & 2.46 \\
Llama 4 Scout & 130.5 & 0.1933 & \textbf{0.237} & 1.060 & 0.909 & 0.1234 & $-$0.0093 & 100.0 & 0.84 \\
\midrule Baseline &  & 0.0739 & 0.268 &  & 0.070 & 0.0636 &  &  &  \\
\bottomrule
\end{tabular}

%% file: data/appendix_tables/freeciv_mirrors.tex
\setlength{\tabcolsep}{3pt}
\begin{tabular}{lrrrrrrr}
\toprule
Model & ECI & Excess Brier & \multicolumn{5}{c}{By horizon (turn)} \\
\cmidrule(lr){4-8}
 & & & 90 & 120 & 150 & 180 & 210 \\
\midrule
Fable & 162.5 & 0.5080 & 0.550 & 0.526 & 0.509 & 0.574 & 0.381 \\
Opus 5 & 161.5 & 0.4441 & 0.433 & 0.419 & 0.490 & 0.469 & 0.409 \\
GPT-5.6 Sol & 161.1 & 0.2684 & 0.325 & 0.188 & 0.321 & 0.270 & 0.238 \\
GPT-5.5 & 158.6 & 0.3810 & 0.466 & 0.330 & 0.429 & 0.347 & 0.334 \\
Gemini 3.7 Flash & 156.8 & 0.2711 & 0.330 & 0.239 & 0.300 & 0.275 & 0.211 \\
GPT-5.6 Luna & 156.2 & 0.3788 & 0.427 & 0.235 & 0.418 & 0.350 & 0.464 \\
Sonnet 5 & 155.9 & 0.5459 & 0.391 & 0.532 & 0.565 & 0.597 & 0.644 \\
DeepSeek V4 Flash & 154.4 & 0.4194 & 0.407 & 0.287 & 0.532 & 0.448 & 0.423 \\
Gemini 3 Flash & 150.7 & 0.2630 & 0.352 & 0.185 & 0.272 & 0.262 & 0.244 \\
GPT-5 & 150.0 & 0.3146 & 0.346 & 0.253 & 0.368 & 0.309 & 0.298 \\
o3 & 147.1 & 0.3354 & 0.359 & 0.257 & 0.355 & 0.395 & 0.311 \\
GPT-5.4 Nano & 146.5 & 0.5527 & 0.541 & 0.443 & 0.580 & 0.575 & 0.624 \\
o4-mini & 146.5 & 0.4981 & 0.497 & 0.404 & 0.621 & 0.564 & 0.406 \\
GPT-5 Mini & 145.8 & 0.4242 & 0.357 & 0.324 & 0.509 & 0.434 & 0.497 \\
Gemini 3.1 Flash Lite & 144.7 & 0.3913 & 0.283 & 0.380 & 0.364 & 0.490 & 0.439 \\
Qwen3.5 Flash & 144.1 & 0.3146 & 0.253 & 0.293 & 0.299 & 0.388 & 0.339 \\
Haiku 4.5 & 142.9 & 0.3680 & 0.294 & 0.310 & 0.434 & 0.440 & 0.361 \\
GPT-5 Nano & 140.8 & 0.6645 & 0.684 & 0.662 & 0.572 & 0.721 & 0.684 \\
Kimi K2 & 140.4 & 0.3402 & 0.285 & 0.207 & 0.431 & 0.361 & 0.418 \\
Gemini 2.5 Flash & 140.3 & 0.3093 & 0.259 & 0.226 & 0.378 & 0.307 & 0.376 \\
Qwen3 235B & 139.6 & 0.4712 & 0.481 & 0.360 & 0.393 & 0.649 & 0.472 \\
GPT-4.1 & 137.5 & 0.2866 & 0.290 & 0.314 & 0.371 & 0.323 & 0.135 \\
DeepSeek V3 & 133.0 & 0.3976 & 0.353 & 0.259 & 0.511 & 0.523 & 0.342 \\
Llama 4 Scout & 130.5 & 0.6067 & 0.581 & 0.763 & 0.431 & 0.623 & 0.635 \\
\midrule Baseline &  & 0.1586 & 0.193 & 0.093 & 0.209 & 0.175 & 0.122 \\
\bottomrule
\end{tabular}

%% file: data/appendix_tables/freeciv_horizon_binary.tex
\setlength{\tabcolsep}{3pt}
\begin{tabular}{lrrrrrrrrrrr}
\toprule
Model & ECI & \multicolumn{5}{c}{Mid-range excess Brier, by horizon (turn)} & \multicolumn{5}{c}{Tail excess bits, by horizon (turn)} \\
\cmidrule(lr){3-7}\cmidrule(lr){8-12}
 & & 90 & 120 & 150 & 180 & 210 & 90 & 120 & 150 & 180 & 210 \\
\midrule
Fable & 162.5 & 0.1671 & 0.1794 & 0.1643 & 0.1589 & 0.1595 & \textbf{0.175} & \textbf{0.208} & \textbf{0.186} & \textbf{0.171} & 0.288 \\
Opus 5 & 161.5 & 0.1508 & 0.1674 & 0.1576 & 0.1467 & 0.1411 & \textbf{0.222} & \textbf{0.236} & \textbf{0.168} & \textbf{0.197} & 0.294 \\
GPT-5.6 Sol & 161.1 & 0.1212 & 0.1139 & 0.1163 & 0.1182 & 0.1157 & \textbf{0.273} & 0.328 & 0.317 & 0.413 & 0.659 \\
GPT-5.5 & 158.6 & 0.1461 & 0.1378 & 0.1321 & 0.1216 & 0.1147 & \textbf{0.171} & \textbf{0.225} & 0.228 & \textbf{0.240} & 0.376 \\
Gemini 3.7 Flash & 156.8 & 0.1019 & 0.1168 & 0.1207 & 0.1013 & 0.0977 & \textbf{0.299} & 0.360 & 0.367 & 0.365 & 0.576 \\
GPT-5.6 Luna & 156.2 & 0.1353 & 0.1391 & 0.1384 & 0.1126 & 0.1218 & \textbf{0.259} & \textbf{0.256} & 0.390 & 0.413 & 0.486 \\
Sonnet 5 & 155.9 & 0.1602 & 0.1779 & 0.1809 & 0.1719 & 0.1758 & \textbf{0.149} & \textbf{0.145} & \textbf{0.151} & \textbf{0.144} & \textbf{0.166} \\
DeepSeek V4 Flash & 154.4 & 0.1534 & 0.1677 & 0.1705 & 0.1510 & 0.1646 & \textbf{0.209} & \textbf{0.192} & \textbf{0.214} & \textbf{0.215} & 0.355 \\
Gemini 3 Flash & 150.7 & 0.1028 & 0.1016 & 0.1137 & 0.1013 & 0.0977 & \textbf{0.318} & 0.457 & 0.331 & 0.400 & 0.482 \\
GPT-5 & 150.0 & 0.1152 & 0.1189 & 0.1237 & 0.1167 & 0.0986 & \textbf{0.283} & 0.326 & 0.370 & 0.451 & 0.519 \\
o3 & 147.1 & 0.1119 & 0.1248 & 0.1221 & 0.1143 & 0.1116 & \textbf{0.284} & \textbf{0.290} & 0.277 & 0.290 & 0.443 \\
GPT-5.4 Nano & 146.5 & 0.1526 & 0.1661 & 0.1719 & 0.1639 & 0.1673 & \textbf{0.260} & \textbf{0.288} & 0.240 & \textbf{0.232} & 0.262 \\
o4-mini & 146.5 & 0.1680 & 0.1734 & 0.1875 & 0.1607 & 0.1496 & \textbf{0.215} & \textbf{0.166} & 0.273 & \textbf{0.233} & 0.269 \\
GPT-5 Mini & 145.8 & 0.1429 & 0.1567 & 0.1647 & 0.1478 & 0.1503 & \textbf{0.221} & \textbf{0.212} & 0.315 & 0.344 & 0.352 \\
Gemini 3.1 Flash Lite & 144.7 & 0.1184 & 0.1235 & 0.1396 & 0.1343 & 0.1421 & \textbf{0.237} & 0.346 & 0.272 & 0.301 & 0.360 \\
Qwen3.5 Flash & 144.1 & 0.1315 & 0.1457 & 0.1317 & 0.1124 & 0.1274 & 0.618 & 0.732 & 0.632 & 0.963 & 0.573 \\
Haiku 4.5 & 142.9 & 0.1056 & 0.1155 & 0.1257 & 0.1063 & 0.1020 & 0.465 & 0.414 & 0.349 & 0.480 & 0.501 \\
GPT-5 Nano & 140.8 & 0.1679 & 0.1976 & 0.1843 & 0.2002 & 0.2085 & \textbf{0.229} & \textbf{0.165} & \textbf{0.195} & \textbf{0.139} & \textbf{0.171} \\
Kimi K2 & 140.4 & 0.1078 & 0.1281 & 0.1279 & 0.1169 & 0.1224 & \textbf{0.297} & 0.398 & 0.424 & 0.413 & 0.495 \\
Gemini 2.5 Flash & 140.3 & 0.1453 & 0.1534 & 0.1567 & 0.1594 & 0.1560 & \textbf{0.233} & 0.401 & 0.484 & 0.594 & 0.554 \\
Qwen3 235B & 139.6 & 0.1249 & 0.1421 & 0.1486 & 0.1390 & 0.1355 & 0.379 & \textbf{0.294} & 0.292 & 0.371 & 0.374 \\
GPT-4.1 & 137.5 & 0.1214 & 0.1576 & 0.1425 & 0.1249 & 0.1341 & \textbf{0.291} & \textbf{0.301} & 0.444 & 0.402 & 0.628 \\
DeepSeek V3 & 133.0 & 0.1243 & 0.1464 & 0.1316 & 0.1246 & 0.1283 & \textbf{0.344} & 0.331 & 0.373 & 0.395 & 0.479 \\
Llama 4 Scout & 130.5 & 0.1915 & 0.2143 & 0.1778 & 0.1796 & 0.2032 & \textbf{0.238} & \textbf{0.146} & 0.334 & 0.302 & \textbf{0.166} \\
\midrule Baseline &  & 0.0962 & 0.0842 & 0.0708 & 0.0609 & 0.0574 & 0.349 & 0.324 & 0.224 & 0.243 & 0.200 \\
\bottomrule
\end{tabular}

%% file: data/appendix_tables/freeciv_horizon_cont.tex
\setlength{\tabcolsep}{3pt}
\begin{tabular}{lrrrrrrrrrr}
\toprule
Model & ECI & \multicolumn{5}{c}{Continuous nCRPS, by horizon (turn)} & \multicolumn{4}{c}{Natural-conditional excess Brier} \\
\cmidrule(lr){3-7}\cmidrule(lr){8-11}
 & & 90 & 120 & 150 & 180 & 210 & 120 & 150 & 180 & 210 \\
\midrule
Fable & 162.5 & 0.389 & 0.781 & 0.760 & 0.950 & 1.079 & 0.1227 & 0.1224 & 0.1275 & 0.0826 \\
Opus 5 & 161.5 & 0.375 & 0.747 & 0.783 & 0.986 & 1.053 & 0.1173 & 0.1172 & 0.1246 & 0.0888 \\
GPT-5.6 Sol & 161.1 & 0.272 & 0.475 & 0.504 & 0.681 & 0.830 & 0.0845 & 0.0995 & 0.1060 & 0.1108 \\
GPT-5.5 & 158.6 & 0.292 & 0.528 & 0.550 & 0.714 & 0.861 & 0.0875 & 0.1022 & 0.0954 & 0.0926 \\
Gemini 3.7 Flash & 156.8 & 0.296 & 0.524 & 0.572 & 0.728 & 0.819 & 0.1207 & 0.1272 & 0.1441 & 0.0934 \\
GPT-5.6 Luna & 156.2 & 0.408 & 0.809 & 0.723 & 0.812 & 0.869 & 0.1030 & 0.1101 & 0.1041 & 0.1060 \\
Sonnet 5 & 155.9 & 0.473 & 0.923 & 0.901 & 1.022 & 1.058 & 0.1079 & 0.1200 & 0.1206 & 0.1029 \\
DeepSeek V4 Flash & 154.4 & 0.363 & 0.737 & 0.644 & 0.810 & 0.883 & 0.1416 & 0.1289 & 0.1251 & 0.1377 \\
Gemini 3 Flash & 150.7 & 0.288 & 0.489 & 0.476 & 0.662 & 0.945 & 0.1302 & 0.1157 & 0.1248 & 0.0846 \\
GPT-5 & 150.0 & 0.361 & 0.793 & 0.788 & 0.952 & 0.984 & 0.0995 & 0.1099 & 0.1089 & 0.1069 \\
o3 & 147.1 & 0.392 & 0.777 & 0.748 & 0.879 & 0.781 & \textbf{0.0806} & 0.0919 & 0.0989 & 0.0843 \\
GPT-5.4 Nano & 146.5 & 0.413 & 0.894 & 0.790 & 0.930 & 0.953 & 0.1138 & 0.1213 & 0.1082 & 0.1058 \\
o4-mini & 146.5 & 0.443 & 0.876 & 0.813 & 0.983 & 1.138 & 0.1333 & 0.1487 & 0.1348 & 0.1651 \\
GPT-5 Mini & 145.8 & 0.530 & 1.064 & 1.052 & 1.254 & 1.299 & 0.1104 & 0.1208 & 0.1152 & 0.1218 \\
Gemini 3.1 Flash Lite & 144.7 & 0.391 & 0.715 & 0.783 & 0.879 & 0.946 & 0.1279 & 0.1200 & 0.1342 & 0.1030 \\
Qwen3.5 Flash & 144.1 & 0.393 & 0.698 & 0.694 & 0.939 & 0.845 & 0.1454 & 0.1546 & 0.1494 & 0.1681 \\
Haiku 4.5 & 142.9 & 0.430 & 0.900 & 0.897 & 0.999 & 0.937 & 0.1110 & 0.1050 & 0.1007 & 0.0956 \\
GPT-5 Nano & 140.8 & 0.710 & 1.288 & 1.239 & 1.511 & 1.653 & 0.1032 & 0.1227 & 0.1017 & 0.1184 \\
Kimi K2 & 140.4 & 0.313 & 0.614 & 0.563 & 0.710 & 0.891 & 0.1097 & 0.1215 & 0.1210 & 0.0977 \\
Gemini 2.5 Flash & 140.3 & 0.438 & 0.843 & 0.827 & 1.020 & 1.004 & 0.2003 & 0.1922 & 0.1672 & 0.1442 \\
Qwen3 235B & 139.6 & 0.509 & 1.036 & 0.930 & 1.136 & 1.182 & 0.1222 & 0.1042 & 0.1011 & 0.1185 \\
GPT-4.1 & 137.5 & 0.448 & 0.883 & 0.858 & 1.031 & 1.110 & 0.1307 & 0.1440 & 0.1456 & 0.1334 \\
DeepSeek V3 & 133.0 & 0.446 & 0.925 & 0.855 & 1.155 & 1.257 & 0.1277 & 0.1202 & 0.1210 & 0.1125 \\
Llama 4 Scout & 130.5 & 0.573 & 1.027 & 0.975 & 1.295 & 1.429 & 0.1111 & 0.1523 & 0.1090 & 0.1211 \\
\midrule Baseline &  & 0.124 & 0.169 & 0.218 & 0.283 & 0.308 & 0.0837 & 0.0570 & 0.0658 & 0.0478 \\
\bottomrule
\end{tabular}

%% file: data/appendix_tables/freeciv_natcond_blocks.tex
\setlength{\tabcolsep}{3pt}
\begin{tabular}{lrrrrrrrrrrr}
\toprule
Model & ECI & \multicolumn{5}{c}{Turn-2 excess Brier, by block} & \multicolumn{5}{c}{Gain over not updating, by block} \\
\cmidrule(lr){3-7}\cmidrule(lr){8-12}
 & & A & B & C1 & C2 & D & A & B & C1 & C2 & D \\
\midrule
Fable & 162.5 & 0.134 & 0.119 & 0.142 & 0.085 & 0.139 & 0.007 & 0.009 & $-$0.073 & $-$0.022 & 0.013 \\
Opus 5 & 161.5 & 0.156 & 0.123 & 0.132 & 0.078 & 0.138 & 0.015 & 0.007 & $-$0.082 & $-$0.003 & 0.020 \\
GPT-5.6 Sol & 161.1 & 0.090 & 0.114 & 0.098 & 0.103 & 0.096 & 0.002 & 0.004 & 0.011 & 0.030 & 0.003 \\
GPT-5.5 & 158.6 & 0.088 & 0.121 & 0.098 & 0.088 & 0.097 & 0.007 & 0.000 & $-$0.014 & 0.035 & 0.010 \\
Gemini 3.7 Flash & 156.8 & 0.108 & 0.099 & 0.152 & 0.134 & 0.095 & $-$0.000 & 0.005 & $-$0.047 & 0.006 & 0.006 \\
GPT-5.6 Luna & 156.2 & 0.121 & 0.107 & 0.126 & 0.108 & 0.085 & 0.004 & $-$0.007 & $-$0.045 & 0.006 & 0.006 \\
Sonnet 5 & 155.9 & 0.151 & 0.127 & 0.121 & 0.075 & 0.155 & 0.012 & 0.009 & $-$0.068 & 0.002 & 0.020 \\
DeepSeek V4 Flash & 154.4 & 0.147 & 0.095 & 0.160 & 0.134 & 0.127 & 0.003 & $-$0.002 & $-$0.069 & 0.011 & $-$0.008 \\
Gemini 3 Flash & 150.7 & 0.083 & 0.138 & 0.152 & 0.117 & 0.090 & 0.001 & $-$0.018 & $-$0.057 & $-$0.001 & 0.001 \\
GPT-5 & 150.0 & 0.123 & 0.138 & 0.125 & 0.100 & 0.089 & 0.001 & $-$0.002 & $-$0.039 & 0.010 & 0.002 \\
o3 & 147.1 & 0.112 & 0.118 & 0.084 & 0.076 & 0.094 & $-$0.009 & 0.002 & $-$0.011 & 0.012 & 0.014 \\
GPT-5.4 Nano & 146.5 & 0.143 & 0.103 & 0.127 & 0.108 & 0.104 & 0.013 & 0.013 & $-$0.072 & $-$0.011 & 0.021 \\
o4-mini & 146.5 & 0.158 & 0.149 & 0.146 & 0.146 & 0.138 & 0.006 & 0.011 & $-$0.019 & 0.006 & 0.015 \\
GPT-5 Mini & 145.8 & 0.166 & 0.146 & 0.142 & 0.105 & 0.095 & 0.007 & $-$0.001 & $-$0.047 & 0.009 & 0.005 \\
Gemini 3.1 Flash Lite & 144.7 & 0.117 & 0.161 & 0.164 & 0.105 & 0.113 & 0.015 & $-$0.002 & $-$0.058 & 0.016 & 0.004 \\
Qwen3.5 Flash & 144.1 & 0.231 & 0.178 & 0.180 & 0.155 & 0.103 & 0.006 & 0.011 & $-$0.061 & $-$0.011 & 0.005 \\
Haiku 4.5 & 142.9 & 0.118 & 0.114 & 0.167 & 0.093 & \textbf{0.076} & 0.020 & $-$0.010 & $-$0.067 & $-$0.005 & 0.012 \\
GPT-5 Nano & 140.8 & 0.141 & 0.156 & 0.139 & 0.086 & 0.114 & 0.014 & 0.004 & $-$0.088 & 0.004 & 0.029 \\
Kimi K2 & 140.4 & 0.130 & 0.128 & 0.129 & 0.104 & 0.106 & 0.019 & $-$0.002 & $-$0.058 & $-$0.008 & 0.018 \\
Gemini 2.5 Flash & 140.3 & 0.165 & 0.124 & 0.232 & 0.163 & 0.190 & 0.005 & 0.002 & $-$0.047 & $-$0.013 & $-$0.025 \\
Qwen3 235B & 139.6 & 0.124 & 0.119 & 0.141 & 0.106 & 0.097 & 0.026 & 0.016 & $-$0.028 & 0.024 & 0.017 \\
GPT-4.1 & 137.5 & 0.124 & 0.143 & 0.184 & 0.142 & 0.110 & 0.009 & 0.008 & $-$0.053 & 0.019 & $-$0.004 \\
DeepSeek V3 & 133.0 & 0.103 & 0.167 & 0.153 & 0.108 & 0.113 & 0.014 & $-$0.008 & $-$0.044 & 0.012 & $-$0.000 \\
Llama 4 Scout & 130.5 & 0.152 & 0.155 & 0.172 & 0.115 & 0.088 & $-$0.000 & 0.029 & $-$0.132 & $-$0.010 & 0.044 \\
\midrule Baseline &  & 0.068 & 0.045 & 0.045 & 0.061 & 0.084 &  &  &  &  &  \\
\bottomrule
\end{tabular}

%% file: data/appendix_tables/freeciv_selection_weights.tex
\setlength{\tabcolsep}{4pt}
\begin{tabular}{@{}lrr@{}}
\toprule
 & Equal weights & Selection weights \\
\midrule
Mean excess Brier over models & 0.141 & 0.143 \\
$\rho$ with ECI [95\% CI] & 0.18 [\ensuremath{-}0.25, 0.55] & 0.19 [\ensuremath{-}0.26, 0.57] \\
Rank correlation with equal weights & --- & 0.98 \\
Constant 0.5, excess Brier & 0.075 & 0.096 \\
Squared bias, share of excess Brier & 29\% & 21\% \\
Discrimination: $\rho$ with ECI [95\% CI] & 0.72 [0.48, 0.85] & 0.47 [0.07, 0.75] \\
Mean $q$ & 0.501 & 0.462 \\
\bottomrule
\end{tabular}

%% file: data/appendix_tables/freeciv_family_difficulty.tex
\begin{tabular}{@{}l r r r r r r r@{}}
\toprule
Family & Items & T90 & T120 & T150 & T180 & T210 & All \\
\midrule
\multicolumn{8}{@{}l}{\emph{Mid-range: excess Brier score}} \\
S7\_any\_destroyed & 2 & 0.448 & -- & -- & -- & -- & 0.448 \\
S4\_gov\_change\_count & 32 & 0.204 & 0.266 & 0.258 & 0.152 & 0.151 & 0.213 \\
W4\_lose\_k & 35 & 0.128 & 0.167 & 0.232 & 0.240 & 0.378 & 0.211 \\
S3\_wars\_at\_T & 3 & 0.206 & -- & -- & -- & -- & 0.206 \\
EX\_government\_at & 46 & 0.085 & 0.177 & 0.278 & 0.199 & 0.194 & 0.191 \\
NB1\_threshold & 48 & 0.207 & 0.194 & 0.170 & 0.190 & 0.180 & 0.187 \\
W3\_capture\_k & 38 & 0.071 & 0.165 & 0.214 & 0.234 & 0.244 & 0.180 \\
NB6\_event & 7 & 0.091 & 0.393 & -- & -- & -- & 0.177 \\
W1\_wonder\_race & 36 & 0.195 & 0.202 & 0.171 & 0.157 & 0.159 & 0.175 \\
NW2\_diplo & 44 & 0.176 & 0.112 & 0.153 & 0.125 & 0.169 & 0.146 \\
W2\_directed\_conquest & 49 & 0.133 & 0.147 & 0.162 & 0.139 & 0.123 & 0.141 \\
EX\_tech\_discovered & 50 & 0.103 & 0.159 & 0.145 & 0.142 & 0.156 & 0.141 \\
NB4\_drawdown & 44 & 0.092 & 0.116 & 0.162 & 0.133 & 0.153 & 0.139 \\
NW1\_war\_at & 49 & 0.126 & 0.124 & 0.098 & 0.156 & 0.142 & 0.130 \\
NB1\_value\_threshold & 51 & 0.121 & 0.164 & 0.118 & 0.143 & 0.103 & 0.129 \\
W6\_peace\_at & 12 & 0.101 & 0.096 & 0.335 & -- & 0.077 & 0.117 \\
NW5\_wonder & 34 & 0.112 & 0.109 & 0.109 & 0.111 & 0.102 & 0.108 \\
S6\_city\_founding & 34 & 0.135 & 0.112 & 0.087 & 0.078 & 0.092 & 0.102 \\
S5\_civ\_wonders & 43 & 0.155 & 0.112 & 0.098 & 0.082 & 0.059 & 0.101 \\
W5\_tech\_lead & 23 & 0.133 & 0.069 & 0.056 & 0.036 & 0.017 & 0.070 \\
EX\_comparative & 53 & 0.068 & 0.110 & 0.038 & 0.066 & 0.052 & 0.067 \\
NEW\_civil\_war & 17 & -- & 0.006 & 0.016 & 0.049 & 0.057 & 0.039 \\
\emph{all families} & 750 & 0.136 & 0.149 & 0.147 & 0.137 & 0.139 & 0.141 \\
\midrule
\multicolumn{8}{@{}l}{\emph{Tail set: excess bits}} \\
EX\_government\_at & 22 & 0.770 & 0.567 & 0.704 & 0.482 & 0.614 & 0.624 \\
W6\_peace\_at & 20 & 0.523 & 0.572 & 0.505 & 0.564 & 0.556 & 0.546 \\
NW5\_wonder & 21 & 0.178 & 0.250 & 0.419 & 0.665 & 0.974 & 0.543 \\
NW1\_war\_at & 15 & 0.543 & 0.367 & 0.287 & 0.384 & 0.411 & 0.395 \\
NW4\_survival & 16 & 0.326 & 0.469 & 0.204 & 0.431 & 0.571 & 0.393 \\
NW2\_diplo & 21 & 0.175 & 0.529 & 0.441 & 0.342 & 0.339 & 0.372 \\
NB1\_value\_threshold & 20 & 0.144 & 0.370 & 0.349 & 0.430 & 0.438 & 0.371 \\
EX\_tech\_discovered & 22 & 0.256 & 0.311 & 0.326 & 0.504 & 0.338 & 0.357 \\
S6\_city\_founding & 2 & 0.297 & -- & -- & -- & -- & 0.297 \\
NB6\_event & 3 & 0.280 & -- & -- & -- & -- & 0.280 \\
W4\_lose\_k & 3 & 0.265 & -- & -- & -- & -- & 0.265 \\
S4\_gov\_change\_count & 6 & 0.285 & 0.172 & -- & -- & -- & 0.248 \\
S5\_civ\_wonders & 21 & 0.157 & 0.203 & 0.191 & 0.253 & 0.329 & 0.236 \\
NB4\_drawdown & 20 & 0.173 & 0.207 & 0.318 & 0.184 & 0.260 & 0.233 \\
W5\_tech\_lead & 15 & 0.188 & 0.261 & 0.220 & 0.036 & 0.314 & 0.213 \\
W3\_capture\_k & 3 & 0.210 & -- & -- & -- & -- & 0.210 \\
W2\_directed\_conquest & 20 & 0.197 & 0.171 & 0.257 & 0.225 & 0.191 & 0.209 \\
W1\_wonder\_race & 21 & 0.158 & 0.160 & 0.151 & 0.231 & 0.250 & 0.193 \\
EX\_comparative & 9 & 0.245 & 0.130 & 0.121 & -- & -- & 0.180 \\
NEW\_civil\_war & 20 & 0.065 & 0.090 & 0.039 & 0.047 & 0.049 & 0.057 \\
\emph{all families} & 300 & 0.278 & 0.301 & 0.318 & 0.353 & 0.410 & 0.332 \\
\midrule
\multicolumn{8}{@{}l}{\emph{Continuous set: normalised CRPS}} \\
P6\_world\_techs & 20 & 2.037 & 3.931 & 3.962 & 4.029 & 4.736 & 3.463 \\
P5\_techs\_at\_T & 31 & 0.781 & 1.357 & 1.728 & 2.029 & 1.795 & 1.546 \\
NC14\_world\_wonders & 31 & 0.156 & 0.338 & 0.706 & 1.063 & 1.504 & 0.763 \\
NC5\_world\_captures & 32 & 0.085 & 0.430 & 0.762 & 0.935 & 1.184 & 0.697 \\
P2\_civ\_losses & 31 & 0.103 & 0.280 & 0.289 & 0.860 & 0.704 & 0.456 \\
S7\_world\_razings & 33 & 0.154 & 0.317 & 0.500 & 0.574 & 0.648 & 0.451 \\
NC1\_value\_at\_T & 60 & 0.225 & 0.395 & 0.435 & 0.489 & 0.691 & 0.447 \\
P3\_civ\_founds & 31 & 0.311 & 0.428 & 0.461 & 0.500 & 0.424 & 0.427 \\
P1\_civ\_conquests & 31 & 0.076 & 0.203 & 0.280 & 0.325 & 0.422 & 0.262 \\
\emph{all families} & 300 & 0.415 & 0.807 & 0.784 & 0.964 & 1.036 & 0.801 \\
\bottomrule
\end{tabular}

%% file: data/combined_matched_panel.tex
\begin{tabular}{lrr}
\toprule
Comparator & $\rho$ & Models, 95\% \\
\midrule
ECI & $0.794$ & $[0.502,0.939]$ \\
ForecastBench & $0.374$ & $[-0.216,0.815]$ \\
ForecastBench, given ECI & $-0.249$ & $[-0.702,0.487]$ \\
\bottomrule
\end{tabular}

%% file: data/combined_score_table.tex
\begin{tabular}{@{}l r r@{}}
\toprule
Model & ECI & $\theta$ \\
\midrule
GPT-5.6 Sol & 161.1 & $+0.42$ \\
GPT-5.5 & 158.6 & $+0.42$ \\
Opus 5 & 161.5 & $+0.32$ \\
DeepSeek V4 Flash & 154.4 & $+0.26$ \\
Fable & 162.5 & $+0.25$ \\
o3 & 147.1 & $+0.24$ \\
Gemini 3.7 Flash & 156.8 & $+0.22$ \\
GPT-5.6 Luna & 156.2 & $+0.20$ \\
Sonnet 5 & 155.9 & $+0.18$ \\
Gemini 3.1 Flash Lite & 144.7 & $+0.10$ \\
Gemini 3 Flash & 150.7 & $+0.08$ \\
GPT-5.4 Nano & 146.5 & $+0.07$ \\
GPT-5 & 150.0 & $+0.06$ \\
Kimi K2 & 140.4 & $-0.07$ \\
GPT-4.1 & 137.5 & $-0.08$ \\
o4-mini & 146.5 & $-0.09$ \\
GPT-5 Mini & 145.8 & $-0.12$ \\
Haiku 4.5 & 142.9 & $-0.19$ \\
Gemini 2.5 Flash & 140.3 & $-0.22$ \\
Llama 4 Scout & 130.5 & $-0.36$ \\
Qwen3 235B & 139.6 & $-0.39$ \\
DeepSeek V3 & 133.0 & $-0.42$ \\
GPT-5 Nano & 140.8 & $-0.43$ \\
Qwen3.5 Flash & 144.1 & $-0.46$ \\
\bottomrule
\end{tabular}

%% file: data/rank_table.tex
\setlength{\tabcolsep}{3pt}
\begin{tabular}{@{}l r r rrr rrrr rr@{}}
\toprule
 & & & \multicolumn{3}{c}{Micropolis} & \multicolumn{4}{c}{FreeCiv} & \multicolumn{2}{c}{Starsim} \\
\cmidrule(lr){4-6}\cmidrule(lr){7-10}\cmidrule(l){11-12}
 & & Avg & Mid- &  &  & Mid- &  &  & Nat. &  &  \\
Model & $\theta$ & Rank & range & Tail & Cont. & range & Tail & Cont. & cond. & Cont. & Interv. \\
\midrule
GPT-5.6 Sol & \cellcolor[RGB]{27,120,55}\color{white}\textbf{{\boldmath$+0.42$}} & \cellcolor[RGB]{43,134,66}\color{white}\textbf{2} & \cellcolor[RGB]{27,120,55}\color{white}\textbf{1} & \cellcolor[RGB]{60,148,77}\textbf{3} & \cellcolor[RGB]{60,148,77}\textbf{3} & \cellcolor[RGB]{113,188,116}\textbf{6} & \cellcolor[RGB]{165,128,182}\textbf{19} & \cellcolor[RGB]{27,120,55}\color{white}\textbf{1} & \cellcolor[RGB]{60,148,77}\textbf{3} & \cellcolor[RGB]{93,176,100}\textbf{5} & \cellcolor[RGB]{27,120,55}\color{white}\textbf{1} \\
GPT-5.5 & \cellcolor[RGB]{27,120,55}\color{white}\textbf{{\boldmath$+0.42$}} & \cellcolor[RGB]{27,120,55}\color{white}\textbf{1} & \cellcolor[RGB]{76,162,88}\textbf{4} & \cellcolor[RGB]{27,120,55}\color{white}\textbf{1} & \cellcolor[RGB]{43,134,66}\color{white}\textbf{2} & \cellcolor[RGB]{194,229,190}\textbf{10} & \cellcolor[RGB]{153,211,149}\textbf{8} & \cellcolor[RGB]{76,162,88}\textbf{4} & \cellcolor[RGB]{43,134,66}\color{white}\textbf{2} & \cellcolor[RGB]{133,199,133}\textbf{7} & \cellcolor[RGB]{61,149,78}\textbf{3} \\
Opus 5 & \cellcolor[RGB]{68,155,82}\textbf{{\boldmath$+0.32$}} & \cellcolor[RGB]{60,148,77}\textbf{3} & \cellcolor[RGB]{60,148,77}\textbf{3} & \cellcolor[RGB]{93,176,100}\textbf{5} & \cellcolor[RGB]{76,162,88}\textbf{4} & \cellcolor[RGB]{199,172,210}\textbf{16} & \cellcolor[RGB]{76,162,88}\textbf{4} & \cellcolor[RGB]{236,243,236}\textbf{12} & \cellcolor[RGB]{173,221,168}\textbf{9} & \cellcolor[RGB]{60,148,77}\textbf{3} & \cellcolor[RGB]{79,164,89}\textbf{4} \\
DeepSeek V4 Flash & \cellcolor[RGB]{95,177,101}\textbf{{\boldmath$+0.26$}} & \cellcolor[RGB]{113,188,116}\textbf{6} & \cellcolor[RGB]{43,134,66}\color{white}\textbf{2} & \cellcolor[RGB]{76,162,88}\textbf{4} & \cellcolor[RGB]{236,243,236}\textbf{12} & \cellcolor[RGB]{176,142,191}\textbf{18} & \cellcolor[RGB]{113,188,116}\textbf{6} & \cellcolor[RGB]{113,188,116}\textbf{6} & \cellcolor[RGB]{155,114,173}\textbf{20} & \cellcolor[RGB]{43,134,66}\color{white}\textbf{2} & \textbf{--} \\
Fable & \cellcolor[RGB]{104,183,109}\textbf{{\boldmath$+0.25$}} & \cellcolor[RGB]{93,176,100}\textbf{5} & \cellcolor[RGB]{153,211,149}\textbf{8} & \cellcolor[RGB]{113,188,116}\textbf{6} & \cellcolor[RGB]{27,120,55}\color{white}\textbf{1} & \cellcolor[RGB]{155,114,173}\textbf{20} & \cellcolor[RGB]{60,148,77}\textbf{3} & \cellcolor[RGB]{240,236,242}\textbf{13} & \cellcolor[RGB]{240,236,242}\textbf{13} & \cellcolor[RGB]{76,162,88}\textbf{4} & \cellcolor[RGB]{159,215,154}\textbf{8} \\
o3 & \cellcolor[RGB]{108,185,112}\textbf{{\boldmath$+0.24$}} & \cellcolor[RGB]{76,162,88}\textbf{4} & \cellcolor[RGB]{113,188,116}\textbf{6} & \cellcolor[RGB]{194,229,190}\textbf{10} & \cellcolor[RGB]{173,221,168}\textbf{9} & \cellcolor[RGB]{93,176,100}\textbf{5} & \cellcolor[RGB]{236,243,236}\textbf{12} & \cellcolor[RGB]{153,211,149}\textbf{8} & \cellcolor[RGB]{27,120,55}\color{white}\textbf{1} & \cellcolor[RGB]{113,188,116}\textbf{6} & \cellcolor[RGB]{97,178,103}\textbf{5} \\
Gemini 3.7 Flash & \cellcolor[RGB]{116,190,119}\textbf{{\boldmath$+0.22$}} & \cellcolor[RGB]{133,199,133}\textbf{7} & \cellcolor[RGB]{93,176,100}\textbf{5} & \cellcolor[RGB]{43,134,66}\color{white}\textbf{2} & \cellcolor[RGB]{113,188,116}\textbf{6} & \cellcolor[RGB]{43,134,66}\color{white}\textbf{2} & \cellcolor[RGB]{187,156,201}\textbf{17} & \cellcolor[RGB]{60,148,77}\textbf{3} & \cellcolor[RGB]{176,142,191}\textbf{18} & \cellcolor[RGB]{173,221,168}\textbf{9} & \cellcolor[RGB]{179,146,194}\textbf{17} \\
GPT-5.6 Luna & \cellcolor[RGB]{126,195,127}\textbf{{\boldmath$+0.20$}} & \cellcolor[RGB]{153,211,149}\textbf{8} & \cellcolor[RGB]{215,236,213}\textbf{11} & \cellcolor[RGB]{240,236,242}\textbf{13} & \cellcolor[RGB]{153,211,149}\textbf{8} & \cellcolor[RGB]{153,211,149}\textbf{8} & \cellcolor[RGB]{226,215,231}\textbf{14} & \cellcolor[RGB]{173,221,168}\textbf{9} & \cellcolor[RGB]{93,176,100}\textbf{5} & \cellcolor[RGB]{215,236,213}\textbf{11} & \cellcolor[RGB]{44,135,66}\color{white}\textbf{2} \\
Sonnet 5 & \cellcolor[RGB]{139,203,137}\textbf{{\boldmath$+0.18$}} & \cellcolor[RGB]{226,240,224}\textbf{11.5} & \cellcolor[RGB]{173,221,168}\textbf{9} & \cellcolor[RGB]{236,243,236}\textbf{12} & \cellcolor[RGB]{93,176,100}\textbf{5} & \cellcolor[RGB]{136,79,152}\color{white}\textbf{22} & \cellcolor[RGB]{27,120,55}\color{white}\textbf{1} & \cellcolor[RGB]{165,128,182}\textbf{19} & \cellcolor[RGB]{236,243,236}\textbf{12} & \cellcolor[RGB]{236,243,236}\textbf{12} & \cellcolor[RGB]{118,190,120}\textbf{6} \\
Gemini 3.1 Flash Lite & \cellcolor[RGB]{178,223,173}\textbf{{\boldmath$+0.10$}} & \cellcolor[RGB]{226,240,224}\textbf{11.5} & \cellcolor[RGB]{236,243,236}\textbf{12} & \cellcolor[RGB]{215,236,213}\textbf{11} & \cellcolor[RGB]{194,229,190}\textbf{10} & \cellcolor[RGB]{236,243,236}\textbf{12} & \cellcolor[RGB]{215,236,213}\textbf{11} & \cellcolor[RGB]{194,229,190}\textbf{10} & \cellcolor[RGB]{187,156,201}\textbf{17} & \cellcolor[RGB]{153,211,149}\textbf{8} & \cellcolor[RGB]{138,203,137}\textbf{7} \\
Gemini 3 Flash & \cellcolor[RGB]{191,228,187}\textbf{{\boldmath$+0.08$}} & \cellcolor[RGB]{194,229,190}\textbf{10} & \cellcolor[RGB]{133,199,133}\textbf{7} & \cellcolor[RGB]{212,194,221}\textbf{15} & \cellcolor[RGB]{215,236,213}\textbf{11} & \cellcolor[RGB]{27,120,55}\color{white}\textbf{1} & \cellcolor[RGB]{176,142,191}\textbf{18} & \cellcolor[RGB]{43,134,66}\color{white}\textbf{2} & \cellcolor[RGB]{226,215,231}\textbf{14} & \cellcolor[RGB]{226,215,231}\textbf{14} & \cellcolor[RGB]{204,180,214}\textbf{15} \\
GPT-5.4 Nano & \cellcolor[RGB]{196,229,192}\textbf{{\boldmath$+0.07$}} & \cellcolor[RGB]{240,236,242}\textbf{13} & \cellcolor[RGB]{212,194,221}\textbf{15} & \cellcolor[RGB]{199,172,210}\textbf{16} & \cellcolor[RGB]{212,194,221}\textbf{15} & \cellcolor[RGB]{165,128,182}\textbf{19} & \cellcolor[RGB]{173,221,168}\textbf{9} & \cellcolor[RGB]{226,215,231}\textbf{14} & \cellcolor[RGB]{194,229,190}\textbf{10} & \cellcolor[RGB]{27,120,55}\color{white}\textbf{1} & \cellcolor[RGB]{168,131,184}\textbf{18} \\
GPT-5 & \cellcolor[RGB]{200,231,197}\textbf{{\boldmath$+0.06$}} & \cellcolor[RGB]{173,221,168}\textbf{9} & \cellcolor[RGB]{194,229,190}\textbf{10} & \cellcolor[RGB]{153,211,149}\textbf{8} & \cellcolor[RGB]{133,199,133}\textbf{7} & \cellcolor[RGB]{76,162,88}\textbf{4} & \cellcolor[RGB]{199,172,210}\textbf{16} & \cellcolor[RGB]{215,236,213}\textbf{11} & \cellcolor[RGB]{113,188,116}\textbf{6} & \cellcolor[RGB]{187,156,201}\textbf{17} & \cellcolor[RGB]{181,224,176}\textbf{9} \\
Kimi K2 & \cellcolor[RGB]{231,222,235}\textbf{{\boldmath$-0.07$}} & \cellcolor[RGB]{226,215,231}\textbf{14} & \cellcolor[RGB]{240,236,242}\textbf{13} & \cellcolor[RGB]{155,114,173}\textbf{20} & \cellcolor[RGB]{176,142,191}\textbf{18} & \cellcolor[RGB]{133,199,133}\textbf{7} & \cellcolor[RGB]{155,114,173}\textbf{20} & \cellcolor[RGB]{93,176,100}\textbf{5} & \cellcolor[RGB]{215,236,213}\textbf{11} & \cellcolor[RGB]{199,172,210}\textbf{16} & \cellcolor[RGB]{190,160,204}\textbf{16} \\
GPT-4.1 & \cellcolor[RGB]{227,215,232}\textbf{{\boldmath$-0.08$}} & \cellcolor[RGB]{187,156,201}\textbf{17} & \cellcolor[RGB]{155,114,173}\textbf{20} & \cellcolor[RGB]{173,221,168}\textbf{9} & \cellcolor[RGB]{187,156,201}\textbf{17} & \cellcolor[RGB]{240,236,242}\textbf{13} & \cellcolor[RGB]{145,97,162}\color{white}\textbf{21} & \cellcolor[RGB]{176,142,191}\textbf{18} & \cellcolor[RGB]{145,97,162}\color{white}\textbf{21} & \cellcolor[RGB]{212,194,221}\textbf{15} & \cellcolor[RGB]{233,225,236}\textbf{13} \\
o4-mini & \cellcolor[RGB]{223,211,229}\textbf{{\boldmath$-0.09$}} & \cellcolor[RGB]{176,142,191}\textbf{18} & \cellcolor[RGB]{145,97,162}\color{white}\textbf{21} & \cellcolor[RGB]{187,156,201}\textbf{17} & \cellcolor[RGB]{165,128,182}\textbf{19} & \cellcolor[RGB]{145,97,162}\color{white}\textbf{21} & \cellcolor[RGB]{93,176,100}\textbf{5} & \cellcolor[RGB]{187,156,201}\textbf{17} & \cellcolor[RGB]{136,79,152}\color{white}\textbf{22} & \cellcolor[RGB]{194,229,190}\textbf{10} & \cellcolor[RGB]{147,99,164}\color{white}\textbf{20} \\
GPT-5 Mini & \cellcolor[RGB]{213,194,221}\textbf{{\boldmath$-0.12$}} & \cellcolor[RGB]{199,172,210}\textbf{16} & \cellcolor[RGB]{187,156,201}\textbf{17} & \cellcolor[RGB]{133,199,133}\textbf{7} & \cellcolor[RGB]{155,114,173}\textbf{20} & \cellcolor[RGB]{212,194,221}\textbf{15} & \cellcolor[RGB]{194,229,190}\textbf{10} & \cellcolor[RGB]{136,79,152}\color{white}\textbf{22} & \cellcolor[RGB]{212,194,221}\textbf{15} & \cellcolor[RGB]{165,128,182}\textbf{19} & \cellcolor[RGB]{247,247,247}\textbf{12} \\
Haiku 4.5 & \cellcolor[RGB]{187,156,201}\textbf{{\boldmath$-0.19$}} & \cellcolor[RGB]{212,194,221}\textbf{15} & \cellcolor[RGB]{226,215,231}\textbf{14} & \cellcolor[RGB]{165,128,182}\textbf{19} & \cellcolor[RGB]{199,172,210}\textbf{16} & \cellcolor[RGB]{60,148,77}\textbf{3} & \cellcolor[RGB]{136,79,152}\color{white}\textbf{22} & \cellcolor[RGB]{199,172,210}\textbf{16} & \cellcolor[RGB]{76,162,88}\textbf{4} & \cellcolor[RGB]{155,114,173}\textbf{20} & \cellcolor[RGB]{218,202,225}\textbf{14} \\
Gemini 2.5 Flash & \cellcolor[RGB]{180,146,194}\textbf{{\boldmath$-0.22$}} & \cellcolor[RGB]{141,88,157}\color{white}\textbf{21.5} & \cellcolor[RGB]{127,60,141}\color{white}\textbf{23} & \cellcolor[RGB]{136,79,152}\color{white}\textbf{22} & \cellcolor[RGB]{226,215,231}\textbf{14} & \cellcolor[RGB]{187,156,201}\textbf{17} & \cellcolor[RGB]{127,60,141}\color{white}\textbf{23} & \cellcolor[RGB]{212,194,221}\textbf{15} & \cellcolor[RGB]{118,42,131}\color{white}\textbf{24} & \cellcolor[RGB]{240,236,242}\textbf{13} & \cellcolor[RGB]{203,232,200}\textbf{10} \\
Llama 4 Scout & \cellcolor[RGB]{142,89,158}\color{white}\textbf{{\boldmath$-0.36$}} & \cellcolor[RGB]{118,42,131}\color{white}\textbf{24} & \cellcolor[RGB]{165,128,182}\textbf{19} & \cellcolor[RGB]{176,142,191}\textbf{18} & \cellcolor[RGB]{145,97,162}\color{white}\textbf{21} & \cellcolor[RGB]{118,42,131}\color{white}\textbf{24} & \cellcolor[RGB]{133,199,133}\textbf{7} & \cellcolor[RGB]{127,60,141}\color{white}\textbf{23} & \cellcolor[RGB]{165,128,182}\textbf{19} & \cellcolor[RGB]{118,42,131}\color{white}\textbf{24} & \cellcolor[RGB]{157,117,174}\textbf{19} \\
Qwen3 235B & \cellcolor[RGB]{135,76,151}\color{white}\textbf{{\boldmath$-0.39$}} & \cellcolor[RGB]{165,128,182}\textbf{19} & \cellcolor[RGB]{199,172,210}\textbf{16} & \cellcolor[RGB]{145,97,162}\color{white}\textbf{21} & \cellcolor[RGB]{136,79,152}\color{white}\textbf{22} & \cellcolor[RGB]{226,215,231}\textbf{14} & \cellcolor[RGB]{240,236,242}\textbf{13} & \cellcolor[RGB]{145,97,162}\color{white}\textbf{21} & \cellcolor[RGB]{153,211,149}\textbf{8} & \cellcolor[RGB]{136,79,152}\color{white}\textbf{22} & \cellcolor[RGB]{137,80,153}\color{white}\textbf{21} \\
DeepSeek V3 & \cellcolor[RGB]{130,65,144}\color{white}\textbf{{\boldmath$-0.42$}} & \cellcolor[RGB]{155,114,173}\textbf{20} & \cellcolor[RGB]{176,142,191}\textbf{18} & \cellcolor[RGB]{127,60,141}\color{white}\textbf{23} & \cellcolor[RGB]{240,236,242}\textbf{13} & \cellcolor[RGB]{215,236,213}\textbf{11} & \cellcolor[RGB]{212,194,221}\textbf{15} & \cellcolor[RGB]{155,114,173}\textbf{20} & \cellcolor[RGB]{199,172,210}\textbf{16} & \cellcolor[RGB]{145,97,162}\color{white}\textbf{21} & \cellcolor[RGB]{118,42,131}\color{white}\textbf{23} \\
GPT-5 Nano & \cellcolor[RGB]{126,58,140}\color{white}\textbf{{\boldmath$-0.43$}} & \cellcolor[RGB]{141,88,157}\color{white}\textbf{21.5} & \cellcolor[RGB]{136,79,152}\color{white}\textbf{22} & \cellcolor[RGB]{226,215,231}\textbf{14} & \cellcolor[RGB]{118,42,131}\color{white}\textbf{24} & \cellcolor[RGB]{127,60,141}\color{white}\textbf{23} & \cellcolor[RGB]{43,134,66}\color{white}\textbf{2} & \cellcolor[RGB]{118,42,131}\color{white}\textbf{24} & \cellcolor[RGB]{133,199,133}\textbf{7} & \cellcolor[RGB]{127,60,141}\color{white}\textbf{23} & \cellcolor[RGB]{128,61,142}\color{white}\textbf{22} \\
Qwen3.5 Flash & \cellcolor[RGB]{118,42,131}\color{white}\textbf{{\boldmath$-0.46$}} & \cellcolor[RGB]{127,60,141}\color{white}\textbf{23} & \cellcolor[RGB]{118,42,131}\color{white}\textbf{24} & \cellcolor[RGB]{118,42,131}\color{white}\textbf{24} & \cellcolor[RGB]{127,60,141}\color{white}\textbf{23} & \cellcolor[RGB]{173,221,168}\textbf{9} & \cellcolor[RGB]{118,42,131}\color{white}\textbf{24} & \cellcolor[RGB]{133,199,133}\textbf{7} & \cellcolor[RGB]{127,60,141}\color{white}\textbf{23} & \cellcolor[RGB]{176,142,191}\textbf{18} & \cellcolor[RGB]{225,239,223}\textbf{11} \\
\bottomrule
\end{tabular}

%% file: data/rank_correlation_table.tex
\setlength{\tabcolsep}{3pt}
\begin{tabular}{@{}l r rrr rrrr rr@{}}
\toprule
 & & \multicolumn{3}{c}{Micropolis} & \multicolumn{4}{c}{FreeCiv} & \multicolumn{2}{c}{Starsim} \\
\cmidrule(lr){3-5}\cmidrule(lr){6-9}\cmidrule(l){10-11}
 & & Mid- &  &  & Mid- &  &  & Nat. &  &  \\
 & $\theta_m$ & range & Tail & Cont. & range & Tail & Cont. & cond. & Cont. & Interv. \\
\midrule
$\theta_m$ &  & \cellcolor[RGB]{43,134,66}\color{white}\textbf{{\boldmath$.92$}} & \cellcolor[RGB]{61,149,78}\textbf{{\boldmath$.82$}} & \cellcolor[RGB]{52,142,72}\color{white}\textbf{{\boldmath$.87$}} & \cellcolor[RGB]{197,230,194}\textbf{{\boldmath$.20$}} & \cellcolor[RGB]{180,224,176}\textbf{{\boldmath$.27$}} & \cellcolor[RGB]{94,176,100}\textbf{{\boldmath$.65$}} & \cellcolor[RGB]{153,211,149}\textbf{{\boldmath$.39$}} & \cellcolor[RGB]{56,145,75}\textbf{{\boldmath$.85$}} & \cellcolor[RGB]{76,162,88}\textbf{{\boldmath$.74$}} \\
\midrule
Micropolis mid-range & \cellcolor[RGB]{43,134,66}\color{white}\textbf{{\boldmath$.92$}} &  & \cellcolor[RGB]{73,159,86}\textbf{{\boldmath$.76$}} & \cellcolor[RGB]{62,150,78}\textbf{{\boldmath$.82$}} & \cellcolor[RGB]{158,214,154}\textbf{{\boldmath$.37$}} & \cellcolor[RGB]{195,229,191}\textbf{{\boldmath$.21$}} & \cellcolor[RGB]{91,175,98}\textbf{{\boldmath$.66$}} & \cellcolor[RGB]{128,197,129}\textbf{{\boldmath$.50$}} & \cellcolor[RGB]{91,175,98}\textbf{{\boldmath$.66$}} & \cellcolor[RGB]{103,181,107}\textbf{{\boldmath$.61$}} \\
Micropolis tail & \cellcolor[RGB]{61,149,78}\textbf{{\boldmath$.82$}} & \cellcolor[RGB]{73,159,86}\textbf{{\boldmath$.76$}} &  & \cellcolor[RGB]{87,171,95}\textbf{{\boldmath$.68$}} & \cellcolor[RGB]{221,238,219}\textbf{{\boldmath$.11$}} & \cellcolor[RGB]{162,217,157}\textbf{{\boldmath$.35$}} & \cellcolor[RGB]{151,210,147}\textbf{{\boldmath$.40$}} & \cellcolor[RGB]{174,222,169}\textbf{{\boldmath$.30$}} & \cellcolor[RGB]{105,183,110}\textbf{{\boldmath$.60$}} & \cellcolor[RGB]{111,186,114}\textbf{{\boldmath$.58$}} \\
Micropolis cont. & \cellcolor[RGB]{52,142,72}\color{white}\textbf{{\boldmath$.87$}} & \cellcolor[RGB]{62,150,78}\textbf{{\boldmath$.82$}} & \cellcolor[RGB]{87,171,95}\textbf{{\boldmath$.68$}} &  & \cellcolor[RGB]{182,224,177}\textbf{{\boldmath$.27$}} & \cellcolor[RGB]{203,232,199}\textbf{{\boldmath$.18$}} & \cellcolor[RGB]{118,190,120}\textbf{{\boldmath$.55$}} & \cellcolor[RGB]{155,213,151}\textbf{{\boldmath$.38$}} & \cellcolor[RGB]{87,171,95}\textbf{{\boldmath$.68$}} & \cellcolor[RGB]{82,167,92}\textbf{{\boldmath$.71$}} \\
FreeCiv mid-range & \cellcolor[RGB]{197,230,194}\textbf{{\boldmath$.20$}} & \cellcolor[RGB]{158,214,154}\textbf{{\boldmath$.37$}} & \cellcolor[RGB]{221,238,219}\textbf{{\boldmath$.11$}} & \cellcolor[RGB]{182,224,177}\textbf{{\boldmath$.27$}} &  & \cellcolor[RGB]{149,104,167}\color{white}\textbf{{\boldmath$-.70$}} & \cellcolor[RGB]{91,174,97}\textbf{{\boldmath$.66$}} & \cellcolor[RGB]{163,217,158}\textbf{{\boldmath$.35$}} & \cellcolor[RGB]{247,247,247}\textbf{{\boldmath$.00$}} & \cellcolor[RGB]{189,227,185}\textbf{{\boldmath$.24$}} \\
FreeCiv tail & \cellcolor[RGB]{180,224,176}\textbf{{\boldmath$.27$}} & \cellcolor[RGB]{195,229,191}\textbf{{\boldmath$.21$}} & \cellcolor[RGB]{162,217,157}\textbf{{\boldmath$.35$}} & \cellcolor[RGB]{203,232,199}\textbf{{\boldmath$.18$}} & \cellcolor[RGB]{149,104,167}\color{white}\textbf{{\boldmath$-.70$}} &  & \cellcolor[RGB]{192,162,205}\textbf{{\boldmath$-.35$}} & \cellcolor[RGB]{216,236,214}\textbf{{\boldmath$.13$}} & \cellcolor[RGB]{181,224,177}\textbf{{\boldmath$.27$}} & \cellcolor[RGB]{247,247,247}\textbf{{\boldmath$.00$}} \\
FreeCiv cont. & \cellcolor[RGB]{94,176,100}\textbf{{\boldmath$.65$}} & \cellcolor[RGB]{91,175,98}\textbf{{\boldmath$.66$}} & \cellcolor[RGB]{151,210,147}\textbf{{\boldmath$.40$}} & \cellcolor[RGB]{118,190,120}\textbf{{\boldmath$.55$}} & \cellcolor[RGB]{91,174,97}\textbf{{\boldmath$.66$}} & \cellcolor[RGB]{192,162,205}\textbf{{\boldmath$-.35$}} &  & \cellcolor[RGB]{202,232,199}\textbf{{\boldmath$.18$}} & \cellcolor[RGB]{111,187,115}\textbf{{\boldmath$.57$}} & \cellcolor[RGB]{121,192,123}\textbf{{\boldmath$.53$}} \\
FreeCiv nat.\ cond. & \cellcolor[RGB]{153,211,149}\textbf{{\boldmath$.39$}} & \cellcolor[RGB]{128,197,129}\textbf{{\boldmath$.50$}} & \cellcolor[RGB]{174,222,169}\textbf{{\boldmath$.30$}} & \cellcolor[RGB]{155,213,151}\textbf{{\boldmath$.38$}} & \cellcolor[RGB]{163,217,158}\textbf{{\boldmath$.35$}} & \cellcolor[RGB]{216,236,214}\textbf{{\boldmath$.13$}} & \cellcolor[RGB]{202,232,199}\textbf{{\boldmath$.18$}} &  & \cellcolor[RGB]{217,237,215}\textbf{{\boldmath$.12$}} & \cellcolor[RGB]{150,210,147}\textbf{{\boldmath$.40$}} \\
Starsim cont. & \cellcolor[RGB]{56,145,75}\textbf{{\boldmath$.85$}} & \cellcolor[RGB]{91,175,98}\textbf{{\boldmath$.66$}} & \cellcolor[RGB]{105,183,110}\textbf{{\boldmath$.60$}} & \cellcolor[RGB]{87,171,95}\textbf{{\boldmath$.68$}} & \cellcolor[RGB]{247,247,247}\textbf{{\boldmath$.00$}} & \cellcolor[RGB]{181,224,177}\textbf{{\boldmath$.27$}} & \cellcolor[RGB]{111,187,115}\textbf{{\boldmath$.57$}} & \cellcolor[RGB]{217,237,215}\textbf{{\boldmath$.12$}} &  & \cellcolor[RGB]{105,183,109}\textbf{{\boldmath$.60$}} \\
Starsim interv. & \cellcolor[RGB]{76,162,88}\textbf{{\boldmath$.74$}} & \cellcolor[RGB]{103,181,107}\textbf{{\boldmath$.61$}} & \cellcolor[RGB]{111,186,114}\textbf{{\boldmath$.58$}} & \cellcolor[RGB]{82,167,92}\textbf{{\boldmath$.71$}} & \cellcolor[RGB]{189,227,185}\textbf{{\boldmath$.24$}} & \cellcolor[RGB]{247,247,247}\textbf{{\boldmath$.00$}} & \cellcolor[RGB]{121,192,123}\textbf{{\boldmath$.53$}} & \cellcolor[RGB]{150,210,147}\textbf{{\boldmath$.40$}} & \cellcolor[RGB]{105,183,109}\textbf{{\boldmath$.60$}} &  \\
\bottomrule
\end{tabular}

%% file: data/simulator_correlation_table.tex
\setlength{\tabcolsep}{6pt}
\renewcommand{\arraystretch}{1.15}
\begin{tabular}{@{}l rl rl@{}}
\toprule
 & \multicolumn{2}{c}{Spearman} & \multicolumn{2}{c}{Partial, given ECI} \\
\cmidrule(lr){2-3}\cmidrule(l){4-5}
Pair & $\rho$ & 95\% CI & $\rho$ & 95\% CI \\
\midrule
ECI--Micropolis & \cellcolor[RGB]{51,141,71}\color{white}\textbf{0.87} & [0.70, 0.93] & -- &  \\
ECI--FreeCiv & \cellcolor[RGB]{67,154,81}\textbf{0.79} & [0.60, 0.89] & -- &  \\
ECI--Starsim & \cellcolor[RGB]{66,153,81}\textbf{0.79} & [0.60, 0.91] & -- &  \\
\midrule
Micropolis--FreeCiv & \cellcolor[RGB]{62,150,78}\textbf{0.82} & [0.61, 0.93] & \cellcolor[RGB]{144,206,142}\textbf{0.43} & [\ensuremath{-}0.02, 0.79] \\
Micropolis--Starsim & \cellcolor[RGB]{72,159,85}\textbf{0.76} & [0.52, 0.90] & \cellcolor[RGB]{190,227,185}\textbf{0.24} & [\ensuremath{-}0.22, 0.69] \\
FreeCiv--Starsim & \cellcolor[RGB]{88,172,96}\textbf{0.68} & [0.40, 0.83] & \cellcolor[RGB]{214,236,211}\textbf{0.14} & [\ensuremath{-}0.28, 0.54] \\
\bottomrule
\end{tabular}

%% file: data/generalization_table.tex
\setlength{\tabcolsep}{5pt}
\begin{tabular}{@{}l l r r l@{}}
\toprule
\multicolumn{5}{@{}l}{\emph{(a) Half vs half: $\theta_m$ fitted on two disjoint halves of the worlds}} \\
Worlds split & Worlds per half & Splits & Mean $\rho$ & Min--max \\
\midrule
All three (nine cells) & 7--8 + 4 + 2 & 1{,}000 & 0.94 & 0.89--0.98 \\
Micropolis & 7 vs 8 & 6{,}435 & 0.94 & 0.82--0.99 \\
FreeCiv & 4 vs 4 & 35 & 0.85 & 0.69--0.95 \\
Starsim & 2 vs 2 & 3 & 0.86 & 0.80--0.93 \\
\midrule
\multicolumn{5}{@{}l}{\emph{(b) Leave one simulator out: $\theta_m$ fitted on two simulators vs the third}} \\
Held-out simulator & Fitted on & & $\rho$ & 95\% CI \\
\midrule
Micropolis & FreeCiv + Starsim & & 0.84 & [0.67, 0.92] \\
FreeCiv & Micropolis + Starsim & & 0.79 & [0.58, 0.89] \\
Starsim & Micropolis + FreeCiv & & 0.79 & [0.55, 0.90] \\
\bottomrule
\end{tabular}

%% file: appendix/D_ablations.tex
\section{Ablations and diagnostics}
\label{app:ablations}

This appendix reports prompt ablations, the Micropolis knowledge test and the FreeCiv no-news control. The effort check is summarized in \cref{app:roster}; the historical Starsim hold set is defined in \cref{app:full-starsim}.

\input{appendix/D1_micropolis_batching}

\paragraph{FreeCiv.}
\label{app:ablation-grouping}

\Cref{tab:freeciv-grouping} compares one-question prompts on September 9 with the reported grouped run of September 20. \FCGroupDirection. ECI correlations change by at most \FCGroupRhoChangeMax, but per-model rank correlations are \FCGroupRankCorrLo--\FCGroupRankCorrHi{} and maximum rank movements are \FCGroupMoves{} places respectively for mid-range, tail, mirror and continuous. On 231 overlapping mid-range questions, mean forecasts fall from \FCGroupOverlapSingle{} to \FCGroupOverlapGrouped{} against mean truth \FCGroupOverlapTruth. The dates, prompt context and two requested hosts also differ, so the comparison does not isolate an effect of batching or a mechanism specific to rare events.

A first grouped run on the same day shuffled question sets and used 50 questions for both kinds. Scores were within \FCGroupShuffledDiffMax{} and rank correlations \FCGroupShuffledRankLo--\FCGroupShuffledRankHi, but o4-mini declined four of eight continuous prompts containing 38--48 questions; the reported run therefore caps continuous prompts at 20. Binary prompts are capped at 50 rather than 100 to limit long-context demands; these comparisons do not establish an optimal batch size.

\begin{table}[htbp]
\caption{FreeCiv scores under one question per prompt (September 9, 2026) and under the grouped prompts of September 20, mean over the 24 models, with the sign-adjusted Spearman correlation with ECI and the rank correlation of the per-model scores between the two runs. Lower scores are better.}
\label{tab:freeciv-grouping}
\centering
\small
\setlength{\tabcolsep}{5pt}
\input{data/appendix_tables/freeciv_grouping}
\end{table}

\input{appendix/D3_micropolis_prompt_variants}
\input{appendix/D4_micropolis_knowledge}

\subsection{Answering again with nothing revealed (FreeCiv)}
\label{app:ablation-nonews}

We score natural conditionals by turn-2 excess Brier loss against $p(Y\mid X)$, and define gain relative to the unchanged turn-1 forecast. The no-news control measures forecast changes when the question is asked again without a revealed fact. For 100 of the 400 cells, stratified over the five structural groups of \cref{sec:freeciv}, we continued the turn-1 exchange with a statement that the game had continued and that no new information was available. These 100 selected cells correspond to 99 distinct question prompts: two cells share one no-news answer. The analysis continues to weight cells, rather than distinct prompts. We then asked the question again. We call the change in the forecast the \emph{drift}. \Cref{tab:nonews} lists it by model. The mean absolute drift over the 24 models is 0.027, from below 0.001 (GPT-5.5) to 0.081 (Kimi K2). Six models drift by less than 0.005 on average. GPT-5.5 and Sonnet 5 return exactly their turn-1 forecast on 96\% and 94\% of cells; GPT-5 Nano, Gemini 3 Flash Preview and Kimi K2 return it on 2\%. The drift is mostly downward: its signed mean is $-0.010$ over models, $-0.064$ for Kimi K2 and $-0.038$ for Haiku 4.5. On the same 100 cells the mean absolute move after a revealed fact is 0.187, about 7 times the drift, so the moves in \cref{fig:freeciv-natcond} are not produced by asking again alone. Drift falls with capability: the Spearman correlation between ECI and mean absolute drift is $-0.75$ [$-0.89$, $-0.48$], $p<0.001$, $n=24$, with a percentile bootstrap over models (10,000 samples). The mean observed accuracy change was small. The excess Brier score of the no-news answer against $p(Y)$ differs from that of the turn-1 answer by less than 0.001 on average over models, and by at most $+0.014$ (Gemini 3 Flash Preview). The control was asked one question per prompt, as the natural-conditional arm was; the rows of DeepSeek V3 and DeepSeek V4 Flash come from their rerun of September 20 (\cref{app:worlds-freeciv}).

\begin{table}[t]
\caption{The no-news control on 100 FreeCiv natural-conditional cells. Drift is the turn-2 forecast under the no-news statement minus the turn-1 forecast; the table gives its mean absolute value and its signed mean. ``Unchanged'' is the share of cells on which the two forecasts are equal, and ``$>0.05$'' the share on which they differ by more than 0.05. ``$|$move$|$'' is the mean absolute move on the same cells when the fact is revealed. Models are ordered by ECI. Llama 4 Scout has 99 cells; every other model has 100. Recomputed from the released per-question scores.}
\label{tab:nonews}
\centering
\small
\setlength{\tabcolsep}{4pt}
\begin{tabular}{@{}lrrrrrr@{}}
\toprule
Model & ECI & $|$drift$|$ & drift & Unchanged & $>0.05$ & $|$move$|$ \\
\midrule
Fable & 162.5 & 0.003 & $-0.003$ & 80\% & 0\% & 0.164 \\
Opus 5 & 161.5 & 0.018 & $+0.003$ & 39\% & 3\% & 0.187 \\
GPT-5.6 Sol & 161.1 & 0.005 & $-0.002$ & 78\% & 1\% & 0.141 \\
GPT-5.5 & 158.6 & 0.001 & $-0.001$ & 96\% & 0\% & 0.167 \\
Gemini 3.7 Flash & 156.8 & 0.003 & $+0.000$ & 77\% & 0\% & 0.170 \\
GPT-5.6 Luna & 156.2 & 0.015 & $-0.005$ & 45\% & 3\% & 0.139 \\
Sonnet 5 & 155.9 & 0.001 & $-0.001$ & 94\% & 0\% & 0.157 \\
DeepSeek V4 Flash & 154.4 & 0.019 & $-0.016$ & 67\% & 9\% & 0.203 \\
Gemini 3 Flash Preview & 150.7 & 0.034 & $+0.013$ & 2\% & 13\% & 0.214 \\
GPT-5 & 150.0 & 0.004 & $-0.001$ & 77\% & 0\% & 0.123 \\
o3 & 147.1 & 0.036 & $-0.002$ & 30\% & 16\% & 0.159 \\
GPT-5.4 Nano & 146.5 & 0.026 & $-0.014$ & 26\% & 12\% & 0.202 \\
o4-mini & 146.5 & 0.019 & $-0.017$ & 55\% & 4\% & 0.164 \\
GPT-5 Mini & 145.8 & 0.010 & $-0.009$ & 80\% & 6\% & 0.197 \\
Gemini 3.1 Flash Lite & 144.7 & 0.031 & $+0.001$ & 43\% & 13\% & 0.225 \\
Qwen3.5 Flash & 144.1 & 0.027 & $+0.004$ & 51\% & 8\% & 0.209 \\
Haiku 4.5 & 142.9 & 0.053 & $-0.038$ & 4\% & 36\% & 0.193 \\
GPT-5 Nano & 140.8 & 0.046 & $-0.025$ & 2\% & 27\% & 0.155 \\
Kimi K2 & 140.4 & 0.081 & $-0.064$ & 2\% & 48\% & 0.215 \\
Gemini 2.5 Flash & 140.3 & 0.040 & $-0.030$ & 58\% & 13\% & 0.252 \\
Qwen3 235B & 139.6 & 0.027 & $-0.017$ & 55\% & 8\% & 0.175 \\
GPT-4.1 & 137.5 & 0.015 & $-0.009$ & 46\% & 4\% & 0.212 \\
DeepSeek V3 & 133.0 & 0.053 & $-0.029$ & 12\% & 19\% & 0.225 \\
Llama 4 Scout & 130.5 & 0.074 & $+0.012$ & 18\% & 42\% & 0.239 \\
\bottomrule
\end{tabular}
\end{table}

\subsection{Starsim reasoning and aggregation robustness}
\label{app:starsim-robustness}
\textbf{Historical target only.} The following checks used the original matched-seed continuous truths and have not been rescored against fixed-state continuations; they do not establish robustness of the revised results. The same items elicited on September 8 at vendor-default reasoning, before the lowest-effort setting of \cref{sec:starsim} was adopted, give, under the same scoring rule, a pooled interventional $\rho(\mathrm{ECI},-\mathrm{loss})$ of $0.66$ with the same five models at the top and a mean excess of 0.47 rather than 0.65. Resampling one answer per item in place of the median gives correlations of $0.72$ [$0.64$, $0.80$] for the September 8 run and $0.73$ [$0.62$, $0.82$] for the September 16 run. These intervals resample answers, not models. The runs are separate elicitations, so their contrast does not isolate a causal effect of reasoning effort. The largest observed difference is at 50 percent coverage, where the mid-sized reasoning models (GPT-5 Mini, o4-mini) give more variable forecasts without a thinking budget. At this coverage, $\rho$ moves from $0.70$ to $0.46$.

\input{appendix/D5_real_world_forecasting}

%% file: appendix/D1_micropolis_batching.tex
\subsection{Questions per prompt}
\label{app:ablation-batching}

\paragraph{Micropolis.}
Eight models (Llama 4 Scout, DeepSeek V3, GPT-4.1, GPT-5 Mini, Gemini 3 Flash Preview, Haiku 4.5, Gemini 3.7 Flash and Fable) answered binary questions for five cities (bruce, haight, freds, happisle and joffburg), with 1, 2, 4, 7--8, 14--15, 25, 50 or 100 questions per prompt. By 7--8 (resp 14--15) we mean some prompts in those groups had 7 (resp 14) question while others had 8 (resp 15) --- this was necessary because the 100 questions could not be evenly divided at this size. \Cref{tab:micropolis-batching} reports the scores for each batch size, parse rates, costs and ECI associations; 50 is the main-run setting.

Tail loss falls from \MPDBatchTailOne\ excess bits at one question per prompt to \MPDBatchTailTwo\ at two and \MPDBatchTailFour\ at four; the paired one-to-four change is \MPDBatchDeltaTailFour\ \MPDBatchDeltaTailFourCIModels. Mid-range changes are smaller and their paired-model intervals mostly include zero; the largest is \MPDBatchDeltaMidEight\ \MPDBatchDeltaMidEightCIModels\ at \MPDBatchSizeEight\ questions. The tail loss with four questions minus that with 50 is $+0.038$ bits, paired-model 95\% interval $[0.010,0.070]$, versus mid-range $+0.00049$ [$-0.0058,0.0070$]. Costs fall approximately as $1/n$. The macro-reported comparisons use \MPDResamples\ paired model resamples; the additional four-versus-50 comparisons use 3,000, seed 20260923. Each draw is shared across settings. These exploratory comparisons do not establish equivalence across batch sizes. Positive ECI associations do not imply stable rankings: rank correlations between settings reach 0.43, with movements up to four places.

The answer-format examples are 0.65 and 0.03 (\cref{app:prompts-micropolis}). \MPDBatchAnchorModel\ answers exactly 0.65 on \MPDBatchAnchorShareSmallest\% of one-question prompts and \MPDBatchAnchorShareSecond\% of two-question prompts; \MPDBatchCopierModel\ also frequently reproduces the example block. These are finished answers, not parsing errors. Excluding 0.65 changes the pooled one-question tail score from \MPDBatchTailOne\ to \MPDBatchTailSmallestNoExample\ bits; a batching difference remains. Excluding both examples gives one-/two-question means 0.215/0.151, versus 0.208/0.139 after excluding only 0.65 and 0.241/0.154 without exclusions. Deletion changes the sample, can remove useful low forecasts, and does not identify copying or a causal mechanism. Tail forecast-minus-truth bias falls from \MPDBatchTailBiasOne\ at one question to \MPDBatchTailBiasFour\ at four and \MPDBatchTailBiasSixtyFour\ at 50, while forecast--truth correlations range from \MPDBatchDiscrimMin\ to \MPDBatchDiscrimMax. 

\begin{table}[htbp]
\caption{The questions-per-prompt ablation by setting: the number of questions per prompt and the number of prompts needed per model; the pooled mid-range excess Brier score and tail excess bits (means of per-model means); the lowest parse rate over the \MPDBatchNModels\ models; the total cost in dollars over all \MPDBatchNModels\ models; and the sign-adjusted Spearman correlation of each score with ECI over the \MPDBatchNModels\ models. $\dagger$: the paper's own run, whose prompts this setting reproduced. 
}
\label{tab:micropolis-batching}
\centering
\small
\input{data/appendix_tables/micropolis_batching}
\end{table}

%% file: data/appendix_tables/micropolis_batching.tex
\setlength{\tabcolsep}{3.5pt}
\begin{tabular}{rrrrrrrr}
\toprule
\multicolumn{2}{c}{Per model} & \multicolumn{2}{c}{Pooled score} & Parsed & Total cost & \multicolumn{2}{c}{$\rho$ with ECI} \\
\cmidrule(lr){1-2}\cmidrule(lr){3-4}\cmidrule(lr){7-8}
Q/prompt & Prompts & Mid-range & Tail & (\%) & (\$) & Mid-range & Tail \\
\midrule
1 & 1,000 & 0.0776 & 0.241 & 97.9 & 52.53 & 0.60 & 0.60 \\
2 & 500 & 0.0758 & 0.154 & 96.4 & 28.70 & 0.67 & 0.69 \\
4 & 250 & 0.0675 & 0.137 & 90.4 & 16.26 & 0.86 & 0.52 \\
7--8 & 130 & 0.0623 & 0.138 & 86.2 & 10.02 & 0.95 & 0.74 \\
14--15 & 70 & 0.0648 & 0.131 & 94.2 & 6.70 & 0.86 & 0.67 \\
25 & 40 & 0.0654 & 0.095 & 100.0 & 4.75 & 0.79 & 0.90 \\
50$^\dagger$ & 20 & 0.0670 & 0.100 & 100.0 & 2.83 & 0.81 & 0.60 \\
100 & 10 & 0.0667 & 0.134 & 100.0 & 2.17 & 0.81 & 0.88 \\
\bottomrule
\end{tabular}

%% file: data/appendix_tables/freeciv_grouping.tex
\begin{tabular}{@{}lrrrrr@{}}
\toprule
 & \multicolumn{2}{c}{One per prompt} & \multicolumn{2}{c}{Grouped} & \\
\cmidrule(lr){2-3}\cmidrule(lr){4-5}
Set & Mean & $\rho$ & Mean & $\rho$ & Rank corr. \\
\midrule
Mid-range, excess Brier & 0.129 & 0.12 & 0.141 & 0.18 & 0.72 \\
Tail, excess bits & 0.556 & 0.35 & 0.332 & 0.34 & 0.68 \\
Mirror, excess Brier & 0.294 & 0.08 & 0.406 & 0.10 & 0.85 \\
Continuous, excess nCRPS & 0.524 & 0.61 & 0.648 & 0.59 & 0.70 \\
\bottomrule
\end{tabular}

%% file: appendix/D3_micropolis_prompt_variants.tex
\subsection{World report and prompt variants (Micropolis)}
\label{app:ablation-variants}

We compared \MPDVarNVariants\ continuous-prompt configurations using \MPDVarNModels\ models at the main reasoning settings and \MPDVarNCities\ cities (\MPDVarNQuestions\ questions, \MPDVarNPrompts\ prompts per model). \Cref{tab:prompt-variants} specifies the history, snapshot ages, preamble, labels, order and batch size; \MPDVarMain\ is the main run. Scores use the common normalization described in \cref{app:worlds-micropolis}. Some contrasts change several features.

The variants change five things. The \emph{history table} of the report is sampled every 4 turns (\texttt{long}), every 48 turns, one row per game year (\texttt{yfreq}), cut to its last two yearly rows (\texttt{last-2y}), or removed, so that the model sees the current state and the events alone (\texttt{snapshot-only}). The \emph{preamble} is either preamble 1, which assigns the forecaster's role and asks for calibrated percentiles, or preamble 2b, which adds that service funding is fixed at 100\% and never decays, that CRPS scores the answer so that intervals too narrow and too wide both lose, and that the five percentiles must be non-decreasing (preamble2b, shown in \cref{app:prompts-micropolis}). A question is \emph{labeled} by a number, the questions ordered by turn, or by its metric and turn, the questions grouped by metric (\texttt{semantic}). One variant puts one metric's four horizons in a \emph{prompt of its own}, five prompts per snapshot instead of one (\texttt{smallbatch}). And the \emph{snapshot} moves from game years 5 and 10 to years 20 and 30 (\texttt{20yhist}), 30 and 40 (\texttt{semlonger}, \texttt{numlonger}) or 100 and 118 (\texttt{100yhist}), with the history sampled every 24 turns to keep the history table to a manageable length. The four variants built on preamble 1, and \texttt{preamble2} itself, also print the report's three funding-effectiveness lines, which the later variants drop once the preamble states that funding is fixed.

\begin{table}[htbp]
\caption{The \MPDVarNVariants\ prompt variants of the Micropolis continuous set.
``vs'' is the comparison configuration, which may differ in several features. ``Snapshot'' is in game years and ``History'' the sampling interval of the report's history table, with $\ast$ where the report also prints the road, police and fire funding-effectiveness lines. Preamble 1 assigns the role and asks for calibrated percentiles; preamble 2b adds that funding is fixed at 100\% and that CRPS scores the answer. ``Label, order'' says how a question is named and whether the questions are ordered by turn or grouped by metric; ``Per prompt'' is the number of questions one prompt carries. $\dagger$: the main run's prompts. Every variant runs \MPDVarNCities\ cities with disasters on, \MPDVarNModels\ models and horizons of 3, 5, 7 and 10 game years. The \texttt{100yhist} snapshots are turns 4,800 and 5,670.
}
\label{tab:prompt-variants}
\centering
\scriptsize
\input{data/appendix_tables/micropolis_variants_settings}
\end{table}

\begin{table}[htbp]
\caption{Results for each prompt variant: the reference configuration it is compared against (\cref{tab:prompt-variants} gives the settings and snapshots); the pooled excess nCRPS (mean of per-model means); its difference from that reference, paired across the \MPDVarNModels\ models, with the 95\% interval of that paired bootstrap; the share (\%) of the variant's forecasts that did not parse, over all models; the sign-adjusted Spearman correlation of the per-model means with ECI, with a 95\% bootstrap interval over models; and the cost of the run in dollars over all models. $\dagger$: the main run, using its cached answers.
}
\label{tab:micropolis-variants}
\centering
\scriptsize
\input{data/appendix_tables/micropolis_variants}
\end{table}

Among same-age configurations, removing history produces the largest loss increase; two yearly rows do not recover it. Yearly versus 4-turn history sampling is unresolved. Preamble 2b improves on preamble 1, but bundles several instructions. Semantic labels increase parsing failures, and small batches triple costs without a resolved score improvement (\cref{tab:micropolis-variants}). These contrasts compare whole configurations; the shared-answer checks below address availability, not causal identification.

Comparisons across snapshot ages change the prediction targets as well as the prompts: losses fall from \MPDVarSemanticMean\ at years 5/10 to \MPDVarSemlongerMean\ at years 30/40 and \MPDVarHundredYHistMean\ at years 100/118. All variants use the main run's city-and-metric scales through year 20 (\cref{app:worlds-micropolis}); lower normalized losses do not isolate a dynamical explanation or prompt effect.

The correlation with ECI is present in every variant, between \MPDVarRhoMin\ (\MPDVarRhoMinVariant) and \MPDVarRhoMax\ (\MPDVarRhoMaxVariant), and no interval reaches zero (\cref{tab:micropolis-variants}). It is weakest where the report says least (\texttt{snapshot-only}, \texttt{last-2y}) and where the questions are easiest (\texttt{100yhist}), and strongest in the 4-turn-history variants at years 5 and 10. Positive ECI associations do not imply stable rankings: pairwise variant rank correlations reach 0.607, with a maximum 13-place shift. Comparisons of semantic/preamble2 and smallbatch/preamble2 at the same snapshot ages, restricted to answers valid in both configurations, give 0.870 and 0.842.

The post-hoc median-range rule in \cref{app:worlds-micropolis} was motivated by three pathological responses among 10,800 calls (158,400 question-answer opportunities), including incorrect traffic units and population values repeated as answers for other metrics. Before the rule, those replies moved one variant's mean from 0.065 to 216. The rule removes 53 answer entries within 1,353 total invalid opportunities, leaving 157,047 valid answers; it removes none from the main run. These are filtering choices, not evidence that missing answers are random. The selected main configuration uses preamble 2b, numbered questions grouped by metric, one prompt per snapshot, 24-turn history sampling and years 20/30.

%% file: data/appendix_tables/micropolis_variants_settings.tex
\setlength{\tabcolsep}{2.5pt}
\begin{tabular}{@{}llllllc@{}}
\toprule
Variant & vs & Snapshot & History & Preamble & Label, order & Per prompt \\
\midrule
\texttt{snapshot-only} & \texttt{long} & 5, 10 & none$^\ast$ & 1 (snapshot) & number, turn & 20 \\
\texttt{last-2y} & \texttt{yfreq} & 5, 10 & 48 turns, last 2 rows$^\ast$ & 1 & number, turn & 20 \\
\texttt{yfreq} & \texttt{long} & 5, 10 & 48 turns$^\ast$ & 1 & number, turn & 20 \\
\texttt{long} & -- & 5, 10 & 4 turns$^\ast$ & 1 & number, turn & 20 \\
\texttt{preamble2} & \texttt{long} & 5, 10 & 4 turns$^\ast$ & 2b & number, turn & 20 \\
\texttt{semantic} & \texttt{preamble2} & 5, 10 & 4 turns & 2b & metric@turn, metric & 20 \\
\texttt{smallbatch} & \texttt{preamble2} & 5, 10 & 4 turns & 2b & number, metric & 4 \\
\texttt{20yhist}$^\dagger$ & \texttt{numlonger} & 20, 30 & 24 turns & 2b & number, metric & 20 \\
\texttt{semlonger} & \texttt{semantic} & 30, 40 & 24 turns & 2b & metric@turn, metric & 20 \\
\texttt{numlonger} & \texttt{semlonger} & 30, 40 & 24 turns & 2b & number, metric & 20 \\
\texttt{100yhist} & \texttt{numlonger} & 100, 118 & 24 turns & 2b & number, metric & 20 \\
\bottomrule
\end{tabular}

%% file: data/appendix_tables/micropolis_variants.tex
\setlength{\tabcolsep}{2pt}
\begin{tabular}{llrrrrr}
\toprule
Variant & vs & Excess nCRPS & $\Delta$ vs parent & Unparsed & $\rho$ with ECI & Cost (\$) \\
\midrule
\texttt{snapshot-only} & \texttt{long} & 0.115 & $+0.051$ $[0.030,\,0.074]$ & 0.4 & 0.68 $[0.36,\,0.85]$ & 7.30 \\
\texttt{last-2y} & \texttt{yfreq} & 0.119 & $+0.054$ $[0.040,\,0.070]$ & 0.8 & 0.66 $[0.32,\,0.83]$ & 7.36 \\
\texttt{yfreq} & \texttt{long} & 0.065 & $0.000$ $[-0.003,\,0.004]$ & 0.1 & 0.76 $[0.47,\,0.90]$ & 7.56 \\
\texttt{long} & -- & 0.065 & -- & 1.1 & 0.83 $[0.61,\,0.92]$ & 8.89 \\
\texttt{preamble2} & \texttt{long} & 0.055 & $-0.009$ $[-0.015,\,-0.004]$ & 0.4 & 0.86 $[0.63,\,0.96]$ & 9.29 \\
\texttt{semantic} & \texttt{preamble2} & 0.054 & $-0.002$ $[-0.005,\,0.002]$ & 1.8 & 0.78 $[0.51,\,0.90]$ & 9.15 \\
\texttt{smallbatch} & \texttt{preamble2} & 0.058 & $+0.003$ $[-0.001,\,0.007]$ & 0.3 & 0.77 $[0.48,\,0.91]$ & 28.25 \\
\texttt{20yhist}$^\dagger$ & \texttt{numlonger} & 0.039 & $+0.005$ $[0.003,\,0.006]$ & 0.6 & 0.82 $[0.56,\,0.92]$ & 8.47 \\
\texttt{semlonger} & \texttt{semantic} & 0.034 & $-0.020$ $[-0.024,\,-0.016]$ & 2.7 & 0.79 $[0.51,\,0.91]$ & 9.36 \\
\texttt{numlonger} & \texttt{semlonger} & 0.034 & $0.000$ $[-0.003,\,0.003]$ & 0.6 & 0.80 $[0.57,\,0.91]$ & 8.81 \\
\texttt{100yhist} & \texttt{numlonger} & 0.020 & $-0.014$ $[-0.021,\,-0.002]$ & 0.6 & 0.64 $[0.26,\,0.85]$ & 11.34 \\
\bottomrule
\end{tabular}

%% file: appendix/D4_micropolis_knowledge.tex
\subsection{What models know about the Micropolis engine}
\label{app:ablation-knowledge}

We test whether a score that reflects both knowledge and abstention is associated with Micropolis forecasting. To probe how much the models know about the engine, we wrote a test of \MPDKnowNStatements\ statements about MicropolisCore, each to be answered True, False or Unknown: \MPDKnowNTrue\ are true and \MPDKnowNFalse\ false. Most come in \MPDKnowNPairs\ pairs, a true statement and a false twin that changes one thing about it, a number, a direction or a mechanism, so that the false statements are as specific as the true ones. A further \MPDKnowNHoneypots\ statements describe mechanics of later SimCity titles that Micropolis does not have, such as water pipes, subways or city ordinances (we refer to these as \emph{honeypot} questions), and the share a model marks True measures how much of a neighboring game it brings along. Each statement carries a difficulty: level 0 (\MPDKnowNDifficultyZero\ statements) for facts anyone who has seen the game would know, level 1 (\MPDKnowNDifficultyOne) for facts from playing it or reading its manual, and level 2 (\MPDKnowNDifficultyTwo) for facts known only from the source code. Each also carries a topic. \emph{Engine} (\MPDKnowNEngine) covers disasters, zoning, power, the budget and the internals; \emph{cities} (\MPDKnowNCities) covers which sample cities ship with the game and what some of them are like when loaded; \emph{dynamics} (\MPDKnowNDynamics) covers what moves the metrics the world report shows: that the four averages are means over byte-valued maps, that the population is a fixed formula over zone densities and can grow only by densification once nobody zones, how often the census runs, that unpowered zones lose people, that rubble stays and traffic decays. The dynamics statements are the knowledge that could help a model forecast the report. Every statement was checked against the fork's source or the bundled cities' own files.

Each model receives two prompts of \MPDKnowPerPrompt\ statements, built so that the two members of a pair never share a prompt: shown ``one in eight'' beside ``one in two'', a model would know that exactly one is true. A prompt names the engine and its version, lists its statements in one numbered list and asks for one answer per line with no explanation; the exact preamble and both statement lists are \changednew{in the data release} (\cref{app:release}).
It states the scoring rule, $+1$ for a correct answer, $-2$ for a wrong one and 0 for Unknown, so that answering pays only above 67\% confidence. An answer that does not parse counts as wrong. A model's score on a set of statements is its points divided by their number: 1 for all correct, 0 for all Unknown and $-2$ for all wrong, on every subset alike. The test was run on September 22, 2026, on the \MPDKnowNModels\ models of the main run at the main run's reasoning settings, one call per prompt, for \$\MPDKnowCost\ in all.

\begin{figure}[htbp]
\centering
\includegraphics[width=\linewidth]{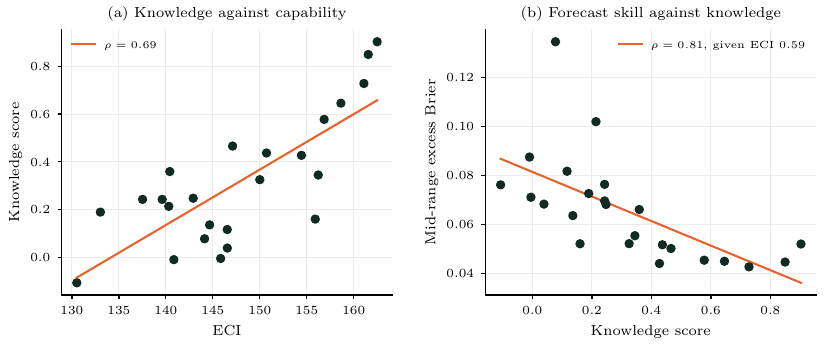}
\caption{The knowledge test against capability and against forecast skill, one point per model (\MPDKnowNModels). (a) Knowledge score over all \MPDKnowNStatements\ statements against ECI, Spearman $\rho$ over the models. (b) The main run's mid-range excess Brier score against the knowledge score; $\rho$ is sign-adjusted so that a positive value means more knowledge, better forecasts, and ``given ECI'' is the partial rank correlation with ECI removed from both. Lines are least-squares fits.}
\label{fig:micropolis-knowledge}
\end{figure}

\Cref{tab:micropolis-knowledge,tab:micropolis-knowledge-corr} report model scores and associations. Engine, dynamics and honeypot scores track ECI; city knowledge does not. Most models score below zero on the city subset, answering rather than abstaining despite frequent errors. Separate intervals do not establish equivalence among difficulty tiers.

The knowledge--forecasting association persists after adjusting for ECI on mid-range questions, but not on tail questions (\cref{tab:micropolis-knowledge-corr}). Continuous results depend on the inferential procedure, as detailed below. These associations do not identify a causal effect of engine knowledge or establish that tail errors reflect calibration rather than understanding.

Two response patterns help explain the scores. First, \MPDKnowNAbstainers\ models answered Unknown to more than half the statements (\MPDKnowAbstainers), \MPDKnowUnknownMaxModel\ to \MPDKnowUnknownMax\% of them. The rule scores an abstention 0, so their scores near zero measure a refusal to guess as much as a lack of knowledge; GPT-5 Nano behaves the same way in the forecasting runs. Second, the honeypots: on average a model marked \MPDKnowHoneypotTrueMean\% of them True, \MPDKnowNHoneypotClean\ models marked none, and \MPDKnowHoneypotTrueMaxModel\ marked \MPDKnowHoneypotTrueMax\%. The mechanics of SimCity 2000 and 3000 are part of what the smaller models believe about Micropolis.

The test has three limitations. The statements were drafted by \MPDKnowAuthor\ with references into the source code, and that model, which is in the panel, scores highest (\MPDKnowAuthorScore, rank \MPDKnowAuthorRank); without it the correlation with ECI is \MPDKnowRhoAllNoAuthor. The source checks do not eliminate wording ambiguity or cross-cues, and a test written by one model may favor what that model finds natural to state. The difficulty levels are subjective. And the test is one exchange per prompt with no repeats, so a model's score carries the noise of a single reply. An earlier test was replaced before these results were used; its details remain in the preserved source.

\paragraph{Abstention, cues and inferential sensitivity.}
The $+1/-2/0$ score combines knowledge, abstention and response strategy. Partial rank associations given ECI are 0.591, 0.120 and 0.488 for mid-range, tail and continuous skill; additionally adjusting for abstention fraction gives 0.597, 0.134 and 0.517. These are exploratory associations, not a causal adjustment or a pure knowledge measure. The continuous partial correlation's model-percentile interval includes zero ($[-0.04,0.78]$), whereas the reported $p=0.018$ comes from a $t$ approximation with $n-3$ degrees of freedom; the two methods do not agree on statistical significance.

Putting each true/false pair in separate prompts does not remove every clue from related statements within a prompt. Ten identified cross-cue items yield partial associations 0.565, 0.139 and 0.473 when deleted from the existing answers. Census wording is also ambiguous: the engine updates the short-term census every four city-time units and the long-term census every 48, while prompt1 item13 and prompt2 item86 do not specify which series. Their historical labels and answers are preserved. Excluding the two ambiguous items gives 0.588, 0.133 and 0.487; excluding them plus the cross-cues (11 unique items) gives 0.564, 0.155 and 0.476. Deleting these items provides limited sensitivity checks, not a complete cue audit or a causal experiment, since models already saw the full prompts.

\begin{table}[htbp]
\caption{The knowledge test by model, ordered by ECI. Scores follow the prompt's rule, $+1$, $-2$ and 0 over the statements of the subset, so 1 is every statement right and 0 is every statement Unknown: over all \MPDKnowNStatements\ statements, over the engine, cities and dynamics topics, and over the \MPDKnowNHoneypots\ honeypots. ``Unknown'' is the share of all statements the model declined to answer and ``Wrong'' counts its wrong and unparsed answers.
}
\label{tab:micropolis-knowledge}
\centering
\footnotesize
\input{data/appendix_tables/micropolis_knowledge}
\end{table}

\begin{table}[htbp]
\caption{Correlations of the knowledge score by statement subset over the \MPDKnowNModels\ models: Spearman $\rho$ with ECI, with a 95\% bootstrap interval over the models; with each of the main run's three forecast scores, sign-adjusted so that a positive value means more knowledge, better forecasts; and the same three with the ECI rank partialled out of both sides.
}
\label{tab:micropolis-knowledge-corr}
\centering
\footnotesize
\input{data/appendix_tables/micropolis_knowledge_corr}
\end{table}

\paragraph{Administered-item examples and preservation.}
Original prompt 1 includes ``A separate water-coverage value is tracked per map location'' and ``Each pass of the simulator is divided into 16 phases.'' The first concerns a mechanic absent from the engine; the second concerns a fact from the source code; \changednew{the complete administered lists and preamble are in the data release, and the historical ambiguous census wording is retained separately and available on request. These records are required} for exact reproduction; the examples above do not replace them.

%% file: data/appendix_tables/micropolis_knowledge.tex
\setlength{\tabcolsep}{3pt}
\begin{tabular}{lrrrrrrrr}
\toprule
Model & ECI & All & Engine & Cities & Dynamics & Honeypot & Unknown (\%) & Wrong \\
\midrule
Claude Fable & 162.5 & 0.90 & 0.96 & 0.54 & 0.96 & 1.00 & 8 & 1 \\
Claude Opus 5 & 161.5 & 0.85 & 0.86 & 0.71 & 0.96 & 1.00 & 5 & 7 \\
GPT-5.6 Sol & 161.1 & 0.73 & 0.93 & -0.43 & 0.77 & 1.00 & 2 & 17 \\
GPT-5.5 & 158.6 & 0.65 & 0.87 & -0.36 & 0.42 & 1.00 & 9 & 18 \\
Gemini 3.7 Flash & 156.8 & 0.58 & 0.78 & -0.50 & 0.54 & 1.00 & 0 & 29 \\
GPT-5.6 Luna & 156.2 & 0.34 & 0.47 & -0.29 & 0.31 & 0.25 & 0 & 45 \\
Claude Sonnet 5 & 155.9 & 0.16 & 0.24 & -0.18 & 0.04 & 0.25 & 11 & 50 \\
DeepSeek V4 Flash 0731 & 154.4 & 0.43 & 0.52 & -0.18 & 0.54 & 0.40 & 12 & 31 \\
Gemini 3 Flash Preview & 150.7 & 0.44 & 0.56 & -0.18 & 0.38 & 1.00 & 13 & 30 \\
GPT-5 & 150.0 & 0.33 & 0.48 & -0.50 & 0.31 & 0.55 & 0 & 46 \\
o3 & 147.1 & 0.47 & 0.57 & 0.00 & 0.35 & 0.70 & 29 & 17 \\
GPT-5.4 Nano & 146.5 & 0.04 & 0.09 & -0.14 & -0.04 & -0.85 & 51 & 31 \\
o4 Mini & 146.5 & 0.12 & 0.22 & -0.14 & -0.19 & 0.25 & 10 & 54 \\
GPT-5 Mini & 145.8 & -0.00 & 0.07 & -0.50 & 0.08 & -0.80 & 1 & 68 \\
Gemini 3.1 Flash Lite Preview & 144.7 & 0.14 & 0.25 & -0.43 & 0.08 & 0.25 & 0 & 59 \\
Qwen3.5-Flash & 144.1 & 0.08 & 0.12 & -0.18 & 0.08 & 0.70 & 0 & 63 \\
Claude Haiku 4.5 & 142.9 & 0.25 & 0.30 & 0.00 & 0.19 & 0.60 & 56 & 13 \\
GPT-5 Nano & 140.8 & -0.01 & -0.01 & 0.00 & 0.00 & -0.10 & 98 & 2 \\
Kimi K2 & 140.4 & 0.36 & 0.47 & -0.14 & 0.23 & 0.55 & 17 & 32 \\
Gemini 2.5 Flash & 140.3 & 0.21 & 0.31 & -0.29 & 0.19 & 0.85 & 0 & 54 \\
Qwen3 235B A22B & 139.6 & 0.24 & 0.30 & 0.00 & 0.19 & 0.25 & 36 & 27 \\
GPT-4.1 & 137.5 & 0.24 & 0.41 & -0.71 & 0.31 & 1.00 & 0 & 52 \\
DeepSeek V3 & 133.0 & 0.19 & 0.14 & 0.21 & 0.42 & 0.25 & 11 & 48 \\
Llama 4 Scout & 130.5 & -0.11 & -0.03 & -0.71 & 0.08 & -0.35 & 0 & 76 \\
\bottomrule
\end{tabular}

%% file: data/appendix_tables/micropolis_knowledge_corr.tex
\setlength{\tabcolsep}{3pt}
\begin{tabular}{lrrrrrrrr}
\toprule
Subset & $n$ & $\rho$ with ECI & \multicolumn{3}{c}{$\rho$ with forecast score} & \multicolumn{3}{c}{$\rho$ with forecast score, given ECI} \\
\cmidrule(lr){4-6}\cmidrule(lr){7-9}
 & & & Mid-range & Tail & Continuous & Mid-range & Tail & Continuous \\
\midrule
All & 24 & 0.69 $[0.34,\,0.90]$ & 0.81 & 0.58 & 0.77 & 0.59 & 0.12 & 0.49 \\
Engine & 24 & 0.71 $[0.38,\,0.89]$ & 0.80 & 0.63 & 0.79 & 0.55 & 0.22 & 0.52 \\
Cities & 24 & 0.07 $[-0.48,\,0.54]$ & 0.03 & -0.30 & -0.01 & -0.05 & -0.53 & -0.11 \\
Dynamics & 24 & 0.53 $[0.08,\,0.83]$ & 0.72 & 0.55 & 0.68 & 0.58 & 0.28 & 0.51 \\
Honeypot & 24 & 0.46 $[0.03,\,0.76]$ & 0.47 & 0.41 & 0.56 & 0.20 & 0.11 & 0.37 \\
Difficulty 0 & 24 & 0.56 $[0.18,\,0.83]$ & 0.61 & 0.50 & 0.69 & 0.31 & 0.14 & 0.48 \\
Difficulty 1 & 24 & 0.76 $[0.45,\,0.91]$ & 0.82 & 0.71 & 0.84 & 0.55 & 0.32 & 0.59 \\
Difficulty 2 & 24 & 0.60 $[0.18,\,0.86]$ & 0.71 & 0.45 & 0.55 & 0.48 & 0.01 & 0.13 \\
\bottomrule
\end{tabular}

%% file: appendix/D5_real_world_forecasting.tex
\subsection{Upper-quantile calibration in FreeCiv and historical measles forecasts}
\label{app:real-world-forecasting}

We measure the proportion of outcomes strictly above each model's predicted 75th percentile. For FreeCiv, this is the proportion of the 1,000 replay outcomes for each question; for measles, it is an indicator for the observed historical outcome. We average available questions within each model at each horizon, then give models equal weight. Strict exceedance handles ties in these discrete outcomes: a calibrated 75th percentile has exceedance probability at most 25\%.

The FreeCiv analysis uses all 300 continuous questions from the eight benchmark worlds and all 24 models, with 60 questions at each horizon. The historical measles cohort contains 1,339 state-seasons with season-start years from 1928 to 1962\changednew{, taken from Project Tycho's weekly state-level counts of reported measles cases \citep{tycho2018measles}}. Models are included if they cover at least 80\% of these series, giving 28 models. There are 7,942 available series--horizon targets; 1,247 state-seasons have an outcome at 20 weeks. The main measles prompts contain unlabeled numerical histories. The historical task selects sufficiently large outbreaks and uses the complete season to choose forecast origins before peaks, so this is a selected retrospective forecasting task, not a prospective evaluation.

\Cref{tab:calibration-horizons} reports the model-averaged exceedance rates at every evaluated horizon. The error bars in \cref{fig:calibration-case-study} are pointwise 95\% percentile intervals from 10,000 cluster-bootstrap draws (seed 20261110), resampling the eight FreeCiv worlds or the 35 measles season-start years while holding the model panels fixed. At the longest horizons the intervals are 49.5--60.8\% for FreeCiv and 80.2--84.3\% for measles. These intervals do not jointly account for variation across models, states and years. FreeCiv exceedance is not monotonic in horizon, and family-specific results differ: eight of nine family point estimates exceed 25\% at 150 turns, while world-wonder forecasts have an exceedance rate of 8.2\%.

For the domain-label sensitivity, we pair unlabeled and explicitly labeled measles forecasts on the same series and horizon within each model, retaining the same year window. At 20 weeks, the equal-model mean exceedance is 85.5\% without labels and 91.8\% with labels; all 28 models exceed 25\% in the labeled condition. The matched subset has 84--247 pairs per model at that horizon. This smaller selected subset shows that the error persists when the domain is disclosed; it does not estimate a randomized context effect for the full cohort.

This diagnostic is exploratory: an initial hypothesis about overly high lower quartiles was not supported in aggregate, and the upper-quartile error was identified in the broader quantile checks. The comparison establishes a shared observable error, not a shared internal mechanism or transfer of model rankings or improvements between domains.

\begin{table}[htbp]
  \caption{Mean proportion of outcomes exceeding the predicted 75th percentile, averaging questions within models and then models equally. The nominal upper bound is 25\%.}
  \label{tab:calibration-horizons}
  \centering
  \begin{tabular}{rrrr}
    \toprule
    FreeCiv horizon (turns) & Exceedance & Measles horizon (weeks) & Exceedance \\
    \midrule
    30 & 58.9\% & 2 & 60.4\% \\
    60 & 69.1\% & 4 & 68.6\% \\
    90 & 62.4\% & 8 & 81.0\% \\
    120 & 61.1\% & 12 & 85.0\% \\
    150 & 55.8\% & 16 & 84.7\% \\
    -- & -- & 20 & 82.3\% \\
    \bottomrule
  \end{tabular}
\end{table}

%% file: appendix/E_validation.tex
\section{Correlations with capabilities and forecasting benchmarks}
\label{sec:validation}

\subsection{Correlation with the Epoch Capabilities Index}

If our benchmark scores measure skill, we expect them to correlate with broad capability measures such as ECI. To validate
this expectation, we examine rank correlations between ECI and our score measures. We find every sign-adjusted correlation is positive, with
seven of the nine confidence intervals excluding zero (\cref{tab:validation}). Correlations are stronger for Micropolis (0.75--0.82) and Starsim (0.67--0.75) and weaker for FreeCiv (\FCRhoMin--\FCRhoMax). The two confidence intervals that include zero are FreeCiv mid-range (\FCMidRho) and tail (\FCTailRho). See \Cref{sec:freeciv-results} for some observations on the source of these weaker FreeCiv numbers. The combined score $\theta_m$, a per-model term fitted to the log losses of all nine scores (\cref{app:combined}), correlates with ECI at $\CombinedRho$ [$\CombinedLo$, $\CombinedHi$]. This is a higher correlation than any in \cref{tab:validation}, which is partly expected: combining the
nine scores cancels some of their noise.

The association with ECI is robust to changes in prompt design (\cref{app:ablation-variants,app:ablation-grouping}), forecast horizon (\MPDRhoHorizonMin--\MPDRhoHorizonMax\ for both binary scores at every horizon, computed from \cref{tab:micropolis-horizon}), question selection (\cref{app:micropolis-robustness}), and to how the nine scores are combined (\cref{app:combined}).
Individual model ranks move more under these changes, for instance by up to 13 places across prompt configurations (\cref{app:ablation-variants}). This is likely in part because separation between model scores is often of the order of their confidence intervals over questions, making the ranking unstable. The upshot is that our benchmark separates more capable models from weaker ones, but a single model's rank is uncertain.

\Cref{app:generalization} reports checks of ranking agreement across disjoint subsets of the observed worlds and across simulators.

\begin{table}[t]
  \caption{Spearman correlation $\rho$ between ECI and each score over 24 models (23 for Starsim interventional forecasts), using $\rho(\mathrm{ECI},-\mathrm{loss})$ so that a positive value means more capable models score better; 95\% CI: percentile bootstrap over models; $p$: two-sided.}
  \label{tab:validation}
  \centering
  \small
\input{data/validation_table_main}
\end{table}

\subsection{Correlations with forecasting benchmarks}
\label{app:validation-forecastbench}

\Cref{tab:validation-forecastbench} repeats the nine correlations of \cref{tab:validation} with the ForecastBench overall score \citep{karger2024forecastbench} in place of ECI. Seventeen of the 24 models have a ForecastBench score\footnote{The seven without are Opus 5, Fable, GPT-5.6 Sol, GPT-5.6 Luna, Gemini 3.7 Flash, DeepSeek V4 Flash and Qwen3.5 Flash}, ranging from 54.6 to 61.1.

Over the 17 models ECI and ForecastBench scores correlate at $\rho = 0.62$ [0.18, 0.88], $p = 0.008$. For each of the nine correlations, the point estimate is positive, but the confidence interval includes zero. Intervals are wider than in \cref{tab:validation} because of the smaller number of models used.
Because ECI and ForecastBench are themselves correlated, \cref{tab:validation-forecastbench} also gives the partial Spearman correlation of each score with ForecastBench given ECI: the ranks of ForecastBench and of the score are each regressed on the ranks of ECI, and the residuals are correlated. These partial correlations run from $-0.26$ to $0.13$, with every interval including zero: once ECI is accounted for, we find no evidence that ForecastBench carries further information about our scores.

\begin{table}[htbp]
  \caption{Spearman correlation $\rho$ between the ForecastBench overall score and each of our 9 scores, over the 17 models covered by ForecastBench. The correlation is sign-adjusted so a positive value means higher ForecastBench score corresponds to lower forecasting loss. CI (models): 95\% percentile bootstrap over models, with 2,000 resamples for the Starsim continuous score and \MPDResamples\ resamples for the remainder, at seed \MPDSeed; $p$ is two-sided. The partial columns give the Spearman correlation with ForecastBench with ECI partialled out of both sides.}
  \label{tab:validation-forecastbench}
  \centering
  \small
\input{data/validation_table_forecastbench}
\end{table}

\paragraph{Other forecasting benchmarks.}
\Cref{tab:external-benchmarks} compares $\theta_m$ and ECI with two further forecasting benchmarks. FutureEval is the Metaculus model leaderboard \citep{metaculus2026futureeval} (skill score, snapshot of 4 September 2026) and covers 18 of our models; it lacks Llama 4 Scout, Qwen3 235B, Qwen3.5 Flash, GPT-5.4 Nano, DeepSeek V4 Flash and Gemini 3.7 Flash. Prophet Arena scores Brier on over 1,300 resolved Kalshi events; its Table 6 \citep{yang2026prophetarena} covers 9 of our models (GPT-5, o3, GPT-4.1, o4-mini, Kimi K2, Gemini 2.5 Flash, DeepSeek V3, Llama 4 Scout and Qwen3 235B). Models are matched by name; where a leaderboard lists several reasoning settings of a model, we take the unqualified entry, or the only one listed. Among the three external benchmarks, ForecastBench is the outlier: on overlapping models it correlates \ExtBetweenFBFE{} with FutureEval and \ExtBetweenFBPA{} with Prophet Arena, while those two correlate \ExtBetweenFEPA{} with each other.
Because $\theta_m$ and all three benchmarks correlate with ECI, \cref{tab:external-benchmarks} also gives partial Spearman correlations, computed as in \cref{app:validation-forecastbench}. After adjusting for ECI, the point estimates of the partial correlations of $\theta_m$ are positive for FutureEval at $\ExtFEThetaGivenEci$ $\ExtFEThetaGivenEciCI$ and for Prophet Arena at $\ExtPAThetaGivenEci$ $\ExtPAThetaGivenEciCI$, and negative for ForecastBench ($\ExtFBThetaGivenEci$ $\ExtFBThetaGivenEciCI$); all three intervals include zero. Conversely, given $\theta_m$, ECI correlates at $\ExtFEEciGivenTheta$, $\ExtPAEciGivenTheta$ and $\ExtFBEciGivenTheta$ with the three benchmarks. The point estimates suggest that for FutureEval, $\theta_m$ and ECI each carry information that the other lacks; for ForecastBench, information lies in ECI; with only 9 models, for Prophet Arena we cannot separate them. Most of these intervals include zero, so this is suggestive rather than conclusive. Panel (b) shows that differing model coverage alone does not explain the weaker correlation of ForecastBench: on the \ExtMatchedN{} models both FutureEval and ForecastBench cover, $\theta_m$ correlates $\ExtMatchedFETheta$ with FutureEval and $\ExtMatchedFBTheta$ with ForecastBench, a difference of $\ExtMatchedDiffTheta$ $\ExtMatchedDiffThetaCI$.

\begin{table}[htbp]
  \caption{Spearman correlation of the combined score $\theta_m$ and of ECI with three forecasting benchmarks, (a) over the models each benchmark covers and (b) over the \ExtMatchedN{} models both FutureEval and ForecastBench cover, with the difference between the two rows. The last two columns are partial Spearman correlations, the other variable partialled out of both sides. Ties are kept. Prophet Arena Brier is sign-flipped so that positive means agreement. 95\% percentile bootstrap over models, 10,000 resamples, seed 2026, the same resamples for every column of a panel. Undefined bootstrap correlations are omitted.}
  \label{tab:external-benchmarks}
  \centering\footnotesize
  \setlength{\tabcolsep}{2.8pt}
  \input{data/external_benchmarks_table}
\end{table}

%% file: data/validation_table_main.tex
\setlength{\tabcolsep}{3pt}
\begin{tabular}{llrrlr}
\toprule
Simulation & Score & $n$ & $\rho$ & 95\% CI & $p$ \\
\midrule
\multirow{3}{*}{Micropolis} & Excess Brier, mid-range & \MPDNModelsBinary & $\MPDRhoBinaryShort$ & \MPDRhoBinaryCIModels & \MPDPBinary \\
 & Excess bits, tail & \MPDNModelsTail & $\MPDRhoTailShort$ & \MPDRhoTailCIModels & \MPDPTail \\
 & Excess nCRPS & \MPDNModelsContinuous & $\MPDRhoContinuousShort$ & \MPDRhoContinuousCIModels & \MPDPContinuous \\
\midrule
\multirow{2}{*}{Starsim} & Excess nCRPS, interventional & 23 & $0.67$ & $[0.34, 0.85]$ & $<$0.001 \\
 & Excess nCRPS, continuous & 24 & $0.75$ & $[0.49, 0.88]$ & $<$0.001 \\
\midrule
\multirow{4}{*}{FreeCiv} & Excess nCRPS & 24 & $0.59$ & $[0.22, 0.82]$ & 0.002 \\
 & Excess bits, tail & 24 & $0.34$ & $[-0.09, 0.67]$ & 0.107 \\
 & Excess Brier, natural conditional & 24 & $0.38$ & $[0.01, 0.65]$ & 0.068 \\
 & Excess Brier, mid-range & 24 & $0.18$ & $[-0.25, 0.55]$ & 0.412 \\
\bottomrule
\end{tabular}

%% file: data/validation_table_forecastbench.tex
\setlength{\tabcolsep}{3pt}
\begin{tabular}{llrlrrlr}
\toprule
 & & \multicolumn{3}{c}{Bivariate} & \multicolumn{3}{c}{Partial, given ECI} \\
\cmidrule(lr){3-5}\cmidrule(lr){6-8}
Simulation & Score & $\rho$ & CI (models) & $p$ & $\rho$ & CI (models) & $p$ \\
\midrule
\multirow{3}{*}{Micropolis} & Excess Brier, mid-range & $\MPDFBRhoBinaryShort$ & \MPDFBRhoBinaryCIModels & \MPDFBPBinary & $-0.13$ & $[-0.59, 0.51]$ & 0.620 \\
 & Excess bits, tail & $\MPDFBRhoTailShort$ & \MPDFBRhoTailCIModels & \MPDFBPTail & $0.12$ & $[-0.41, 0.58]$ & 0.656 \\
 & Excess nCRPS & $\MPDFBRhoContinuousShort$ & \MPDFBRhoContinuousCIModels & \MPDFBPContinuous & $-0.10$ & $[-0.51, 0.44]$ & 0.716 \\
\midrule
\multirow{2}{*}{Starsim} & Excess nCRPS, interventional & $0.45$ & $[-0.14, 0.88]$ & 0.071 & $0.11$ & $[-0.51, 0.77]$ & 0.689 \\
 & Excess nCRPS, continuous & $0.24$ & $[-0.35, 0.82]$ & 0.345 & $-0.26$ & $[-0.73, 0.59]$ & 0.324 \\
\midrule
\multirow{4}{*}{FreeCiv} & Excess nCRPS & $0.21$ & $[-0.32, 0.67]$ & 0.416 & $-0.22$ & $[-0.57, 0.33]$ & 0.419 \\
 & Excess bits, tail & $0.12$ & $[-0.38, 0.56]$ & 0.653 & $-0.10$ & $[-0.48, 0.41]$ & 0.709 \\
 & Excess Brier, natural conditional & $0.38$ & $[-0.18, 0.81]$ & 0.129 & $0.13$ & $[-0.45, 0.73]$ & 0.639 \\
 & Excess Brier, mid-range & $0.28$ & $[-0.24, 0.67]$ & 0.279 & $0.13$ & $[-0.38, 0.57]$ & 0.626 \\
\bottomrule
\end{tabular}

%% file: data/external_benchmarks_table.tex
\begin{tabular}{lrcccc}
\toprule
Benchmark & $n$ & $\theta_m$ & ECI & $\theta_m$ given ECI & ECI given $\theta_m$ \\
\midrule
\multicolumn{6}{@{}l}{\emph{(a) Models each benchmark covers}} \\
FutureEval & 18 & $0.90$ {\scriptsize$[0.71,0.96]$} & $0.89$ {\scriptsize$[0.67,0.97]$} & $0.59$ {\scriptsize$[-0.04,0.88]$} & $0.52$ {\scriptsize$[0.02,0.88]$} \\
Prophet Arena & 9 & $0.86$ {\scriptsize$[0.50,1.00]$} & $0.74$ {\scriptsize$[0.11,1.00]$} & $0.67$ {\scriptsize$[-0.47,1.00]$} & $0.22$ {\scriptsize$[-0.53,1.00]$} \\
ForecastBench & 17 & $0.37$ {\scriptsize$[-0.22,0.81]$} & $0.62$ {\scriptsize$[0.18,0.88]$} & $-0.25$ {\scriptsize$[-0.70,0.49]$} & $0.57$ {\scriptsize$[0.14,0.84]$} \\
\midrule
\multicolumn{6}{@{}l}{\emph{(b) The 14 models both FutureEval and ForecastBench cover}} \\
FutureEval & 14 & $0.85$ {\scriptsize$[0.51,0.97]$} & $0.80$ {\scriptsize$[0.42,0.95]$} & $0.62$ {\scriptsize$[-0.10,0.93]$} & $0.48$ {\scriptsize$[-0.10,0.87]$} \\
ForecastBench & 14 & $0.46$ {\scriptsize$[-0.16,0.89]$} & $0.73$ {\scriptsize$[0.35,0.91]$} & $-0.19$ {\scriptsize$[-0.73,0.71]$} & $0.66$ {\scriptsize$[-0.02,0.88]$} \\
Difference & & $+0.38$ {\scriptsize$[+0.03,+0.79]$} & $+0.07$ {\scriptsize$[-0.25,+0.39]$} & $+0.81$ {\scriptsize$[+0.04,+1.25]$} & $-0.18$ {\scriptsize$[-0.69,+0.31]$} \\
\bottomrule
\end{tabular}

%% file: appendix/F_release.tex
\section{Release and documentation}
\label{app:release}

\changednew{The data release (\url{https://zenodo.org/records/24401629}) contains the questions with their truths, the reports and prompts, every stored model response behind the paper's results and every input of the analysis code, with SHA-256 checksums of all its files. Its README lists the few responses that are missing (below and \cref{app:ablation-batching}). The analysis code is in the analysis repository.}

\subsection{Artifacts, provenance and remaining dependencies}
\label{app:release-contents}

\paragraph{FreeCiv.}
Draw v1.8 has six item files: 750 mid-range, 300 tail, 50 mirror and 300 continuous questions, 400 natural-conditional cells and 124 extra turn-1 questions. They specify text, generated resolution criteria, replay truths and normalization constants; eight turn-60 reports and prompt builders define the elicitation (\cref{app:worlds-freeciv,app:prompts-freeciv}). \changednew{The data release includes a three-file export of aligned natural-conditional replay data}: \texttt{natcond\_replays.npz}, \texttt{natcond\_replays\_cells.csv} and \texttt{natcond\_replay\_tags.csv}. Its boolean $X/Y$ arrays, metadata and tag maps reproduce all 400 cells, accepted probabilities/counts and 9,584 paired forecast scores (\cref{app:full-freeciv}). The original game-save extraction was not independently re-audited.\changednew{ The candidate export, every question the generator enumerated before the draw with its replay truths (23,241 binary and 1,462 continuous candidates), is released; the FreeCiv baselines are computed from it (\cref{sec:design-baselines}).}

\changednew{The data release holds the original model replies behind every FreeCiv forecast used in the analyses.} The 92~GB replay archive is documented as available on request; the derived truth files are 1.1~GB and the compressed corpus 311~MB. The simulator uses CivRealm \citep{qi2024civrealm} and freeciv-web \citep{freecivweb2026}.

\paragraph{Micropolis.}
\changednew{The data release contains the scored binary/continuous forecasts with the original model responses and their predicted quantiles}, usage summaries, \texttt{versions.txt} and \texttt{data/micropolis/continuous\_replays.tar.gz}. The verified archive has 30 JSONL files (15 cities, years 20/30), four horizons and 1,000 outcomes per metric: 840,000 values for seven metrics. All 840 means, 3,000 binary Yes-count probabilities and 600 continuous identifiers match exported summaries. The benchmark repository at commit \texttt{e0f49b5} reproduces it from source: rerunning all 30 snapshots with 1,000 replays each gives every one of its values, and the trunk runs from which the prompts were built are reproduced byte for byte. The repository vendors a prebuilt engine in \path{worlds/micropolis/MicropolisCore}; its command-line runner requires Node and pnpm. The archive records a range of seeds without identifying the seed for each array element. Binary questions keep only Yes counts, not each replay's outcome, so the archive alone does not allow joint resampling across questions; rerunning from source could recover them. \changednew{The other variant ages and the trunk histories used for scales are also in the data release.} \changednew{Six of Llama 4 Scout's original responses for the batching ablation study (\cref{app:ablation-batching}) are missing: in the 50-question setting, the responses behind 300 of its 1,000 forecasts were accidentally overwritten after scoring.} The horizon records share snapshots and do not represent independent sets of continuations. The numerical agreement verifies the available archive.

\paragraph{Starsim.}
The headline uses 32 continuous items per model; 24 historical binary items remain appendix-only. Five historical item files support seven scored sets, and their reference-distribution file belongs to older skill scoring. Draw v4 does not contain repaired continuous truths. \changednew{The corrected continuous run, in the data release,} retains original scores, new truth samples, score/uncertainty outputs, coverage records and hashes. \changednew{The seven call files in the data release match byte for byte the 5,045 successful scored September 16 responses}; they do not cover every failed attempt or establish the numerical precision used to serve the models. Original forecasts are reused, normalization is unchanged, and binary truths are unrepaired (\cref{app:starsim-rescore}).

\changednew{\paragraph{Historical measles.}
The inputs of the measles case study (\cref{sec:real-world-forecasting,app:real-world-forecasting}) are in the data release: the 1,339 state-season series derived from Project Tycho's US measles counts \citep{tycho2018measles}, each model's original responses, parsed forecasts and saved scores, and the mapping from models to ECI values. The analysis repository's \texttt{fbsim-paper build} command checks the saved scores and targets against the series and recomputes the comparison; its parser audit re-parses the original responses and checks them against the parsed forecasts.}

\paragraph{Scripts and exact supporting material.}
The analysis repository's \texttt{fbsim-paper build} command regenerates every generated table, figure and macro of the paper from the data release, with the scripts under \path{src/fbsim_paper/assets/}. Its \texttt{README.md} lists the scripts behind each simulation's outputs, and \path{production/README.md} describes how the inputs were produced. \changednew{The data release also holds the knowledge test's exact prompts (preamble and statement lists) and the OpenRouter identifiers, hosts and per-call usage and costs}, which the knowledge test and some ablations need; the forecasting instructions and question templates are in \cref{app:prompts}.
Limitations of the cost and host records are in \cref{app:roster}.

\paragraph{Licenses and notices.}
\label{app:release-licenses}
The repository's root license does not determine the licenses of its dependencies. Starsim 3.3.4 is MIT; the vendored MicropolisCore engine includes GPL and name/trademark terms, kept in its license files. \changednew{The data release is licensed under CC BY 4.0; the model responses in it remain subject to their providers' terms. Its \texttt{licenses/} folder holds the Micropolis, FreeCiv (GPL version 2) and Starsim (MIT) licenses. The historical measles counts are from Project Tycho, which distributes them under CC BY 4.0.}
SimCity and Civilization are trademarks of their owners; this work is not affiliated with either.

\subsection{Refresh and versioning limits}
\label{app:release-refresh}
Selecting questions from existing replays is inexpensive; fresh simulations and model calls cost money. New seeds and draws can reduce answer reuse, but cannot guarantee harder questions, contamination immunity or freedom from saturation: simulator knowledge, released information and other exposure remain relevant. Preserve old draws and version new ones with source and run identifiers; check score comparability across draws. Recorded run costs (\cref{app:roster}) are not full-refresh prices.